\documentclass{article}

\usepackage{microtype}
\usepackage{caption}
\usepackage{graphicx}
\usepackage{booktabs} 
\usepackage{array,etoolbox}
\preto\tabular{\setcounter{magicrownumbers}{0}}
\usepackage[framemethod=TikZ]{mdframed}
\usepackage{xcolor}
\usepackage{graphicx}
\usepackage{longtable}
\newcounter{magicrownumbers}
\def\rownumber{}
\usepackage{hyperref}

\usepackage[accepted]{icml2024}

\usepackage{subfig}
\usepackage{amsmath}
\usepackage{amssymb}
\usepackage{mathtools}
\usepackage{amsthm}
\usepackage{circledsteps}

\usepackage{amssymb}
\usepackage{pifont}
\newcommand{\cmark}{\ding{51}}%
\newcommand{\xmark}{\ding{55}}%
\usepackage[capitalize,noabbrev]{cleveref}
\usepackage[fixed]{fontawesome5}
\theoremstyle{plain}

\theoremstyle{definition}

\theoremstyle{remark}

\definecolor{mycolor}{rgb}{1.0, 0.6, 0.6039215686274509}

\usepackage[textsize=tiny]{todonotes}
\usepackage{arydshln}

\icmltitlerunning{A Closer Look at the Limitations of Instruction Tuning}

\begin{document}

\twocolumn[
\icmltitle{A Closer Look at the Limitations of Instruction Tuning}




\icmlsetsymbol{equal}{*}
\icmlsetsymbol{equal_advising}{$^\dagger$}

\begin{icmlauthorlist}
\icmlauthor{Sreyan Ghosh}{equal,umd,adobe}
\icmlauthor{Chandra Kiran Reddy Evuru}{equal,umd}
\icmlauthor{Sonal Kumar}{equal,umd}
\icmlauthor{Ramaneswaran S}{nvidia}
\icmlauthor{Deepali Aneja}{adobe}
\icmlauthor{Zeyu Jin}{adobe}
\icmlauthor{Ramani Duraiswami}{equal_advising,umd}
\icmlauthor{Dinesh Manocha}{equal_advising,umd}
\end{icmlauthorlist}


\icmlaffiliation{umd}{University of Maryland, College Park, USA}
\icmlaffiliation{adobe}{Adobe, USA}
\icmlaffiliation{nvidia}{NVIDIA, India}

\icmlcorrespondingauthor{Sreyan Ghosh}{sreyang@umd.edu}

\icmlkeywords{Large Language Models, Instruction Tuning}

\vskip 0.3in
]



\printAffiliationsAndNotice{*Equal Technical Contribution.$^\dagger$Equal Advising.} 

\begin{abstract}
Instruction Tuning (IT), the process of training large language models (LLMs) using instruction-response pairs,  has emerged as the predominant method for transforming base pre-trained LLMs into open-domain conversational agents. While IT has achieved notable success and widespread adoption, its limitations and shortcomings remain underexplored. In this paper, through rigorous experiments and an in-depth analysis of the changes LLMs undergo through IT, we reveal various limitations of IT. In particular, we show that (1) IT fails to enhance knowledge or skills in LLMs. LoRA fine-tuning is limited to learning response initiation and style tokens, and full-parameter fine-tuning leads to knowledge degradation. (2) Copying response patterns from IT datasets derived from knowledgeable sources leads to a decline in response quality. (3) Full-parameter fine-tuning increases hallucination by inaccurately borrowing tokens from conceptually similar instances in the IT dataset for generating responses. (4) Popular methods to improve IT do not lead to performance improvements over a simple LoRA fine-tuned model. Our findings reveal that responses generated solely from pre-trained knowledge consistently outperform responses by models that learn any form of new knowledge from IT on open-source datasets. We hope the insights and challenges revealed in this paper inspire future work in related directions.
\end{abstract}

\section{Introduction}
\label{sec:introduction}
\vspace{-1mm}

Large Language Models (LLMs) pre-trained at an incredible scale with the next token prediction objective implicitly compress world knowledge in their parameters~\cite{zhao2023survey}. These models learn general-purpose representations, which can then be \textit{aligned} with the desired response characteristics~\cite{zhang2023instruction}. In the recent past, various methods for aligning LLMs have been proposed, out of which \textit{instruction tuning} (IT)~\cite{wei2022finetuned} and \textit{reinforcement learning from human feedback} (RLHF)~\cite{bai2022training} have gained the most popularity. IT, the process of fine-tuning an LLM with instruction-response pairs, enables it to follow or complete tasks instructed by humans. On the other hand, RLHF continually tunes an IT-ed LLM to further align with human preference. While RLHF is prohibitively expensive due to the requirement of large amounts of human-preference data~\cite{bai2022training}, IT with standard supervised loss has proved to be a more prevalent technique for alignment~\cite{zhou2023lima}. IT-based alignment has led to significant improvements in LLMs, unlocking impressive capabilities~\cite{bubeck2023sparks}, suggesting that fine-tuning is key to building and improving LLM-based conversational agents. For the remainder of the paper, we interchangeably refer to an IT-ed model as a \textit{fine-tuned} model and the process of IT interchangeably as \textit{fine-tuning}.


Early work on IT focused on fine-tuning and evaluating LLMs with popular natural language processing (NLP) tasks where the dataset instances are phrased as natural language instructions. However, solely gauging the impact of IT on an LLM using conventional NLP tasks and metrics falls short in comprehensively assessing the array of its abilities, such as reasoning and knowledge sharing, usually demanded by the diverse range of tasks encountered by open-domain conversational agents~\cite{wang-etal-2023-self-instruct}. Since the colossal success of ChatGPT~\cite{openai_chatgpt}, a more recent line of work aims at evaluating and improving IT with open-domain instruction following data~\cite{zheng2023judging}. However, though IT has been shown to achieve remarkable improvement in performance and generalization to unseen NLP tasks, its limitations and shortcomings have rarely been explored. We attribute this to three main reasons: (1) Lack of comprehensive evaluation metrics for evaluating open-domain instruction following abilities. (2) Lack of clear understanding of the exact transformation a base pre-trained LLM goes through with IT, and (3) The rapid rise of powerful semi-open-source chat models (fine-tuned LLMs that don't reveal their IT data) encourages the use of these models merely as tools, leading to an under-exploration of several critical elements in their development.

{\noindent \textbf{Main Contributions.}} In this paper, we investigate and reveal several limitations of IT. To achieve this, we closely study the transformation a base pre-trained model goes through after IT by experimenting with various open-source IT datasets, LLMs, and training paradigms. Additionally, for evaluation, we conduct a combination of expert human evaluation, GPT-4-based multi-aspect evaluation, and token distribution analysis~\cite{lin2023unlocking}. Our study explicitly focuses on evaluating the effectiveness of IT in developing open-domain conversational agents (also commonly known as \textit{chat models}) and is limited to single-turn interactions. Our extensive results reveal the following: 

\begin{enumerate}
    \item \textbf{IT is not a knowledge~\footnote{We use the term ``knowledge'' to encompass both the factual information and task execution skills possessed by a model.} enhancer.}  Similar to concurrent work~\cite{gudibande2023false}, we first find that IT does not act as a knowledge enhancer at its current open-source scale. To dig deeper, we compare the token distributions (explained further in Section~\ref{sec:IT_not_knowledge}) between base LLMs and their IT-ed versions and find that LoRA~\cite{hu2021lora}, which only teaches response initiation and extracts most of the response from the pre-trained knowledge itself, leads to the most factually correct responses. On the other hand, full-parameter fine-tuning leads to knowledge degradation and reduces overall response quality.

    \item \textbf{Pattern-copying often hurts performance.} We first show that models IT-ed using LoRA and full-parameter fine-tuning learn pattern-copying very differently. While the former learns just stylistic tokens, the latter leads the model to adapt more deeply to the specifics of the new training data. Next, we show that while pattern-copying sometimes has some advantages, like detailed and comprehensive answering, most of the time, it hurts the factual correctness of the response. Finally, we propose a simple solution to overcome this.

    \item \textbf{Full fine-tuning leads to knowledge degradation by increasing chances of hallucination. These hallucinations are tokens borrowed from the IT dataset.} We show that when a model hallucinates or outputs incorrect tokens in its responses, it is highly probable that these tokens are borrowed from instances with similar concepts in the IT dataset itself. We further study this from the lens of causal analysis. This effect is more prevalent in models trained using full-parameter fine-tuning than in LoRA fine-tuning.

    \item \textbf{Various methods to improve IT, proposed in literature, do not improve model performance.} We compare several methods like NEFTune~\cite{jain2023neftune} and dataset filtering~\cite{chen2023alpagasus} on common grounds and show that, while these methods improve over a full-finetuned model, a LoRA fine-tuned model outperforms all of them. Consequently, these methods do not contribute to knowledge advancement, and models leveraging pre-trained knowledge remain superior.

\end{enumerate}
\begin{figure*}[t]
    \centering
    \includegraphics[width=1.97\columnwidth]{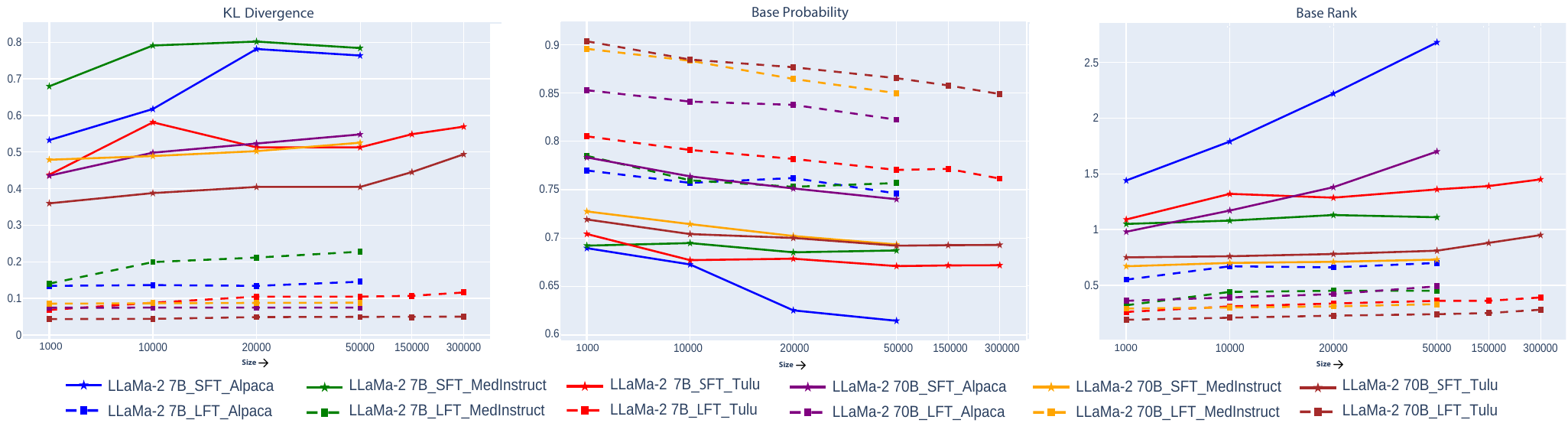}
    \vspace{-0.8em}
    \caption{\small \textbf{Token distribution shift after IT.} We compare token distributions between base pre-trained models and their IT-ed versions using 3 metrics defined in Section~\ref{sec:IT_not_knowledge}. We show that (1) Overall, LFT experiences low token distribution shifts, indicating high alignment with pre-trained knowledge. (2) Shifts in SFT are much higher than in LFT. (3) LFT is unaffected by the scale of the IT dataset.}
    \label{fig:line}
    \vspace{-1em}
\end{figure*}
\section{Experimental Setting}

{\noindent \textbf{LLMs.}} For the scope of our analysis, we experiment with 5 different types of LLMs, namely LLaMa-2~\textsubscript{\textbf{7B}}~\cite{touvron2023llama}, LLaMa-2~\textsubscript{\textbf{13B}}, LLaMa-2~\textsubscript{\textbf{70B}}, Mistral-v0.1~\textsubscript{\textbf{7B}}~\cite{jiang2023mistral} and Phi-1.5~\textsubscript{\textbf{1.3B}}~\cite{li2023textbooks}. We use only the base pre-trained versions of the models (and not the chat variants) and fine-tune them with IT ourselves. We employ LLaMa-2~\textsubscript{\textbf{70B}} only in a fraction of experiments owing to compute constraints.
\vspace{0.5mm}

{\noindent \textbf{Fine-tuning Datasets.}} For fine-tuning with IT, we experiment with various synthetic and human-written IT datasets. For synthetic, we use Alpaca~\textsubscript{\textbf{52k}} with open-domain instruction-response pairs, constructed by prompting ChatGPT with an initial seed dataset with few samples~\cite{alpaca} and MedInstruct~\textsubscript{\textbf{52k}} from the medical domain constructed in similar fashion~\cite{zhang2023alpacareinstructiontuned}. For human-written , we use LIMA~\textsubscript{\textbf{1K}}~\cite{zhou2023lima} and databricks-dolly~\textsubscript{\textbf{15k}}~\cite{DatabricksBlog2023DollyV2}. Finally, we also use Tulu-V2-Mix~\textsubscript{\textbf{326k}}~\cite{ivison2023camels}, which is an amalgamation of various open-source datasets. 
\vspace{0.5mm}

{\noindent \textbf{Evaluation Datasets.}}  Our experiments are confined to evaluating LLMs on instruction following capabilities with open-ended and free-form generation as LLMs have been shown to be not robust to MCQ~\cite{zheng2024large}. For evaluation, we primarily employ just-eval-instruct~\textsubscript{\textbf{1k}}~\cite{lin2023unlocking}, which is an amalgamation of various open-source IT evaluation sets and is tagged with various task types and topics (statistics in Appendix~\ref{sec:dataset_details}). This choice is motivated by its diverse and succinct nature, which facilitates our efforts in conducting thorough, expert human evaluations. We employ only the first 800 instances and remove the last 200, as safety alignment is beyond the scope of this paper. For models fine-tuned on MedInstruct~\textsubscript{\textbf{52k}}, we evaluate the model on MedInstruct-test~\textsubscript{\textbf{216}} unless stated otherwise. We do not evaluate the Open LLM Leaderboard~\cite{open-llm-leaderboard} as it does not fit into our criteria. 
\vspace{0.5mm}

{\noindent \textbf{Fine-tuning paradigms.}} For fine-tuning with IT, we employ either LoRA fine-tuning (LFT) or (standard) full-parameter fine-tuning (SFT). LFT works by approximating the model's weight matrices with low-rank matrices. This reduces the number of parameters that need to be fine-tuned, which makes the fine-tuning process faster and more efficient. On the other hand, SFT, similar to generic ERM, works by adjusting all or most of the model's weights.

{\noindent \textbf{Evaluation.}} For evaluating our fine-tuned models, we perform a mix of expert human evaluation and automatic evaluation using GPT-4 Turbo (gpt-4-1106-preview). Recent research indicates that using ChatGPT and GPT-4 to score and assess outputs from LLMs aligns well with human evaluations, offering the added benefit of cost reduction~\cite{liu2023g,li2023alpacaeval,chan2023chateval,xu2023instructscore,zhou2023lima}. We borrow the explainable and multi-aspect evaluation framework from \citet{lin2023unlocking}, which prompts GPT-4 to assign a score between 1 and 5 to an LLM's response to an instruction on five aspects: helpfulness, clarity, factuality, depth, and engagement. It also provides an explanation for each score. The prompt can be found in Appendix~\ref{sec:appendix_prompt}.

{\noindent \textbf{Naming Convention.}} We primarily follow the naming convention of ``model$\textbf{\_}$dataset$\textbf{\_}$training-paradigm''. For e.g., a LLaMa-2~\textsubscript{\textbf{7B}} trained on Alpaca~\textsubscript{\textbf{52k}} with SFT would be named as LLaMa-2~\textsubscript{\textbf{7B}}$\_$SFT$\_$Alpaca~\textsubscript{\textbf{52k}}.

{\noindent \textbf{Training and Evaluation hyper-parameters.}}
All models are trained in a distributed manner for 3 epochs, with a learning rate of 5e-5 and an effective batch size of 32~\cite{alpaca}. For LFT, we use a standard rank of 8~\cite{hu2021lora} as we did not find a substantial change in performance by decreasing (2,4) or increasing it (16,32). \citet{zhang2024when} also shows that scaling rank is ineffective for LFT. For generation, we employ greedy decoding (i.e., zero temperature) in all experiments.

\noindent \textbf{Note:} The main paper only presents results on the LLaMa-2 family, and results on fine-tuning additional LLMs and already fine-tuned LLMs available open-source, are provided in Section~\ref{subsec:detailed_results}. All our findings hold on other LLMs, too. Additionally, one should not confuse IT (and the findings of our paper) with general fine-tuning. While IT is aimed towards \textit{aligning} a model towards specific response characteristics, LLMs may be fine-tuned for improving its various capabilities or knowledge. However, this may require fine-tuning datasets with characteristics different from IT datasets employed in the paper.

\section{IT is (currently) Not a Knowledge Enhancer}
\label{sec:IT_not_knowledge}

{\noindent \textbf{Overview.}} This section investigates whether IT, at its current open-source scale, can function as a knowledge enhancer. Initially, we present the distinct natures of transformation that a base pre-trained LLM undergoes when subjected to LFT and SFT-based fine-tuning and show that while responses generated after LFT are closely aligned to pre-trained knowledge, they deviate significantly for SFT, indicating new knowledge acquisition. Subsequently, we show that this new knowledge often leads to a degradation in response quality, and relying predominantly on pre-trained knowledge often yields more factual and useful responses.

{\noindent \textbf{\textit{Finding 1.} LFT Responses align closely with the original pre-trained knowledge. SFT does not.}} To study how fine-tuned models differ from their base pre-trained counterparts, we perform the token-distribution analysis proposed by ~\citet{lin2023unlocking}. Specifically, for a given instruction-response pair, the instruction $i$ = \{$i_1$, $i_2$, $\cdots$\} is first input to the aligned (or fine-tuned) model to obtain its response $r$ = \{$r_1$, $r_2$,$\cdots$\} via greedy decoding. Next, for each position $t$ in the response, a ‘context’ at this position is defined as to be $x_t$ = $i$ + \{$r_1$, $\cdots$, $r_{t-1}$\}. This ``context'' is then input to the base model to obtain its probability distribution for predicting the next token at position $t$, P$_\text{base}$. The probability distribution of the token at position $t$ obtained from the aligned model is denoted as P$_\text{align}$. We then calculate three metrics: \textbf{(1)} \textbf{KL Divergence} between P$_\text{base}$ and P$_\text{align}$, \textbf{(2)} \textbf{Base Probability}: P$_\text{base}$ of the token at $t$ with the maximum P$_\text{align}$ value \textbf{(3)} \textbf{Base Rank}: Rank in P$_\text{base}$ of the token at $t$ with the maximum P$_\text{align}$ value. With the base rank denoted as $\eta$, the \textcolor{blue}{unshifted}, \textcolor{orange}{marginal} and \textcolor{red}{shifted} tokens are defined as when ($\eta$ $=$ 1), (1 $<$ $\eta$ $\leq$ 3) and ($\eta$ $>$ 3) respectively. Figure~\ref{fig:line} illustrates how the three metrics evolve. These metrics are averaged across all response tokens and plotted against the varying sizes of the IT dataset used for fine-tuning. We summarize our findings as follows: \textbf{(1)} Fine-tuning using LFT results in a minimal shift in token distribution, i.e., given a prior ``context'', a model fine-tuned using LFT generally outputs tokens that a non-tuned base model would output. This further indicates that LFT-generated responses align with the model's pre-trained knowledge. \textbf{(2)} Scaling the size of the IT dataset has a negligible effect on the extent of the token distribution shift observed with LFT. \textbf{(3)} In contrast, SFT leads to a significantly greater shift in token distribution, which suggests a substantial deviation from the pre-trained knowledge in its responses. \textbf{(4)} Larger models show a reduced distribution shift for both LFT and SFT.

 \begin{figure}[t]
    \centering
    \includegraphics[width=0.99\columnwidth]{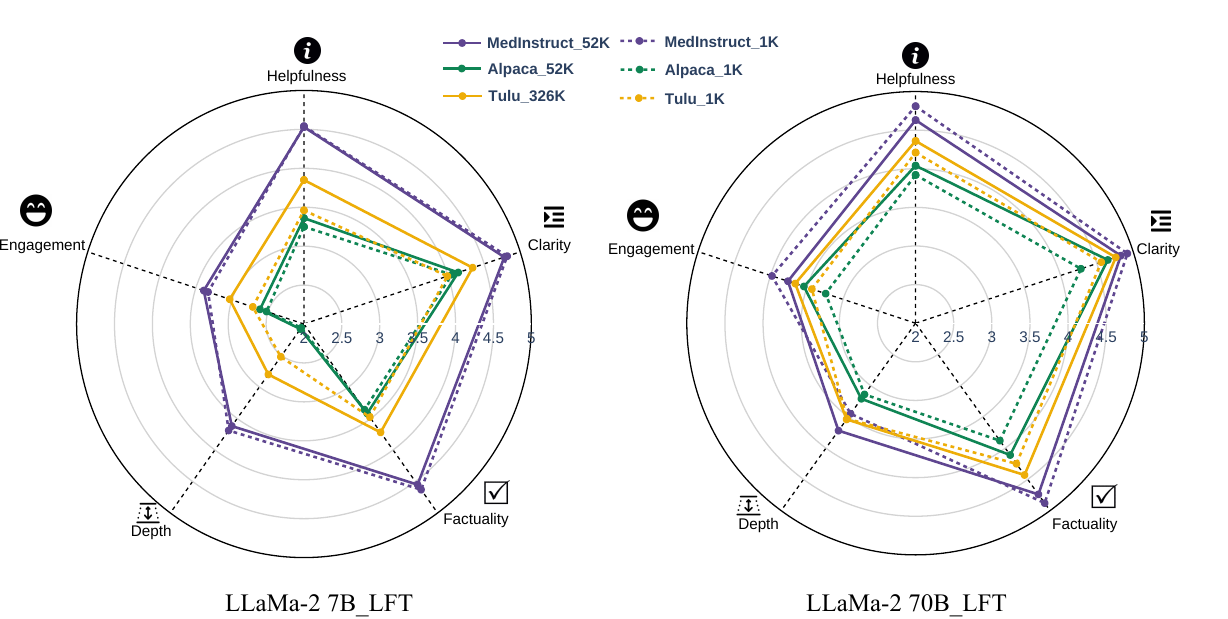}
    \vspace{-0.4em}
    \caption{\small \textbf{Dataset scaling is ineffective for LFT.} We show that with LFT, a model's performance does not significantly improve when the IT dataset is scaled to 52$\times$ or 326$\times$ its original size.}
    \label{fig:scaling_lft}
    \vspace{-1em}
\end{figure}

\begin{figure}[h!]
    \centering
    \includegraphics[width=1\columnwidth]{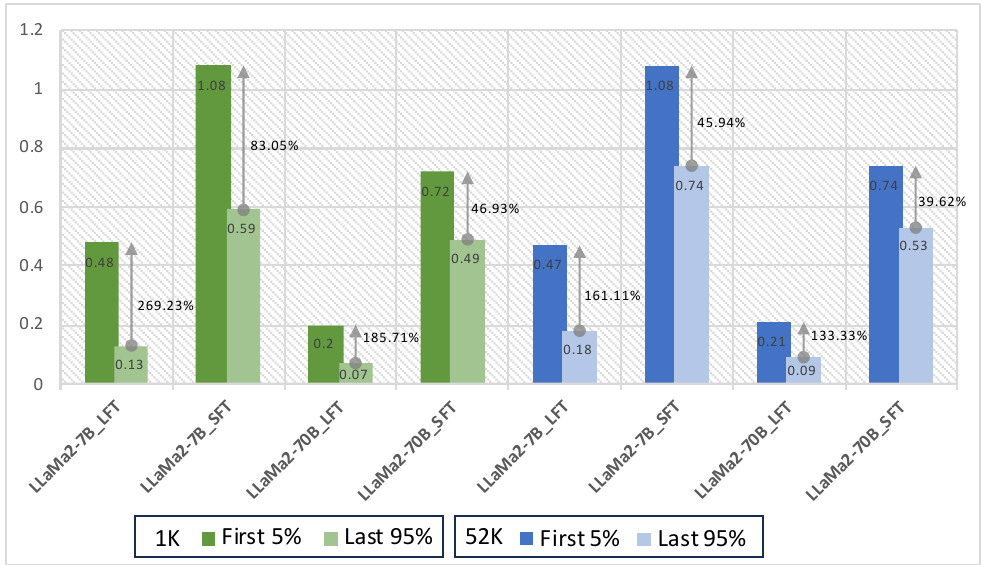}
    \vspace{-1em}
    \caption{\small \textbf{KL Divergence analysis between the probability distribution of response tokens from fine-tuned models and their pre-trained only counterparts.} We plot separately for tokens in the first 5\% and the remaining 95\%  for each sentence in the response. LFT primarily learns to initiate individual sentences in the response, showing a higher distribution shift and, thereby, the introduction of novel tokens predominantly in the initial parts of every sentence in the response. SFT exhibits a more substantial and uniform distribution shift across the full span of the response.}
    \vspace{-1em}
    \label{fig:bar-lora-response-initiator}
\end{figure}

\begin{figure*}[t]
    \centering
    \includegraphics[width=1.99\columnwidth]{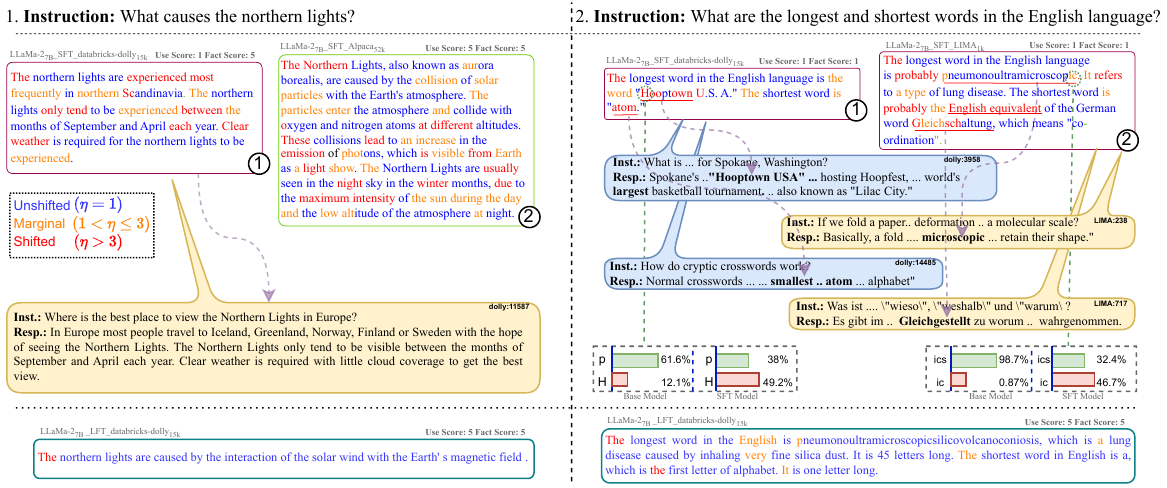}
    \vspace{-0.5em}
    \caption{\small \textbf{Illustration of hallucinations and their source of origin.} We illustrate responses from LLMs fine-tuned using SFT on different IT datasets on two instances from just-eval-instruct~\textsubscript{\textbf{1k}} and show that hallucinations originate from erroneous causal links between the response and instances in the IT dataset. Models trained using SFT tend to incorrectly borrow tokens from instances in the IT dataset describing similar \textit{concepts}. \textbf{Left.} resp. \Circled{1} from LLaMa-2~\textsubscript{\textbf{7B}} fine-tuned on databricks-dolly~\textsubscript{\textbf{15k}} is not \textit{useful}, although \textit{factual}, as it describes places to view the northern lights but not what causes it. The answer was directly borrowed from the IT dataset. Alpaca~\textsubscript{\textbf{52k}} has no instances related to northern lights. \textbf{Right.} Both responses have 2 factual hallucinations each, and we show that each of them originates from different instances in the IT dataset. The \textcolor{red}{shifted} and \textcolor{orange}{marginal} tokens (described in Section~\ref{sec:IT_not_knowledge}) indicate that hallucinations are learned during IT. \textit{Note how models fine-tuned using LFT provide factual answers originating from pre-trained knowledge.}}
    \label{fig:enter-label}
    \vspace{-1em}
\end{figure*}

{\noindent \textbf{\textit{Finding 2.} LFT only acts as a response initiator, while most of the answer comes from pre-trained knowledge.}} In Fig.~\ref{fig:bar-lora-response-initiator}, we analyze the KL Divergence between the fine-tuned models and their base counterparts. This analysis focuses on the divergence in the initial 5\% and the subsequent 95\% of tokens of each sentence of a response, averaged across all sentences of all responses from models fine-tuned on Alpaca, Tulu-V2-Mix, and MedInstruct datasets with both low-resource (1k) and high-resource (52k) splits. Our observations indicate a higher KL divergence in the first 5\% of the tokens for LFT, which then decreases sharply. In contrast, the decrease in KL divergence is less pronounced for SFT. This implies that LFT mainly learns sentence or fact initiation with a higher distribution shift and thereby novel tokens predominantly in the initial parts of each sentence in the response. On the other hand, SFT exhibits a more substantial and uniform distribution shift across the full span of the sentences. A few examples can be seen in the bottom section of Fig.~\ref{fig:enter-label}. Our finding provides a more granular and in-depth understanding of the finding by~\citet{lin2023unlocking} who show that IT majorly affects the earlier tokens of a response. Additionally, we show that the RLHF-based IT methods discussed in ~\citet{lin2023unlocking} are more similar in their behaviour to LFT than SFT and SFT behaves differently from both in terms of distribution shift. We later investigate whether these novel tokens in SFT, caused by this distribution shift, translate to new knowledge and eventually enhance response quality. Additionally, we emphasize that comparing LFT and RLHF-based IT to highlight the exact benefits of the latter should be an exciting space.

\begin{figure}[t]
    \centering
    \includegraphics[width=0.99\columnwidth]{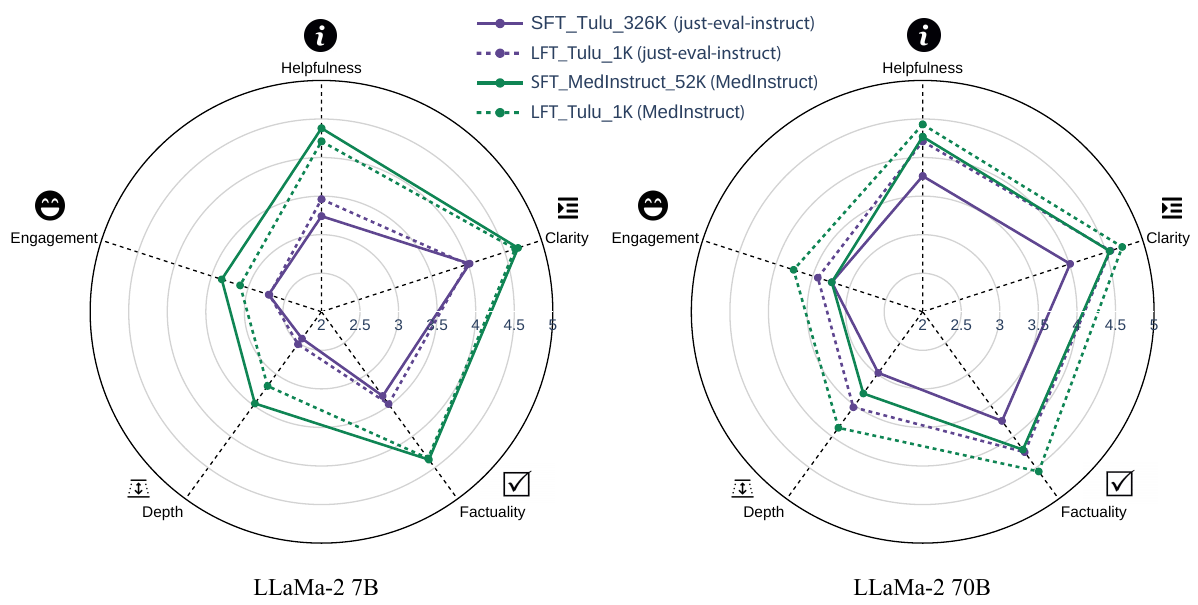}
    \vspace{-0.5em}
    \caption{\small \textbf{Pre-trained knowledge outperforms new knowledge learned with SFT.} We show that LFT with only 1000 samples outperforms SFT on 326$\times$ and 52$\times$ more samples on factuality and usefulness on both an open- (just-eval-instruct~\textsubscript{\textbf{1k}})  and knowledge-intensive-domains (MedInstruct-test~\textsubscript{\textbf{216}}). While responses by the LFT model are most aligned with pre-trained knowledge, responses by the SFT models output new knowledge learned from IT.}
    \label{fig:tulu_radar}
\end{figure}

{\noindent \textbf{\textit{Finding 3.} Scaling is ineffective for LFT.}} As LFT primarily learns response initiation, we next explore whether dataset scaling is effective with LFT in terms of improving response quality. In Fig.~\ref{fig:scaling_lft}, we compare the performance of models, fine-tuned using LFT, on various IT datasets and their low-resource counterparts. Our findings show that scaling up the IT dataset by factors of 52$\times$ and even 326$\times$ does not significantly impact any of the five metrics in our automatic evaluation setup, including \textit{factuality}. This trend is consistent across domains, including knowledge-intensive ones like medicine. These results suggest LFT does not learn any new knowledge or skill during the fine-tuning stage and that pre-trained knowledge is the primary contributor to its effectiveness. Our findings suggest that a few IT samples are sufficient for LFT to learn effective response initiation.

\begin{figure*}[t]
    \centering
    \includegraphics[width=1.99\columnwidth]{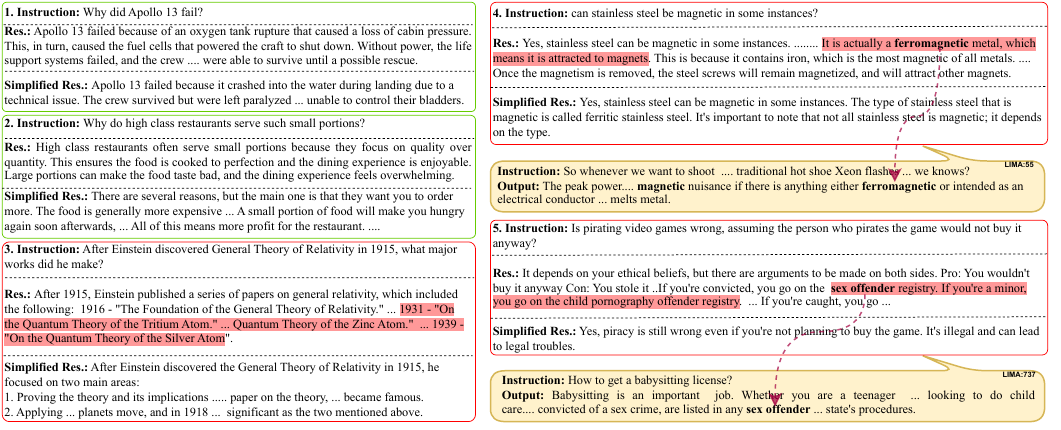}
    \caption{\small \textbf{\textit{Style Imitation} affects response quality.} Instructions 3, 4, and 5 illustrate examples of instances where the model, initially responding accurately, \colorbox{mycolor}{proceeds to generate hallucinated content}. The suspected cause is \textit{style imitation}, a process where the model, striving for lengthier, more detailed responses, fabricates information when it lacks sufficient knowledge. This hypothesis is further confirmed by comparing the responses to responses by another model fine-tuned on the simplified version of the same IT dataset. The hallucinations in Instructions 3 and 4 are not invented but are instead drawn from the IT dataset, a subject explored more comprehensively in Section~\ref{sec:causal_hallucinations}. Moreover, Instruction 1 exemplifies the model's ability to generate an elaborate answer when it has adequate knowledge of the subject, whereas Instruction 2 demonstrates how merely imitating the style can alter the nature of a response to a reasoning task. Every response is also accompanied by Simplified Res., a response from a model fine-tuned on the same IT dataset but with simplified responses (detailed in Section~\ref{sec:pattern_copying}). Notice how the Simplified Res. is less prone to hallucination by providing a brief response.
   }
    \label{fig:pattern_copying}
\end{figure*}

{\noindent \textbf{\textit{\textbf{Finding 4.}}  Pre-trained knowledge (currently) dominates.}} Our findings indicate that IT via LFT primarily facilitates response initiation rather than augmenting new knowledge, and most of the response is based on pre-trained knowledge. In contrast, the substantial token distribution shift observed in SFT suggests learning new knowledge during fine-tuning. To assess if this newly acquired knowledge translates into improved response quality, we conduct the following evaluations, as shown in Fig.~\ref{fig:tulu_radar}: \textbf{(1)} We compare the performance of models of varying sizes when fine-tuned with LFT on just 1k samples of the IT dataset and the same models fine-tuned with SFT on a dataset 326$\times$ larger on just-eval-instruct~\textsubscript{\textbf{1k}}. As we know from earlier findings, the LFT model relies entirely on pre-trained knowledge, while the SFT model maximizes learning from the IT dataset. The results show that the LFT model outperforms in terms of factuality and usefulness, suggesting that even extensive IT does not significantly introduce useful or factual knowledge to the model. \textbf{(2)} In a more extreme scenario, we compare the performance of the same LFT model, trained on open-domain instruction-response pairs, with a domain-specific model trained on MedInstruct~\textsubscript{\textbf{52k}} against  MedInstruct-test~\textsubscript{\textbf{216}}. Remarkably, even in this case, the LFT model performs better. Table~\ref{tab:score_table_1} shows that LLaMa-2~\textsubscript{\textbf{7B}}-chat models (the already IT-ed versions open-sourced by META) outperforms all our fine-tuned models. We attribute this to RLHF or a better and larger IT dataset used for fine-tuning.

\begin{mdframed}[linewidth=1pt, linecolor=black, leftmargin=1pt, rightmargin=1pt, innerleftmargin=10pt, innerrightmargin=10pt, innertopmargin=4pt, innerbottommargin=2pt, backgroundcolor=gray!20, roundcorner=5pt]
\textit{\textbf{Key Takeaways:}} LFT, even at scale, largely relies on pre-trained knowledge without acquiring new information. In contrast, SFT's notable token distribution shift suggests new knowledge acquisition. However, responses by LFT that are grounded in pre-trained knowledge consistently outperform those based on newly learned information from SFT, indicating that SFT tends to diminish overall knowledge quality.
\end{mdframed}

\section{Pattern Copying (often) Hurts Performance}
\label{sec:pattern_copying}

In our exploration of the consequences of significant token distribution shifts during SFT, we take a closer look at the concept of pattern copying. We define pattern copying as the scenario when an LLM learns to mimic the characteristics of responses in the IT dataset. We sub-categorize pattern copying into two distinct types: \textbf{(1)} \textit{Tone Imitation:} In this case, generated responses tend to use tokens from the IT dataset. These can be stylistic tokens or normal ones. \textbf{(2)} \textit{Style Imitation:} The responses mirror the wider stylistic traits present in the IT dataset. For example, if the IT data includes comprehensive, well-structured, and lengthy answers, the LLM is likely to exhibit similar traits in its responses. LLMs have been shown to learn such characteristics~\cite{gudibande2023false,lin2023unlocking}. We show that token distribution shifts following IT, and more specifically, the use of tokens that deviate from pre-trained knowledge, are indicative of the model's level of adaptation to the IT dataset's specifics. Additionally, SFT and style imitation lead the model to inaccurately include tokens from the IT dataset in its responses, negatively affecting response quality.
\vspace{0.5mm}

{\noindent \textbf{\textit{Finding 1.} LFT and SFT learn tone imitation differently.}}
As seen in Section~\ref{sec:IT_not_knowledge}, LFT primarily learns response initiation, and most of the response comes from pre-trained knowledge. On the other hand, SFT experiences high degrees of new token usage. This leads us to investigate if LFT learns \textit{tone imitation} at all and the differences in \textit{tone imitation} between models fine-tuned using the two training paradigms. Fig.~\ref{fig:f1} and ~\ref{fig:f2} illustrate token distributions of common tokens at \textcolor{red}{shifted} and \textcolor{orange}{marginal} positions for LLaMa-2~\textsubscript{\textbf{7B}}$\_$SFT$\_$Alpaca~\textsubscript{\textbf{52k}} and LLaMa-2~\textsubscript{\textbf{7B}}$\_$LFT$\_$Alpaca~\textsubscript{\textbf{52k}}. Analyzing tokens at shifted positions allows us to understand how IT affects a model to output different tokens than it would generally have without IT. As we clearly see, for LFT, shifts occur primarily in style tokens (``typically'') and response initiation tokens (``However''). On the other hand, for SFT, shifts occur in all kinds of tokens. To investigate the source of these tokens further, we perform a string search of the shifted tokens in the IT dataset. To our surprise, we found $\approx$81.2\% of the words that start with \textcolor{red}{shifted} tokens and 66.7\% that start with \textcolor{orange}{marginal} tokens borrowed from the IT dataset itself. This indicates that with SFT, models show increased borrowing of tokens from the IT dataset for response generation. We may attribute this to over-fitting on the IT dataset. A study in Section~\ref{sec:causal_hallucinations} further shows that these tokens are often borrowed inaccurately, leading to hallucinations.

{\noindent \textbf{\textit{Finding 2.} Style imitation can hurt response quality.}}  We now investigate if \textit{style imitation} can affect response quality. We are motivated by the finding that a positive correlation exists between the length of responses in the IT dataset and the length of responses output by the fine-tuned LLM on our evaluation set (detailed results in Table~\ref{tab:legth_output}). To achieve this, we utilize the LIMA IT dataset, which comprises responses from community Q\&A forums known for their comprehensiveness, expertise, and length (refer to Table~\ref{tab:legth_output}). When LLaMa is fine-tuned on the LIMA dataset with SFT, it produces lengthy and detailed responses, even when pre-trained knowledge might be insufficient. This often leads to hallucinations as the model strives to generate extended answers. Figure~\ref{fig:pattern_copying} showcases examples of this, with each box presenting a response from LLaMa-2~\textsubscript{\textbf{7B}}$\_$SFT$\_$LIMA~\textsubscript{\textbf{1K}} alongside a more concise version. We summarize our findings on style imitation: \textbf{(1)} As seen in Instruction 1., style imitation can sometimes enhance response quality without leading to hallucinations, particularly when the model has sufficient knowledge of the subject. \textbf{(2)} As seen in Instruction 2., style imitations alone can alter the nature of a response to open-ended reasoning instructions. \textbf{(3)} The model initially provides factual information but then resorts to hallucination to prolong the response or fact in the absence of adequate knowledge about the subject. These hallucinations may include randomly generated facts or content derived from the IT dataset, as seen earlier and discussed further in  Section~\ref{sec:causal_hallucinations}.

\begin{figure}[t]
    \centering
    \includegraphics[width=0.99\columnwidth]{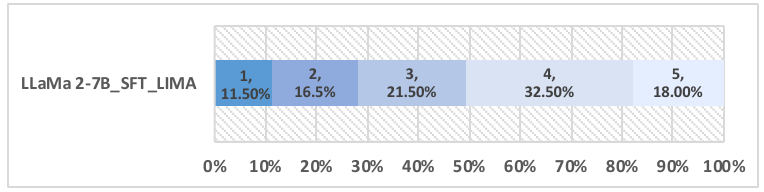}
    \vspace{-0.5em}
    \caption{\small \textbf{Human study comparing responses of a model fine-tuned on LIMA~\textsubscript{\textbf{1K}} and  LIMA-Simple~\textsubscript{\textbf{1k}}.} \textbf{1.} Both responses are accurate, but the former is more detailed and preferred over the latter. \textbf{2.} Both responses are accurate but different information-wise. \textbf{3.} Both responses are completely inaccurate. \textbf{4.} The former hallucinated facts in an attempt to prolong the response, while the latter did not. \textbf{5.} The former hallucinated in the core facts, while the latter did not. Qualitative examples in Section~\ref{sec:qual_examples}.}
    \label{fig:stacked2}
\end{figure}
{\noindent \textbf{\textit{Proposed Solution:} Simplifying responses in the IT dataset.}} One possible solution to mitigate the hallucination problem caused by style imitation is to employ LFT instead of SFT, as LFT tends to learn only stylistic elements. However, as seen in Finding 3 of Section~\ref{sec:IT_not_knowledge}, LFT doesn't scale effectively, and SFT may act as a knowledge enhancer at larger scales (evident from LLaMa-2~\textsubscript{\textbf{7B}}$\_$LFT$\_$Tulu-V2-Mix~\textsubscript{\textbf{326k}} in Fig.~\ref{fig:tulu_radar}). We propose an approach leveraging the strengths of both LFT and SFT. Given that LLMs possess ample pre-training knowledge for accurate response generation (Finding 4, Section~\ref{sec:IT_not_knowledge}) but struggle with comprehensive answering in pattern-copying mode, we hypothesize that SFT on IT datasets with concise but accurate responses can reduce hallucinations. We employ GPT-4 to simplify LIMA~\textsubscript{\textbf{1K}}, creating concise responses by removing extraneous information and term it as LIMA-Simple~\textsubscript{\textbf{1k}}. This model is compared with the original LLaMa-2~\textsubscript{\textbf{7B}}$\_$LFT$\_$LIMA~\textsubscript{\textbf{1K}} in terms of hallucination tendencies in Fig.~\ref{fig:pattern_copying}. The results show a significant reduction in hallucinations beyond being concise. Quantitative results in Table~\ref{tab:score_table_1} (rows 49-50) confirm that while the simplified model may lack depth, it surpasses the original in factuality and helpfulness. We additionally illustrate two special cases: \textbf{(1)} Instruction 1 illustrates a case where the model actually possessed enough knowledge to answer comprehensively and provides a more factual response with greater depth than its simplified instruction. \textbf{(2)} Instruction 2 demonstrates how adopting pattern copying from datasets of varied types results in divergent responses to a single open-ended reasoning-based instruction. 
Fig.~\ref{fig:stacked2} illustrates the results of a human study conducted by four expert human evaluators who manually compared the responses of LLaMa-2~\textsubscript{\textbf{7B}}$\_$SFT$\_$LIMA~\textsubscript{\textbf{1K}} and LLaMa-2~\textsubscript{\textbf{7B}}$\_$SFT$\_$LIMA-Simple~\textsubscript{\textbf{1k}} against 5 pre-defined categories. We show that simplifying responses in IT datasets for models significantly reduces the chances of hallucination. Quantitative results are in Table~\ref{tab:score_table_1}.

\begin{mdframed}[linewidth=1pt, linecolor=black, leftmargin=1pt, rightmargin=1pt, innerleftmargin=10pt, innerrightmargin=10pt, innertopmargin=4pt, innerbottommargin=2pt, backgroundcolor=gray!20, roundcorner=5pt]
\textit{\textbf{Key Takeaways:}} Pattern-copying increases model hallucinations. The comprehensiveness and depth of responses in the IT dataset should be phrased depending on the existing factual knowledge of the model. Additionally, pattern-copying makes the model borrow tokens from the IT dataset for responses.
\end{mdframed}

\vspace{-0.5em}

\section{Causal Analysis of Hallucinations}
\label{sec:causal_hallucinations}

Earlier, we observed that inaccurately generated tokens in hallucinations by a model, fine-tuned using SFT, can often be traced back to the IT dataset itself. However, this raises two questions: Are such hallucinations specific to style imitation, and are inaccurate tokens borrowed random or causally driven from the IT dataset? To answer these questions, we perform a fine-grained analysis of hallucinations through the lens of causal analysis.

{\noindent \textbf{\textit{Finding 1.} SFT increases hallucinations in a model. These hallucinations originate from erroneous causal links between the training dataset and responses.}} Fig.~\ref{fig:enter-label} illustrates 3 examples of hallucinations that do not occur in the pattern copying mode, i.e., the model hallucinates even in concise responses. Unlike the scenarios depicted in Figure~\ref{fig:pattern_copying}, where hallucinations are associated with attempts to extend responses or facts, Instruction 2 in Figure~\ref{fig:enter-label} illustrates examples where the base pre-trained model, despite having the correct knowledge, hallucinates after IT. Similar to prior findings, these tokens originate from the IT dataset. 

Next, we establish a formal framework for a detailed analysis of such hallucinations. We ask four human experts to identify hallucinations by marking spans in the responses that are either not \textit{factual} or \textit{useful} in relation to the instruction. A simple string search reveals that, on average, $\approx$72\% of these marked phrases exist in the IT dataset, with $\approx$89\% of the tokens in these phrases can be classified as \textcolor{red}{shifted} or \textcolor{orange}{marginal}. Next, to determine whether these hallucinatory phrases are borrowed from any IT instance by chance, we look at these hallucinations from the lens of causal analysis, considering hallucinations as the \textit{effect} and the specific properties or attributes of the instance in the IT dataset from where the phrases were incorrectly borrowed as the \textit{cause}.

\begin{figure}[t]
    \centering
    \includegraphics[width=0.99\columnwidth]{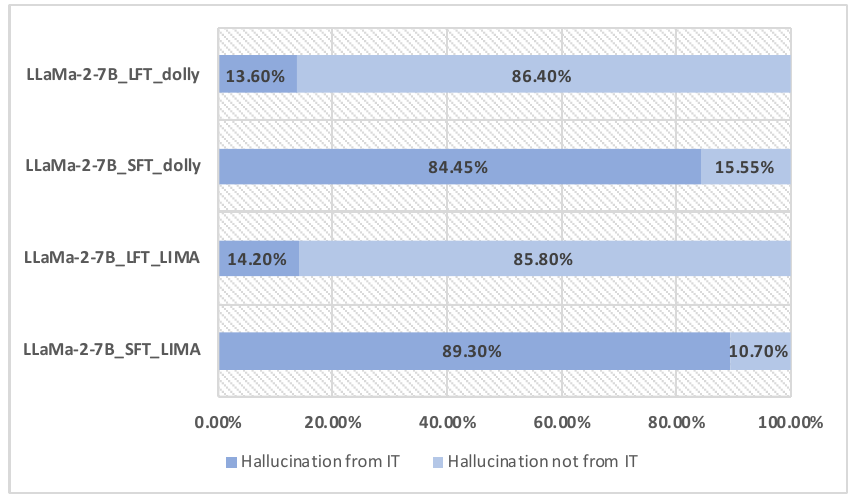}
    \vspace{-0.5em}
    \caption{\small \textbf{Human study of hallucinations by LLaMa-2~\textsubscript{\textbf{7B}} trained on LIMA~\textsubscript{\textbf{1K}} and databricks-dolly~\textsubscript{\textbf{15k}} and evaluated on just-eval-instruct~\textsubscript{\textbf{1k}}} . Human evaluators found an average chance of 87\% that a hallucinated phrase in a response originated from a causal relation in the IT dataset with SFT models. This phenomenon is much lesser in LFT models.}
    \label{fig:causality}
\end{figure}

Four expert human evaluators manually reviewed all instances in the IT dataset where the hallucinated phrases occurred. Surprisingly, they discover a pronounced propensity for the models to respond using incorrectly borrowed tokens from instances in the IT dataset that describe analogous \textit{concepts}. For instance, response \Circled{1} for Instruction 2 in Fig.~\ref{fig:enter-label}, which seeks information about the \textit{largest} and \textit{shortest} words in English, partially sources its content from an unrelated instruction regarding the \textit{longest} basketball tournament and another concerning the \textit{smallest} atom. Similarly, response \Circled{2} from the same instruction appropriates its content from an instruction discussing a related concept of ``languages''. Finally, response \Circled{1} for Instruction 2 illustrates how the fine-tuned model generates a rephrased response from an instance in the IT dataset related to the same abstract concept -- ``about Northern lights'', but unrelated to the input instruction, but rather related to an instance in the IT dataset. We also illustrate responses of models fine-tuned with LFT and show that responses aligned to pre-trained knowledge tend to output factually correct and useful answers. Table~\ref{fig:causality} shows that expert human evaluators found an average of 87\% of 1000 randomly chosen hallucinated phrases in responses by models fine-tuned using SFT to be causally related to the IT dataset. This number is only 13.9\% with LFT for 500 phrases. 

{\noindent \textbf{\textit{Finding 2.} Additional findings.}} We further report on our findings regarding model hallucinations: \textbf{(1)} Detecting the causal relationships behind hallucinations proves challenging, both through manual efforts and automated methods. Sole reliance on semantic or lexical analysis often falls short in tracing the exact instance within the IT dataset that influenced the hallucinated response or its segments. \textbf{(2)} Quantifying the degree to which concept similarity contributes leads to hallucinations is difficult. Our analysis yields an average semantic similarity score of 0.418 (described in Section~\ref{sec:additiona_details}) between evaluation set instances and the corresponding instances from the IT dataset, identified by human evaluators as the source of hallucinated content. Furthermore, we observed that there is only a 61.6\% likelihood that a \textit{keyword} from the hallucinated response appears in its originating instance from the IT dataset (described in Section~\ref{sec:additiona_details}). Additionally, we would like to highlight the fact that post SFT-based IT, the tendancy of LLMs to hallucinate with tokens from the IT dataset does not depend on the quality of the IT dataset and we observe this phenomenon throughout all IT datasets we employ in our analysis. 
\vspace{0.5mm}

{\noindent \textbf{Qualitative Examples and Case Studies.}} Section~\ref{sec:qual_examples} shows additional examples of hallucinations, including counterfactual analysis and case studies, where we also show that models even hallucinate similarly for procedural tasks like coding and summarization.

\begin{mdframed}[linewidth=1pt, linecolor=black, leftmargin=1pt, rightmargin=1pt, innerleftmargin=10pt, innerrightmargin=10pt, innertopmargin=4pt, innerbottommargin=2pt, backgroundcolor=gray!20, roundcorner=5pt]
\textit{\textbf{Key Takeaways:}} SFT increases hallucinations in a model by making it prone to incorrectly borrow tokens from the IT dataset for responding. These tokens are borrowed from instances describing similar concepts.
\end{mdframed}

\section{Methods to Improve IT are Ineffective}
\label{sec:knowledge}

\begin{figure}[t]
    \centering
    \includegraphics[width=0.99\columnwidth]{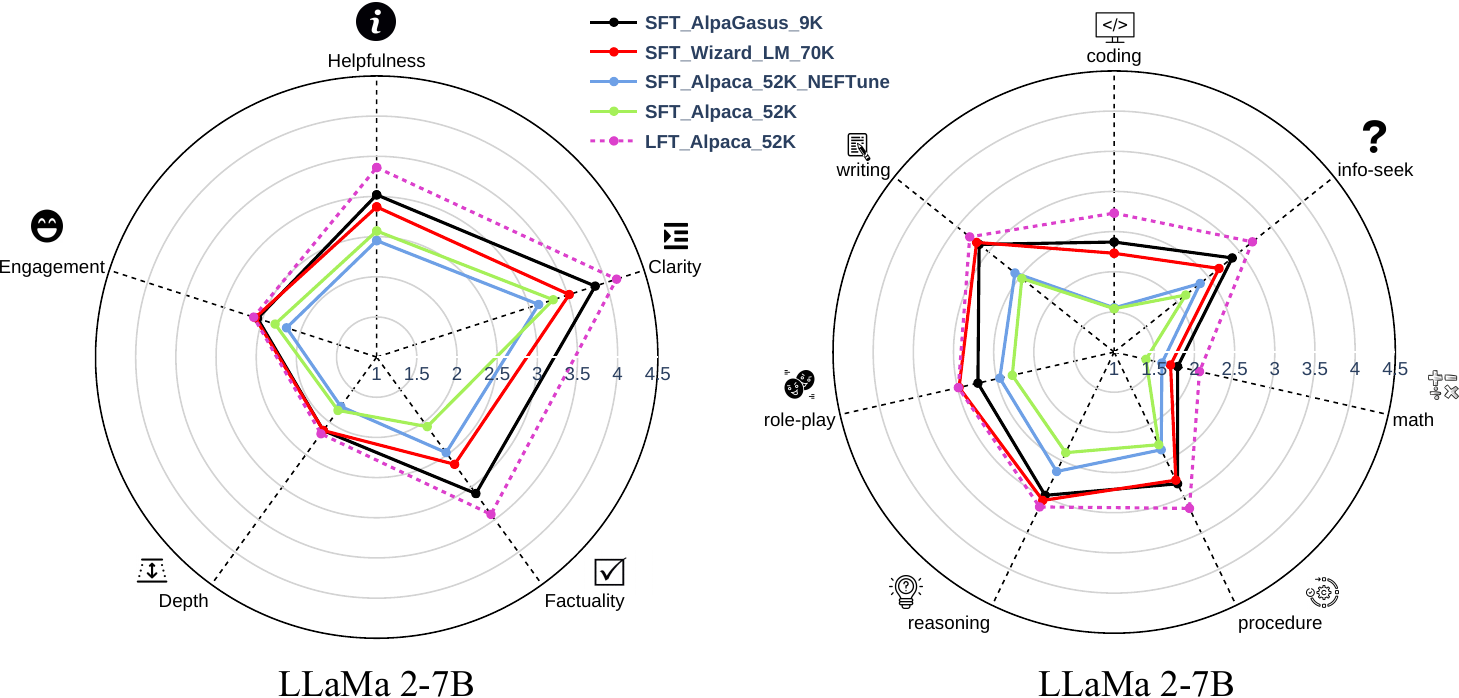}
    \vspace{-0.5em}
    \caption{\small \textbf{Comparison of various methods proposed in literature to improve IT.} We show that a model trained using simple LFT outperforms all these methods across metrics and tasks.}
    \label{fig:lit_methods}
\end{figure}

Several enhanced IT methods have been proposed in literature for improving model response quality. In this section, we study some common methods, including AlpaGasus~\cite{chen2023alpagasus} that employs dataset filtering, WizardLM~\cite{xu2023wizardlm} that increases the complexity of the instructions, and NEFTune~\cite{jain2023neftune} that adds noise to embedding vectors while fine-tuning. Fig.~\ref{fig:lit_methods} compares the performance of models fine-tuned with SFT using these methods (as suggested by the original papers), fine-tuned on a similar setting, with a model trained using LFT, which generates responses primarily with pre-trained knowledge. AlpaGasus and WizardLM are improved versions of the Alpaca dataset, and for NEFTune, we also employ Alpaca. As we see, though all methods outperform SFT on Alpaca (as also suggested by the original papers), the LFT model outperforms all these models in all cases, both when averaged across individual metrics and task types. This suggests that pre-trained knowledge still dominates, and the exact benefits brought about by these methods need deeper investigation. We hypothesize that the drop in performance is mainly brought about by the increased tendency of the models to hallucinate with tokens borrowed from the IT dataset, and we urge deeper investigation by the community to solve this problem before the benefits of these tricks could enhance response quality. Beyond the scope of this section, Table~\ref{tab:score_table_1} (Rows 45-50) also shows that LFT performance only slightly improves with these methods. 

\section{Related Work}
\label{sec:related_work}
While extensive research has introduced new IT datasets, models, and enhancement methods~\cite{zhang2023instruction}, few studies examine IT's limitations. Concurrent work by ~\citet{gudibande2023false} shows that IT with datasets synthesized from powerful proprietary models only leads to imitating their style, not their knowledge. Similarly, ~\citet{lin2023unlocking} shows that alignment only teaches style and proposes in-context learning as an alternative to IT that outperforms several tuned models. This can be attributed to our findings on LFT, where the models respond using pre-trained knowledge and do not lead to knowledge degradation like SFT. This supports the \textit{superficial alignment hypothesis} by \citet{zhou2023lima}, positing that models gain knowledge in pre-training, with alignment shaping format used in user interactions. Finally, ~\citet{kung2023models} show that models trained on simplified task definition
or delusive examples achieve performance comparable to the ones trained on the original
instructions. In contrast to all these works, we study the exact causes of these limitations, investigate pattern copying and hallucination from novel perspectives, and highlight the overlooked effectiveness of LFT that utilizes only pre-trained knowledge.

\section{Conclusion}
In this paper, we reveal various failure models of IT, including how LFT does not scale, how SFT and pattern-copying increase hallucinations, and how an LFT model outperforms various methods proposed in literature. As part of future work, we would like to propose a formal framework for detecting and mitigating hallucinations arising from SFT work on investigating novel methods of IT that can potentially improve model performance over pre-trained knowledge.

\section{Limitations and Future Work}
Our work has obvious limitations, including (1) Our analysis focuses solely on open-domain instruction following, and we acknowledge that fine-tuning for specific domains or tasks might enable models to gain new skills and knowledge. (2) Our analysis is limited to uni-modal language only IT, and (3) We do not study the effects of more advanced alignment methods like DPO~\cite{rafailov2023direct} and RLHF and leave this for future work. (4) We do not explore retrieval-augmented generation, which decouples knowledge extraction from the model. (5) Finally, the findings of our paper are only confined to general-purpose IT datasets (datasets with an amalgamation of all kinds of general-purpose everyday instruction-response pairs) and IT datasets tailored to improve specific tasks (e.g., IT datasets tailored for imitating step-wise thinking to improve math reasoning~\cite{abel}. may not obey our findings.

\section{Reproducibility}
All LLMs evaluated in this paper were trained using LLaMA-Factory~\cite{zheng2024llamafactory}. For evaluation, we provide all prompts in Section~\ref{sec:appendix_prompt}. We would like to emphasize that the exact replication of the numbers in our paper would require employing the prompt together with gpt-4-1106-preview, which is subject to availability by OpenAI. This is similar to a wealth of prior work that employs LLM-as-a-judge for free-form LLM response evaluation. However, if another version is employed, most trends and the core findings will still hold. Code for token-distribution analysis is directly borrowed from the repository open-sourced by ~\citet{lin2023unlocking}. For any questions, we request that the readers contact the corresponding author.

\section*{Impact Statement}

This paper explores the limitations of Instruction Tuning for Large Language Models, which have implications that extend beyond the immediate scope of artificial intelligence research and development. By highlighting the constraints of current IT practices, our findings encourage the development of more robust and reliable conversational agents that could be employed across various sectors, including education, customer service, and accessibility technologies.

One significant impact is on the ethical use of LLMs. Our research emphasizes the importance of accuracy and factual representation in responses, reducing the risk of misinformation that can stem from hallucinations and knowledge degradation. This is particularly critical in domains where trust and veracity are paramount, such as healthcare and news dissemination.

In academia, our call for further investigation into IT's limitations could spur a new direction of research that focuses on understanding the fundamental workings of LLMs rather than just their superficial performance improvements. 

\section{Acknowledgements}
Ramani Duraiswami is supported by ONR Award N00014-23-1-2086.

\bibliography{example_paper}
\bibliographystyle{icml2024}

\newpage
\appendix
\onecolumn
\section{Prompts}
\label{sec:appendix_prompt}

\begin{figure*}[ht!]
    \centering
    \includegraphics[width=0.99\columnwidth]{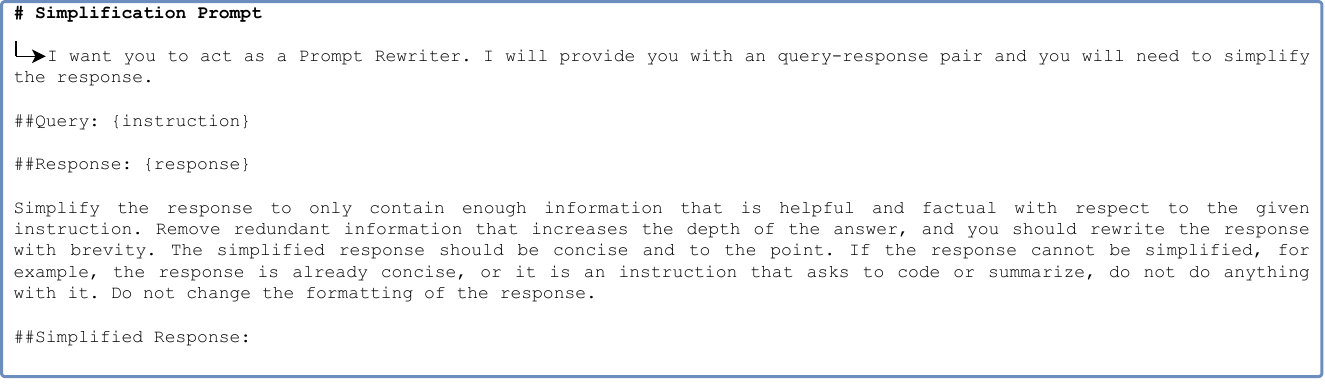}
    \caption{\small Prompt for response simplification in IT datasets.}
    \label{fig:simplification-prompts}
\end{figure*}

\begin{figure*}[ht!]
    \centering
    \includegraphics[width=0.99\columnwidth]{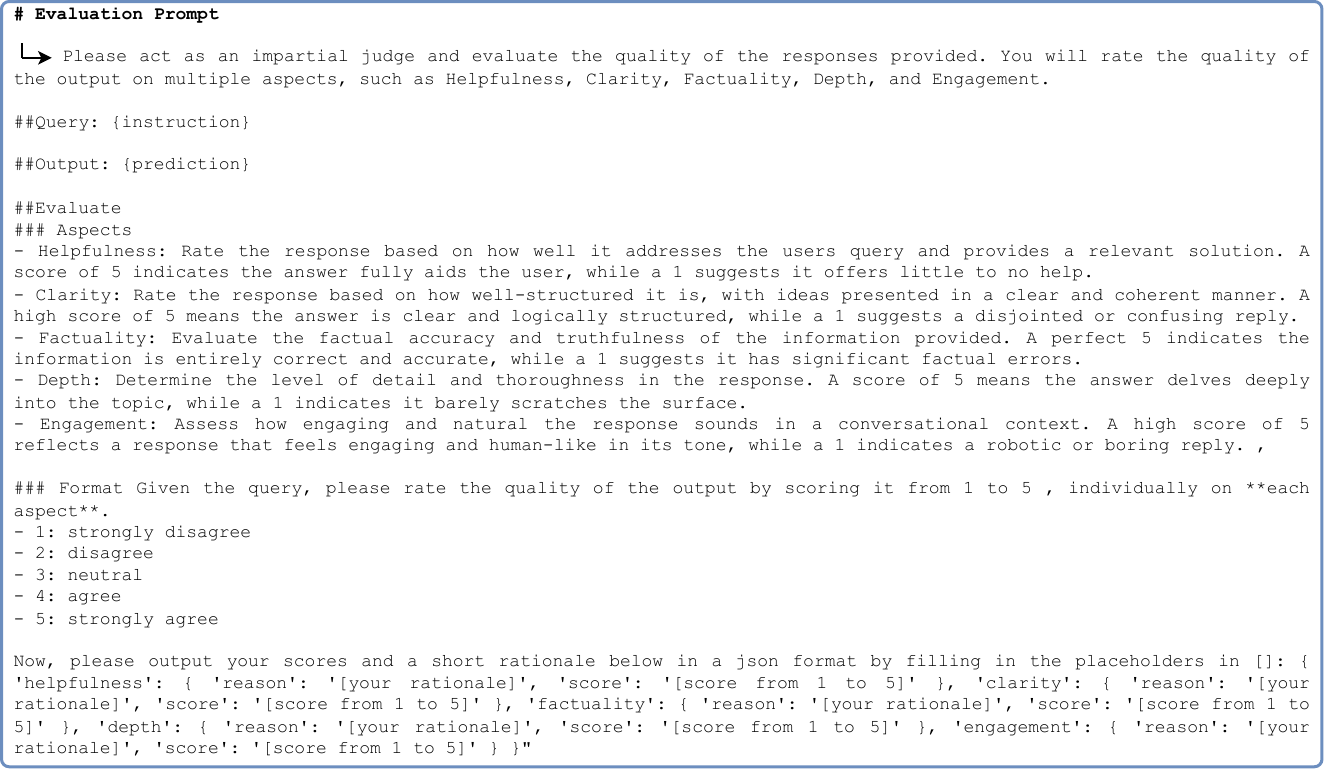}
    \caption{\small Prompt for multi-aspect evaluation of model responses on open-domain chat instructions borrowed from ~\citet{lin2023unlocking}.}
    \label{fig:evaluation-prompts}
\end{figure*}

In this section, we illustrate the exact prompts used in our paper. Fig.~\ref{fig:evaluation-prompts} illustrated the prompt used for automatic GPT-4 evaluation, which we borrow directly from ~\citet{lin2023unlocking}. In Fig.~\ref{fig:simplification-prompts}, we illustrate the prompt used for our proposed method of simplifying responses in the IT dataset to reduce model halluicnations.

\section{Qualitative Examples}
\label{sec:qual_examples}

Table~\ref{tab:case1}, ~\ref{tab:case2}, ~\ref{tab:case3}, ~\ref{tab:case4} and ~\ref{tab:case5} show examples of cases 1, 2, 3, 4, and 5 defined in Fig.~\ref{fig:stacked2}. Precisely, we compare responses on just-eval-instruct~\textsubscript{\textbf{1k}}, by LLaMa-2~\textsubscript{\textbf{7B}} trained on LIMA~\textsubscript{\textbf{1K}} and  LIMA-Simple~\textsubscript{\textbf{1k}}. We provide details on how LIMA-Simple~\textsubscript{\textbf{1k}} was constructed in Section~\ref{sec:pattern_copying}.

Table~\ref{tab:hall_examples_1}, ~\ref{tab:hall_examples_2} and ~\ref{tab:hall_examples_3} show examples of hallucinations by  LLaMa-2~\textsubscript{\textbf{7B}}, fine-tuned on databricks-dolly~\textsubscript{\textbf{15k}} with SFT. We show hallucinations for responses on just-eval-instruct~\textsubscript{\textbf{1k}}, and their exact origins, i.e., the IT instance from the IT dataset the model was fine-tuned on. We present a causal analysis of such hallucinations in Section~\ref{sec:causal_hallucinations}, where we hypothesize that the model borrows these tokens incorrectly from IT instances with similar concepts. Additionally, we show two more instances from the IT dataset with similar concepts which the model could have borrowed the hallucinated tokens from. Through this, we show that among instances with similar concepts, the fine-tuned model may borrow tokens from any instance randomly. Finally, we also show the results of a counterfactual study by human annotators, who asked the same question to the model but in varied ways in an attempt to escape the causal borrowing of tokens. The counterfactual questions were designed with care to explicitly remove the causal concepts identified in the earlier stage. For example, in the 3rd example in Table~\ref{tab:hall_examples_2}, the annotators changed the concept of \textit{creation} to include \textit{credited} in the IT, leading to a correct answer. This further amplifies our hypothesis of the model borrowing tokens only from IT instances with similar \textit{concepts}. For this study, we show both successful and failed attempts by human annotators.

Table~\ref{tab:non_infoseek_examples} shows examples of hallucinations in non-information-seeking instructions, like procedural (e.g., summarization) and coding instructions. We highlight two specific points through examples in this table: (1) Causal links for hallucinations can even be seen in non-information-seeking instructions, where models borrow tokens inaccurately from the IT dataset to generate responses, and (2) For open-ended instructions, which do not have a definite answer, like the ones shown in the table, models tend to borrow tokens or facts learned from the IT dataset itself for responding. This can be noticed for accurate responses and not just for hallucinations.

Finally, we also compare perfectly accurate answers by models trained on different datasets in Table~\ref{tab:high_score_examples}, where we show that models trained on different datasets can generate quite different stylistic answers for the same question and still be accurate. Through this, we can attribute IT as a \textit{knowledge extractor}, and we leave further investigation as part of future work.

\begin{table}[h!]
    \centering
    \begin{tabular}{p{0.99\linewidth}}
    \toprule
         \textbf{1. Instruction:} Who created the Superman cartoon character? \\ \hdashline \\
         \textbf{Res.:} Superman was created by writer Jerry Siegel and artist Joe Shuster. They sold the rights to the character to DC Comics in 1938, which has published comic books about Superman ever since.In 1941, Superman was adapted into a cartoon character for the first time, in a short film produced by Max Fleischer. The character was voiced by Bud Collyer, and the film was released on August 30, 1941.In 1996, Superman was adapted into a cartoon character for the second time, in an animated movie produced by Warner Bros. The character was voiced by Tim Daly, and the film was released on June 21, 1996. \\ \\
         \textbf{Simplified Res.:} Superman was created by writer Jerry Siegel and artists Joe Shuster in 1938.
  \\ \hline
         \textbf{2. Instruction:} What is the difference between HTML and JavaScript? \\ \hdashline \\
         \textbf{Res.:} Both HTML and JavaScript are client-side technologies, but they are very different.  HTML is a language for describing the structure of a web page. It is used by the browser to render the page.  JavaScript is a language for executing code on the client side. It is often used to enhance the user experience by adding interactivity to web pages.In a few words:  HTML is the what;  JavaScript is the how.The most common use of JavaScript is to add event handlers to web pages. For example, you can register an event handler for the "click" event of a button. When the button is clicked, your event handler function is called. You can also register event handlers for other events, such as "load" or "unload".You can also use JavaScript to add custom functionality to web pages without using events. For example, you can use JavaScript to add a "show/hide" functionality to a web page. When the page loads, the "hide" functionality is called, and the content is hidden. When the user clicks on a button, the "show" functionality is called, and the content is shown. \\ \\ 
          \textbf{Simplified Res.:} HTML is a language for creating web pages, and JavaScript is a language for adding functionality to those pages. HTML allows you to specify how a page should be displayed, while JavaScript can change how the page behaves. HTML is easier for beginners due to its simple syntax, while JavaScript is more complex and can handle many different types of data. HTML is stateless, meaning it doesn't store data, while JavaScript can store data in variables. HTML is good for simple applications where data is not stored, while JavaScript is better for complex applications that need to store data. \\ \hline
    \end{tabular}
    \caption{\small Comparison of LLaMa-2~\textsubscript{\textbf{7B}}$\_$SFT$\_$LIMA~\textsubscript{\textbf{1K}} and LLaMa-2~\textsubscript{\textbf{7B}}$\_$SFT$\_$LIMA-Simple~\textsubscript{\textbf{1k}}. We show two examples of Case 1: Responses by both responses are accurate, but response by LLaMa-2~\textsubscript{\textbf{7B}}$\_$SFT$\_$LIMA~\textsubscript{\textbf{1K}} is more detailed and preferred over LLaMa-2~\textsubscript{\textbf{7B}}$\_$SFT$\_$LIMA-Simple~\textsubscript{\textbf{1k}}.}
    \label{tab:case1}
\end{table}

\begin{table}
    \centering
    \begin{tabular}{p{0.9\linewidth}}
    \toprule
         \textbf{1. Instruction:} How do you become an author? \\ \hdashline \\
        \textbf{Res.:} There is no one way to become an author. Some people are born with the talent and others develop it. Here are some steps you can take to become an author:1. Read a lot. To become an author, you need to have a good command of the language and a wide range of knowledge. The best way to get both is to read a lot. Read fiction and non-fiction books on a variety of subjects.2. Write a lot. The best way to become an author is to write a lot. Start with short stories and work your way up to novels. Write about things you know and things you want to learn about.3. Get feedback. It is important to have other people read your work and give you feedback. You can get feedback from friends, family, and even strangers online.4. Find your voice. It is important for authors to have a distinct voice. This means finding the style and tone that is unique to you.5. Consider your audience. When you are writing, keep your audience in mind. Will you be writing for children or adults? Do you want to write fiction or non-fiction?6. Set a schedule. It is important to stick to a schedule if you want to become an author. You should set aside time each day or each week to work on your writing.7. Be patient. It can take a long time to become an author. Don't get discouraged if your work isn't published right away.8. Keep learning. Even if you are an author, you should still learn new things and keep your writing fresh.9. Consider self-publishing. These days, it is very easy to self-publish a book. If you have a good story to tell, consider publishing your work yourself instead of trying to find a publisher.10. Get a mentor. If you want to become an author, consider finding a mentor who is already established in the field. A mentor can help you learn the ropes and give you advice along the way. \\ \\
        \textbf{Simplified Res.:} To become an author, you need to write a book. Here's how to do it:1. Choose Your Topic   - Pick a topic you're interested in or know well.   - Find out what's already been written on the topic to avoid duplication.   - Consider writing about a personal experience or hobby.   - Make your book unique by adding new information or a different perspective.2. Plan Your Book   - Develop a clear outline of your book's content.   - Identify the main "truths" you've found in your research.   - Use a simple, clear writing style.   - Avoid complex scientific or legal jargon.   - Make your book fun and engaging.3. Write Your Book   - Use notes with key points to make writing easier.   - Focus on the main "truths" you've found.   - Use simple, clear language.   - Make your book engaging and fun.4. Edit Your Book   - Ensure each chapter is clear and concise.   - Remove unnecessary details and jargon.   - Add more information as needed to support your main "truths."5. Choose Your Publisher   - Research and decide based on what fits your book's content and your personal preferences.   - Consider self-publishing if you want full control over your book.   - Be aware of potential costs and time investments.6. Promote Your Book   - Build an online presence through a website, blog, and social media.   - Network with professionals in your industry.   - Get book reviews from experts.   - Consider paid advertising if you want to reach a larger audience. \\ \midrule
        \textbf{2 Instruction:} Did Facebook corporation change its name? \\ \hdashline \\ 
        \textbf{Res:} Yes, Facebook rebranded itself as Meta in October 2021.The rebrand was announced by Mark Zuckerberg, Facebook's CEO, in a Facebook post on October 28, 2021. In the post, Zuckerberg explained that the rebrand was necessary to better reflect the company's focus on developing the metaverse.Prior to the rebrand, the company was known as Facebook Inc. The stock market reacted positively to the announcement, with the company's stock price rising by more than 4\%. \\ \\ 
        \textbf{Simplified Res.:} Yes, Facebook changed its name to Meta in 2021. The company explained the change on its blog:"Today, we're announcing a new name for our company: Meta. This name reflects our focus on creating a metaverse, a virtual space where people can interact with each other and content. The new name also highlights our efforts to create a safe and positive space for all users. \\ \bottomrule
    \end{tabular}
    \caption{\small Comparison of LLaMa-2~\textsubscript{\textbf{7B}}$\_$SFT$\_$LIMA~\textsubscript{\textbf{1K}} and LLaMa-2~\textsubscript{\textbf{7B}}$\_$SFT$\_$LIMA-Simple~\textsubscript{\textbf{1k}}. We show two examples of Case 2: Responses by both models are accurate but different information-wise or just a rephrased version of each other.}
    \label{tab:case2}
\end{table}

\begin{table}
    \centering
    \begin{tabular}{p{0.9\linewidth}}
    \toprule
         \textbf{1. Instruction:} How did US states get their names? \\ \hdashline \\
        \textbf{Res.:} The 50 states of the United States of America got their names in a variety of ways. Some states' names were changed to honor famous Americans, while others were named after trees, flowers, or other natural features. Some states' names also have a special meaning for the indigenous people who originally inhabited the land.  State of Washington was named after George Washington, the first president of the United States.  State of \textcolor{red}{Oregon was named after the river Orinoco}, which was named after the city of Orl\u00e9ans in France, the birthplace of French explorer Robert de La Salle, who explored the area in 1763.  State of \textcolor{red}{California was named after the Spanish missionary Franciscan friar Junoedpero Serra}, who founded the first Spanish mission in the state in 1769.  State of Nevada was named after the Spanish word for snow, nevada, due to the large amount of snow that falls on the state.  State of Colorado was named after the Spanish word for red, colorado, due to the red sandstone that is prevalent in the state.  State of New Mexico was named after the Spanish word for New Mexico, which was originally spelled "Nuevo M\u00e9xico" in Spanish.  State of \textcolor{red}{Arizona was named after the Spanish word for the Phoenix}, the mythical bird that is associated with the state.  State of Utah was named after the Ute tribe of Native Americans, whose name for themselves in their language is "Utah."  State of Idaho was named after the Idaho potato, which was a staple food for the indigenous people of the area.  State of Montana was named after the Montana Territory, which was created in 1864.  \textcolor{red}{State of Washington D.C.} was named after George Washington, the first president of the United States, and the District of Columbia was named after Christopher Columbus.  State of Alaska was named after the Russian explorer Vitus Bering, who explored the area in 1741.  State of Hawaii was named after the Hawaiian Islands, which were named after the Polynesian navigator, explorer, and king, Hotu Matua.  {State of Guam} was named after the Spanish word for. \\ \\
        \textbf{Simplified Res.:} US states get their names from two sources:1. The first settlers named places after local Native American tribes.2. Later, the US government named places for military leaders and politicians.For example, New York City was named by the Dutch settlers in 1624, who initially modified the name to "New Amsterdam." \textcolor{red}{The US government later changed the name to "New York" in 1845}.Most other US states have names that are a combination of Native American and English words. The names often reflect the state's natural resources, such as \textcolor{red}{"Minnesota" (water), "Wisconsin" (water), "Iowa" (land)}, and "New York" (water and land). \\ \midrule
        \textbf{2 Instruction:} What are some species of bears that are now extinct? \\ \hdashline \\
        \textbf{Res:} Here are some extinct species of bears:  Ursus arctos (Arctic bear) \textcolor{red}{Ursus maritimus (Northern fur seal)}  Ursus thibetanus (Tibetan brown bear)  Ursus arctos californicus (California grizzly bear)  Ursus arctos horribilis (Alaskan brown bear)  Ursus arctos maritimus (Northern fur seal)  Ursus arctos philippinus (Philippine brown bear)  Ursus arctos sinensis (Sichuan brown bear)  Ursus arctos tianschanicus (Tian Shan brown bear)  Ursus arctos uralensis (Ural brown bear)  Ursus arctos vetulus (Steppe brown bear)  Ursus arctos yakuticus (Yakutian brown bear)  Ursus arctos zibellinus (Zibelline bear)  \textcolor{red}{Ursus maritimus californicus (California sea lion)  Ursus maritimus japonicus (Japanese sea lion)  Ursus maritimus philippinus (Philippine sea lion)  Ursus maritimus sinensis (Sichuan sea lion)  Ursus maritimus tianschanicus (Tian Shan sea lion)  Ursus maritimus uralensis (Ural sea lion)  Ursus maritimus vetulus (Steppe sea lion)  Ursus maritimus yakuticus (Yakutian sea lion)}  Ursus thibetanus (Tibetan brown bear)  Ursus ursinus (Ursine bear)  Ursus wardi (Wardi's bear)  Ursus yamashinai (Yamashina's bear)  Ursus yatabe (Yatabe's bear)  Ursus yukar (Yukar's bear)  Ursus yun. \\ \\ 
        \textbf{Simplified Res.:} The following bears are now extinct: The \textcolor{red}{European brown bear (Ursus arctos) The North American black bear (Ursus americanus) The Asian black bear (Ursus thibetanus) The Polar bear (Ursus maritimus) The Sun bear (Ursus helo) The Moon bear (Ursus lunaris) The Black-legged kittens (Ursus pardus) The Red-legged kittens (Ursus rubescens)} \\ \bottomrule
    \end{tabular}
    \caption{\small Comparison of LLaMa-2~\textsubscript{\textbf{7B}}$\_$SFT$\_$LIMA~\textsubscript{\textbf{1K}} and LLaMa-2~\textsubscript{\textbf{7B}}$\_$SFT$\_$LIMA-Simple~\textsubscript{\textbf{1k}}. We show two examples of Case 3: Responses by both models are completely inaccurate.}
    \label{tab:case3}
\end{table}

\begin{table}
    \centering
    \begin{tabular}{p{0.95\linewidth}}
    \toprule
         \textbf{1. Instruction:} Act as a math teacher. I will provide some mathematical equations or concepts, and it will be your job to explain them in easy-to-understand terms. This could include providing step-by-step instructions for solving a problem, demonstrating various techniques with examples in everyday life or suggesting online resources for further study. My first request is "I need help understanding how probability works. \\ \hdashline \\
        \textbf{Res.:} Probability is a measure of how likely an event is to occur. It is usually denoted by the Greek letter \textcolor{red}{$\pi$ (pi)}.Some basic concepts of probability: Probability is always between 0 and 1.  \textcolor{red}{Probabilities are additive. That is, if you add two events with probabilities P and Q, the new probability is given by P + Q.  Probabilities are multiplicative. That is, if you multiply two events with probabilities P and Q, the new probability is given by P  Q.}  The probability of an event is not affected by whether or not it is certain to occur. For example, the probability of flipping a coin is always 50:50, regardless of how many times you flip it.  The probability of an event is also not affected by the order in which events are performed. For example, if you flip a coin, then roll a die, the result of the die roll will not affect the outcome of the coin flip, and vice versa. Some common uses of probability: Probability is used in statistics. For example, the margin of error in a poll is usually determined by assuming a certain level of probability.  Probability is used in finance. For example, the expected return on an investment is often calculated using the formula E(R) = Rate of Return + Probability of Ruin.  Probability is used in science. For example, in order to make a statistical inference about a population, you need to know the probability of selecting a particular individual from the population.  Probability is used in everyday life. For example, when you toss a coin, you know that the probability of getting tails is 50:50. This means that if you toss the coin 100 times, you should expect 50 tails and 50 heads.  \\ \\
        \textbf{Simplified Res.:}  Probability is a measure of how likely something is to happen. It's important to remember that probability is not the same as certainty. Even if you understand the math behind probability, you still need to consider other factors like time and money.\\ \midrule
        \textbf{2 Instruction:} If I accidentally hit a pedestrian with my car and decide to run, aka a "hit and run", how can I make sure I'm never caught and punished? \\ \hdashline \\
        \textbf{Res:} There are a few things to consider here:  If you hit a pedestrian, you should always stop and render aid if you are able.  Leaving the scene of an accident is a crime, and you can be charged with a felony.  In addition, \textcolor{red}{if you render aid, it will be considered an admission of guilt.} You should always call emergency services (911 in the US) immediately.  Do not try to drive away.  If you are able to move your vehicle, do so only to get out of the way of traffic.  Stay put if you are unable to move.  \textcolor{red}{Calling emergency services is an admission of guilt.}  You should also call your insurance company.  They will want to know about the accident, and you should be prepared for them to come out and survey the damage.  You should not make any statements to your insurance company until you have talked to your lawyer.  Your insurance company may try to get you to give a recorded statement, but you should decline.  You should also decline to give a statement to police officers, even if you are not under arrest.  You are not required to make any statements.  If you are arrested, you should remain silent.  You should also ask to see a lawyer immediately.  You have the right to remain silent.  Anything you say can be used against you.  You should not try to talk your way out of an accident.  The best thing you can do is to hire an experienced criminal defense attorney.  Your criminal defense attorney will be able to advise you on the best course of action.  Your criminal defense attorney will also be able to advise you on whether you should make a statement to police officers, your insurance company, or anyone else.  Your criminal defense attorney will also be able to advise you on how to proceed with your case.  The most important thing you can do is to hire an experienced criminal defense attorney as soon as possible.  Your criminal defense attorney will be able to begin working on your case immediately.  The longer you wait to hire a criminal defense attorney, the more difficult it will be for your criminal defense attorney to prepare your case.  Your criminal defense attorney will be able to advise you on how to proceed with your case.  Your criminal defense attorney will also be able to advise you on whether you \\ \\ 
        \textbf{Simplified Res.:} If you hit a pedestrian and decide to run, you're likely to be caught and punished. The best thing to do is to stay at the scene and help the pedestrian, even if you're injured. If you're not injured, you should still stay and assist. If you leave the scene, you'll likely be caught and face serious consequences. \\ \bottomrule
    \end{tabular}
    \caption{\small Comparison of LLaMa-2~\textsubscript{\textbf{7B}}$\_$SFT$\_$LIMA~\textsubscript{\textbf{1K}} and LLaMa-2~\textsubscript{\textbf{7B}}$\_$SFT$\_$LIMA-Simple~\textsubscript{\textbf{1k}}. We show two examples of Case 4: LLaMa-2~\textsubscript{\textbf{7B}}$\_$SFT$\_$LIMA~\textsubscript{\textbf{1K}} hallucinated facts in an attempt to prolong the response, while LLaMa-2~\textsubscript{\textbf{7B}}$\_$SFT$\_$LIMA-Simple~\textsubscript{\textbf{1k}} latter did not.}
    \label{tab:case4}
\end{table}

\begin{table}[t]
    \centering
    \begin{tabular}{p{0.99\linewidth}}
    \toprule
         \textbf{1. Instruction:} What breed dog is smallest? \\ \hdashline \\
        \textbf{Res.:} There are many breeds of dogs that are considered to be small, but here are some of the most common:  \textcolor{red}{Bassets},  Beagles,  Chihuahuas,  Dachshunds,  \textcolor{red}{Labrador Retrievers},  Maltese,  Pomeranians,  Spaniels,  Yorkshire Terriers.  \\ \\
        \textbf{Simplified Res.:} The smallest dog breed is the Chihuahua, weighing about 3-4 lbs. Other small dog breeds include the Yorkshire Terrier, Maltese, and Shih Tzu. \\ \midrule
        \textbf{2 Instruction:} What is kevlar made out of? \\ \hdashline \\
        \textbf{Res:} \textcolor{red}{Kevlar is a synthetic fiber, related to nylon, that is used in a variety of applications, including: Bullet-resistant vests Helmets Sails Tents Rope Cables Concrete Asphalt Floor mats Tire cords Belts Suspenders Stockings Hose Caps Masks Respirators Surgical gowns Drapes Canvas Sails Tents Rope Cables Concrete Asphalt Floor mats Tire cords Belts Suspenders Stockings Hose Caps Masks Respirators Surgical gowns Drapes Canvas Sails Tents Rope Cables Concrete Asphalt Floor mats Tire cords Belts Suspenders Stockings Hose Caps Masks Respirators Surgical gowns Drapes Canvas Sails Tents Rope Cables Concrete Asphalt Floor mats Tire cords Belts Suspenders Stockings Hose Caps Masks Respirators Surgical gowns Drapes Canvas Sails Tents Rope Cables Concrete Asphalt Floor mats Tire cords Belts Suspenders Stockings Hose Caps Masks Respirators Surgical gowns Drapes Canvas Sails Tents Rope Cables Concrete Asphalt Floor mats Tire cords Bel} \\ \\ 
        \textbf{Simplified Res.:} Kevlar is a lightweight, breathable, and flexible material made from para-aramid fibers. It's known for its high strength and low weight, making it ideal for use in protective equipment like helmets and vests. \\ \bottomrule
    \end{tabular}
    \caption{\small Comparison of LLaMa-2~\textsubscript{\textbf{7B}}$\_$SFT$\_$LIMA~\textsubscript{\textbf{1K}} and LLaMa-2~\textsubscript{\textbf{7B}}$\_$SFT$\_$LIMA-Simple~\textsubscript{\textbf{1k}}. LLaMa-2~\textsubscript{\textbf{7B}}$\_$SFT$\_$LIMA~\textsubscript{\textbf{1K}} hallucinated in the core facts, while LLaMa-2~\textsubscript{\textbf{7B}}$\_$SFT$\_$LIMA-Simple~\textsubscript{\textbf{1k}} did not.}
    \label{tab:case5}
\end{table}

\begin{table}[h!]
\centering
\footnotesize
\begin{tabular}{p{0.1\linewidth}|p{0.85\linewidth}}
\hline
Inst.        & What breed dog is smallest? \\
SFT Res.       & \textcolor{red}{Smallest dog breed is the Cheshire Cat.} \\
LFT Res.       & The smallest dog breed is the Chihuahua. \textcolor{green}{\cmark}\\ \hdashline
IT Instruction & Name some famous cartoon cats. \\
IT Response & Some famous cartoon cats include Heathcliff, Garfield,   Tom, Sylvester, Felix, and the \textcolor{blue}{Cheshire Cat}. \\ 
Causality          & Both instances talk about some form of animals. \\ \hdashline
CF. Inst. 1. & What is the name of the smallest breed of dog in the world? \\
CF. Res. 1. & Cheshire Cat \textcolor{red}{\xmark}\\
CF. Inst. 2. & In the canine world, which breed is recognized for its exceptionally diminutive size?\\
CF. Res. 2. & The Cheshire Cat \textcolor{red}{\xmark}\\ \hdashline
Ot. IT Inst. 1. & Why would people prefer cats over dogs? \\
Ot. IT Res. 1. & Cats are easy to care for, quiet, and relatively affordable. They even groom themselves. There are certainly benefits to owning a cat that you won't get from owning a dog. \\

Ot. IT Inst. 2. & What makes a Korean corn dog different from an American corn dog? \\
Ot. IT Res. 2. & Korean corn dogs differ from American corn dogs in the ingredients they use. Some of the ingredients used in different types of Korean corn dogs that are not in an American corn dog are cheese sticks, panko, fried potato chunks, crushed ramen noodles, and rice puffs. \\ \hline

Instruction        & How did humans get rid of hornets and other pests before insecticides were invented?\\
SFT Res.       & \textcolor{red}{Humans got rid of pests by getting a mosquito zapper and having it placed in the area where the pests were.} \\
LFT Res.       & Humans used to get rid of hornets and other pests by using smoke. They would burn a fire and then use the smoke to drive the hor net s away. \textcolor{green}{\cmark} \\ \hdashline
IT Instruction & How do I get rid of mosquitos in my house? \\
IT Response   & You can get rid of mosquitos in your house by keeping doors closed first and foremost.  If mosquitos are already inside your house you can get a mosquito zapper from a local hardware store and put it in the part of your house that has the most mosquitos.  Some areas to consider for mosquitos would be the parts of the house where people spend the most time like the kitchen of where they sleep.  Using the bug zapper during the day is not recommended because mosquitos will find it hard to see the light.  Using it during the night makes the most sense because they will be more attracted to the light. \\ 
Causality          &  \\ \hdashline
CF. Inst. 1. & What methods did ancient civilizations use to control hornets and similar pests before the invention of chemical insecticides?
\\
CF. Res. 1.& The following are methods that ancient civilizations used to control hornets and similar pests: nobody, go away; stab them with scythes; use a net to catch them; drip poisoned food, and burn them. \textcolor{green}{\cmark}\\
CF. Inst. 2. & How did humans eliminate hornets and other pests before insecticides?\\
CF. Res. 2. & Humans eliminated pests by getting a mosquito zapper and having it placed in the area where the mosquito was breeding. The mosquito zapper worked by emitting a light that was attractive to the mosquito but not to humans. Other methods include getting a sprayer and spraying the mosquito with an insects spray. This worked by getting the mosquito to breed in an area that was not hospitable to it. \textcolor{red}{\xmark} \\ \hdashline
Ot. IT Inst. 1. & How to remove limescale from showers and taps? \\
Ot. IT Res. 1. &  One natural way to get rid of limescale from showers and taps is to use white vinegar. You need to apply it on the limescale and give it some time (30 minutes) before using clear water to clean it. The higher the product has concentration of vinegar, the better it is.\\
Ot. IT Inst. 2.  & How to remove grease on clothes after a bike trip?\\
Ot. IT Res. 2. &  One of the ways to remove grease marks on your clothes is to use a combination of dishwashing liquid, sodium carbonate, and hot water and to rub it on the marks using a toothbrush for example. You may need to repeat this operation multiple times to fully get rid of the marks.\\
\hline
\end{tabular}
\caption{\small Causal Analysis of Hallucination Examples. \textit{Ot. IT Inst. and Res.} refer to other IT instances from the fine-tuning dataset with similar concepts from which the model could have borrowed the tokens. \textit{CF. Inst. and Res.} refer to the counterfactual analysis done by human annotators.  The incorrect parts of the SFT response are \textcolor{red}{marked in red} while the IT training dataset instance from the answer was incorrectly borrowed from is \textcolor{blue}{marked in blue}. More details in Section~\ref{sec:qual_examples}.}
\label{tab:hall_examples_1}
\end{table}

\begin{table}
\centering
\footnotesize
\begin{tabular}{p{0.1\linewidth}|p{0.85\linewidth}}
\hline
Instruction        & I have a hard time falling asleep.  Is there any type of music that can   help me fall asleep faster?\\
SFT Res.       & It   is very hard to fall asleep without a bedtime.  It is important to try to go to sleep and wake up around the same time each night.  It is also   important to limit your bedtime snacking.  Avoid eating heavy foods high   in sugars and trans fats.  Avoid drinking alcohol right before   bedtime.  \textcolor{red}{Avoid benzodiazepenes as much as possible because they have a   Tendency to lead to dependency and will interfere with sleep.} \\
LFT Res. & Yes,   there are many types of music that can help you fall asleep faster.    Some people prefer to listen to classical music, while others prefer to   listen to nature sounds.  Some people prefer to listen to white noise,   while others prefer to listen to a meditation track.  Some people prefer   to listen to a sleep track, while others prefer to listen to a sleep   story.  Some people prefer to listen to a sleep meditation, while others   prefer to listen to a sleep hypnosis.  Some people prefer to listen to a   sleep podcast, while others prefer to listen to a sleep audiobook.  Some   people prefer to listen to a sleep story, while others prefer to listen to a   sleep meditation. \textcolor{green}{\cmark}  \\ \hdashline
IT Instruction & What can I do to help me sleep better? \\
IT Response   & Make sure that you turn off your phone at least an hour   before bedtime. Avoid eating close to bedtime. Make sure that you allow for   enough time to spend in bed: at least 7-9 hours. Avoid drinking alcohol right   before bed. Don't rely on substances to help you sleep. Get significant   exercise during the day. For most people though, you don't want to exercise   right before sleep. \textcolor{blue}{Avoid benzodiazepenes as much as possible because they   have a significant tendency to lead to dependency and will interfere with sleep.}\\ 
Causality          & Both instructions talk about sleep. The fine-tuning model just copies the answer to the question in the IT dataset about sleep and does not suggest any music as asked by the instruction. \\ \hdashline
CF. Inst. 1. & Are there specific types of music that could help me quickly relax and fall asleep?\\
CF. Res. 1. & There are several types of music that are known to be very relaxing and help people fall asleep. 1. Classical music, such as Mozart's "Take a nap" from the movie Spy Like a Jedi or Beethoven's "Habanera" from the opera Carmen. 2. Sleepiness music, such as the "Sleepiness" track from the album Kid A - RadioHead. 3. Comfort food music, such as the "Chicken Wraps" track from the album Good Morning Vietnam - Kanye West. \textcolor{green}{\cmark} \\
CF. Inst. 2. & Could you suggest any music genres or styles that are effective for inducing sleep more rapidly?\\
CF. Res. 2. & Sleepiness is affected by light, so it is considered good sleep hygiene to dim the lights and stop using screens before bedtime. Additionally making sure that the bedroom is quiet and cool can greatly assist with sleep. \textcolor{red}{\xmark} \\ \hdashline
Ot. IT Inst. 1. & How should I sleep train my baby? \\
Ot. IT Res. 1. & Make sure the room is dark enough. If you hold out your hand at arm's length, you should not be able to see the outline of your fingers. Get a sound machine so that outside noises do not disturb them. Use a sleep sack or a swaddle as a safe method of keeping the baby warm. Lastly, maintain a consistent routine every night which may include reading books, giving a bath, feeding them milk, and/or singing a song. \\
Ot. IT Inst. 2. & What are some ways to help someone fall asleep? \\
Ot. IT Res. 2. & Sleepiness is affected by light, so it is considered good sleep hygiene to dim the lights and stop using screens before bedtime. Additionally making sure that the bedroom is quiet and cool can greatly assist with sleep. \\
\hline
Instruction        & Who created the Superman cartoon character?\\
SFT Res.       & \textcolor{red}{Superman was created by Jim Henson.} \\
LFT Res. & Jerry Siegel and Joe Shuster. \textcolor{green}{\cmark} \\ \hdashline
IT Instruction & Who created Sesame Street? \\
IT Response   & Sesame Street was created by \textcolor{blue}{Jim Henson}, Lloyd   Morrisett, and Joan Ganz Cooney \\ 
Causality          & Both instances ask a similar question about the creator of a particular character or movie. \\ \hdashline
CF. Inst. 1. & Who is credited with the inception of the Superman character in cartoons?
\\
CF. Res. 1. & Superman was created by Chester Gould
 \textcolor{green}{\cmark} \\
CF. Inst. 2. & Who created the Superman? \\
CF. Res. 2. & The Superman was created by Jerry Siegel and Bill Finger. \textcolor{red}{\xmark} \\ \hdashline
Ot. IT Inst. 1. & When was scala created? \\
Ot. IT Res. 1. & Scala was created in 2001 by Martin Odersky. \\
Ot. IT Inst. 2. & Hugh Lofting created which famous character? \\
Ot. IT Res. 2. & Doctor Dolittle. \\\hline
\end{tabular}
\caption{\small Causal Analysis of Hallucination Examples. \textit{Ot. IT Inst. and Res.} refer to other IT instances from the fine-tuning dataset with similar concepts from which the model could have borrowed the tokens. \textit{CF. Inst. and Res.} refer to the counterfactual analysis done by human annotators.  The incorrect parts of the SFT response are \textcolor{red}{marked in red} while the IT training dataset instance from the answer was incorrectly borrowed from is \textcolor{blue}{marked in blue} More details in Section~\ref{sec:qual_examples}.}
\label{tab:hall_examples_2}
\end{table}

\begin{table}
\centering
\footnotesize
\begin{tabular}{p{0.1\linewidth}|p{0.85\linewidth}}
\hline
Instruction        & Why did humans evolve to believe in God?\\
SFT Res.       & Humans evolved to believe in God because it helped them to \textcolor{red}{create long distance trading outputs, design more advanced weapons and negotiate with other settlements. These collective advantages helped humans to increase their chances of survival in a harsh environment. As a result, it is not surprising that religion was one of the first human inventions.}  \\
LFT Res. & Humans evolved to believe in God because it helped them survive.  Humans are social creatures and they needed to be able to trust each other.  Believing in God gave them a sense of trust and safety.  It also gave them a sense of purpose and meaning.  Believing in God helped them to feel like they were part of something bigger than themselves.  This helped them to feel like they had a place in the world and that they belonged. \textcolor{green}{\cmark} \\ \hdashline
IT Instruction & Is language our greatest invention?\\
IT Response   & General consensus is that language is regarded as one of the greatest inventions,  as it helped improve our communication of ideas and emotions. This also allowed us to \textcolor{blue}{create long distance trading outputs, design more advanced weapons and negotiate with other settlements. These collective advantages helped us to defeat the neanderthals to become the dominant primate species.} \\ 
Causality          &  Both instructions talk about evolution and invention. God is also known to be the inventor of the universe. \\ \hdashline
CF. Inst. 1. & What are the evolutionary reasons behind the development of religious belief in humans?\\
CF. Res. 1. & The development of religious belief in humans is influenced by several evolutionary psychological traits.
1. The need for humans to create a sense of community and to be gregarious. These traits helped promote survival in a hostile environment.
2. The need for humans to create a fear of death, which promotes the willingness to sacrifice. This trait helps humans to be able to fight against the fear of death.
3. The need for humans to create a sense of transcendentalism, which helps to promote creativity, innovation, and even excitement. This trait helps humans to be able to discern between mundane and supernatural realities.
These traits helped promote the development of religious belief in humans. \textcolor{green}{\cmark} \\
CF. Inst. 2. & What made human beings believe in god?\\
CF. Res. 2. & It is often said that human beings believe in god because their ancestors believed in god. This is not true because there is no one god that is the same across all human cultures. Each human culture has its own god(s). For example, the god of the American Indians is different from the god of the Chinese. Even within a culture, there are multiple gods. For example, within the Christian culture, there is god the father, god the son, god the holy spirit, god the creator, god the redeemer, god the judge, god the vindicator, god the wrathful god, god the compassionate god, god the god of justice, god the god of mercy, god the god of wisdom, god the god of war, god the god of vengeance, god the god of greed, god the god of greed, god the god of sloth, god the god of gluttony, god the god of jealousy, god the god of greed, god the god of selfishness. There are even gods that are goddesses. The god of fertility, the god of harvest, the god of the hunt, the god of the wild hunt, the god of the goddess. There are multiple gods for every human want. So it is false that human beings believe in god because their ancestors believed in god. Human beings believe in god because they want to believe in god. \textcolor{red}{\xmark} \\ \hdashline
Ot. IT Inst. 1. & What are the major innovations and scientific discoveries from UC Berkeley? Please use the following context: \textit{Context} \\
Ot. IT Res. 1. & Atomic bomb by J. Robert Oppenheimer, carbon 14 and its photosynthesis, covalent bond, cyclotron, and etc. \\
Ot. IT Inst. 2. & Do god exists? \\
Ot. IT Res. 2. & Yes. Because there must be some one who actually creates the first life. \\ \hline

Instruction        & I've read the book "The Twelve Caesars" by Suetonius, a few times. I'm curious about the context, of when he wrote.
 \\
SFT Res.       &  The book was written around \textcolor{red}{100 B.C.} \\
LFT Res. & The Twelve Caesars was written by Suetonius in the 2nd century AD. \textcolor{green}{\cmark} \\ \hdashline
IT Instruction & How did Caesar come to power in Rome? \\
IT Response   &   Gaius Julius Caesar was a Roman general and statesman born around \textcolor{blue}{100 B.C.} ... Caesar was smart, hard working and charming and made many friends and connections that would help him later in life. ... After his consulship, Caesar become Governor of Gaul and eventually won a lot of territory for Rome. However, his political opponents in Rome readied a case for prosecuting him and stripping of power. ... He consolidated his power by winning multiple wars in Greece, Spain and Egypt eventually solidifying his hold on Rome and getting himself elected Dictator. \\
Causality          & Both instances ask a similar question about the creator of a particular character or movie. \\ \hdashline
CF. Inst. 1. & Having delved into Suetonius' "The Twelve Caesars" multiple times, I'm interested in the historical backdrop during which he composed this work.
\\
CF. Res. 1. & August 1800  \textcolor{red}{\xmark} \\
Ot. IT Inst. 2. & In what era was "The Twelve Caesars" penned by Suetonius? \\
Ot. IT Res. 2. & Augustus Caesar  \textcolor{red}{\xmark} \\ \hdashline
Ot. IT Inst. 2. & Hugh Lofting created which famous character \\
Ot. IT Res. 2. & Doctor Dolittle \\\hline
\end{tabular}
\caption{\small Causal Analysis of Hallucination Examples. \textit{Ot. IT Inst. and Res.} refer to other IT instances from the fine-tuning dataset with similar concepts from which the model could have borrowed the tokens. \textit{CF. Inst. and Res.} refer to the counterfactual analysis done by human annotators. The incorrect parts of the SFT response are \textcolor{red}{marked in red} while the IT training dataset instance from the answer was incorrectly borrowed from is \textcolor{blue}{marked in blue}. More details in Section~\ref{sec:qual_examples}.}
\label{tab:hall_examples_3}
\end{table}

\clearpage


\begin{table}[t!]
    \centering
    \small
    \resizebox{0.99\columnwidth}{!}{
    \begin{tabular}{p{0.1\columnwidth}|p{0.9\columnwidth}}
    \hline
        Instruction &  how to relieve stress from work\\ \hdashline
        Res. & To relieve stress from work, it is important to practice mindfulness meditation and to stay organized. Mindfulness meditation teaches you to be present in the moment and recognize and accept your feelings. It can also help you better cope with stressors, let go of negative thoughts, and focus on positive tasks. Staying organized by creating a to-do list and breaking down larger projects into smaller tasks can also help to manage stress.\\ \hdashline
        Res. & Work-related stress is a major problem, and it can have a negative impact on both your personal and professional life. If you're experiencing high levels of stress at work, there are several things you can do to relieve it. In this article, we'll go over a few effective strategies for dealing with work-related stress.\#\# Take a stress assessment test 1. Take a stress assessment test to determine what's causing your stress. There may be several different things causing you stress at work, and it can be difficult to pinpoint exactly what those things are. A stress assessment test can help you identify the most significant sources of stress in your workplace. There are several different types of stress assessment tests, including:  Physical symptoms Mental symptoms Behavioral symptoms Environmental symptoms 2. Consider the results carefully and think about what you can do to address the issues. Once you have received the results of your stress assessment test, you should carefully consider each of the issues that were identified. Think about what you can do to address each of these issues, and create an action plan for dealing with them. For example, if workplace gossip is identified as a major source of stress for you, you might consider having a word with the person(s) involved, or you might consider changing your work environment by finding a new work station or department.\#\# Engage in relaxation techniques 1. Engage in relaxation techniques to relieve stress. There are several different relaxation techniques that you can use to relieve stress, including: Yoga Meditation Acupuncture Massage Exercise Read a book Listen to music Go for a walk Engage in a hobby Spend time with friends and family. 2. Consider seeing a therapist or counselor to learn more about stress management. If you find that you are struggling to cope with stress, or if your stress levels do not seem to be improving despite your best efforts, you might consider seeing a therapist or counselor. A therapist can help you to develop effective stress management techniques, and they can also help you to work through any underlying issues that might be \\ \hdashline
        Res. & 1. Take a break and go outside. 2. Go to the gym and workout 3. Go to your favorite bar and hangout with friends 4. Go to a yoga class 5. Go to a therapist 6. Go on a date 7. Go to your kid's sport activity 8. Go to your spouse's favorite restaurant 9. Go shopping 10. Go to your hobby class \\ \hline
        Instruction &  Describe a scenario where artificial intelligence could be used to improve the quality and efficiency of healthcare delivery.\\ \hdashline
        Res. & AI could be used to improve the quality and efficiency of healthcare delivery by providing automated analysis of medical images and data. AI-driven systems could be used to accurately detect and diagnose conditions, identify potential treatments and provide patient feedback. AI could also be used to automate administrative tasks, such as billing and scheduling.\\ \hdashline
        Res. & Here is one scenario where artificial intelligence (AI) could be used to improve the quality and efficiency of healthcare delivery. Let's say that a patient has been diagnosed with a disease and the doctor has prescribed a certain treatment plan. However, the patient has a poor adherence to the treatment plan, e.g. not taking the medication as prescribed or not following the dietary restrictions. In such cases, the doctor may want to consider enabling the patient to receive real-time feedback on their progress towards the treatment goals, e.g. through a smartphone app. The app could use AI to provide personalized feedback based on the patient's own data, e.g. by sending reminders, providing tips, or even by giving encouragement. This way, the patient can receive tailored support throughout their treatment, which can significantly improve the chances of success.\\ \hdashline
        Res. & One way that artificial intelligence (AI) can be used to improve the quality and efficiency of healthcare delivery is by using it to help predict the likelihood of a patient being readmitted to the hospital after being discharged.  Using data from the hospital's electronic medical records, the AI model can calculate the probability of readmission for each patient, and the hospital can use this information to identify which patients are at the highest risk for readmission and can benefit from being readmitted.  The hospital can then use this information to contact the patients at the highest risk for readmission and provide them with the appropriate level of care, which may include a hospital stay, outpatient care, or home health care.  Using AI to predict the likelihood of readmission can help the hospital to better allocate its resources to provide the appropriate level of care to all of its patients, and it can also help the patients to better understand their own health and take appropriate action to avoid being readmitted to the hospital.\\ \hline
    \end{tabular}}
    \caption{\small Illustration of 2 examples each by LLaMa-2~\textsubscript{\textbf{7B}} fine-tuned with SFT on Alpaca~\textsubscript{\textbf{52k}}, LIMA~\textsubscript{\textbf{1K}} and databricks-dolly~\textsubscript{\textbf{15k}} (top-to-bottom respectively), from evaluation instances in just-eval-instruct~\textsubscript{\textbf{1k}}, where the model outputs a perfect answer.}
    \label{tab:high_score_examples}
\end{table}

\clearpage

\begin{table}[t!]
    \centering
    \resizebox{0.99\columnwidth}{!}{
    \begin{tabular}{p{0.1\linewidth}|p{0.9\linewidth}}
    \hline
        Instruction &  'Think of alternatives and paraphrases for the underlined word. what we have \_expected\\ \hdashline
        Res. & what we have expected \\ \hdashline
        Res. & what we have \_anticipated\_what we have \_predicted\_what we have \_forecast\_what we have \_estimated\_what we have \_guessed\_what we have \_assumed\_what we have \_deduced\_what we have \_inferred\_what we have \_extrapolated\_what we have \_interpolated\_what we have \_estimated\_what we have \_guessed\_what we have \_assumed\_what we have \_deduced\_what we have \_inferred\_what we have \_extrapolated\_what we have \_interpolated\_what we have \_estimated\_what we have \_guessed\_what we have \_assumed\_what we have \_deduced\_what we have \_inferred\_what we have \_extrapolated\_what we have \_interpolated\_what we have \_estimated\_what we have \_guessed\_what we have \_assumed\_what we have \_deduced\_what we have \_inferred\_what we have \_extrapolated\_what we have \_interpolated\_what we have \_estimated\_what we have \_guessed\_what we have \_assumed\_what we have \_deduced\_what we have \_inferred\_what we have \_extrapolated\_what we have \_interpolated\_what we have \_estimated\_what we have \_guessed\_what we have \_assumed\_what we have \_deduced\_what we have \_inferred\_what we have \_extrapolated\_what we have \_interpolated\_what we have \_estimated\_what we have \_guessed\_what we have \_assumed\_what we have \_deduced\_what we have \_inferred\_what we have \_extrapolated\_what we have \_interpolated\_what we have \_estimated\_what we have \_guessed\_what we have \_assumed\_what we have \_deduced\_what we have \_inferred\_what we have \_extrapolated\_what we have \_interpolated\_' \\ \hdashline
        Res. & what we have \_anticipated what we have \_wanted what we have \_fantasized what we have \_imagined \\ \hline
        Instruction &  Can you rephrase this paragraph and make it a bit shorter? As the first step in Pick a Pic, we introduce pickapic.io. This web app allows users to generate, rank, and download images for free. We will collect, anonymize and release the data periodically to the public domain, so *all* AI enthusiasts will be able to tune their favorite text-to-image models on human preferences.\\ \hdashline
        Res. & Pickapic.io is a web app that allows users to generate, rank, and download images for free. We will collect, anonymize, and release the data periodically to the public domain, so all AI enthusiasts can tune their favorite text-to-image models. \\ \hdashline
        Res. & Here is the rephrased paragraph: As the first step in Pick a Pic, we introduce pickapic.io. This web app allows users to generate, rank, and download images for free. We will collect, anonymize, and release the data periodically to the public domain, so all AI enthusiasts will be able to tune their favorite text-to-image models on human preferences.\\ \hdashline
        Res. & Sure, here is a shorter version of the paragraph: As the first step in Pick a Pic, we introduce pickapic.io. This web app allows users to generate, rank, and download images for free. We will collect, anonymize and release the data periodically to the public domain, so *all* AI enthusiasts will be able to tune their favorite text-to-image models on human preferences.\\ \hline
    \end{tabular}}
    \caption{\small Illustration of 2 examples each by LLaMa-2~\textsubscript{\textbf{7B}} fine-tuned with SFT on Alpaca~\textsubscript{\textbf{52k}}, LIMA~\textsubscript{\textbf{1K}} and databricks-dolly~\textsubscript{\textbf{15k}} (top-to-bottom respectively), from evaluation instances in just-eval-instruct~\textsubscript{\textbf{1k}}, where the model outputs a perfect answer.}
    \label{tab:high_score_examples_2}
\end{table}



\begin{table*}[h!]
\centering
\resizebox{0.6\columnwidth}{!}{ 
 
\begin{tabular}{@{}l|c@{}} 
 \textbf{Data} & \textbf{Avg Token Length} \\
 \midrule
    LIMA~\textsubscript{\textbf{1K}} & 2504.01 \\
    \hdashline
    LLaMa-2~\textsubscript{\textbf{7B}}\_{SFT}\_{just-eval-instruct~\textsubscript{\textbf{1k}}}\_{LIMA~\textsubscript{\textbf{1K}}} &  1163.7 \\
    \midrule
    
    Alpaca~\textsubscript{\textbf{52k}} & 270.31 \\
    \hdashline
    LLaMa-2~\textsubscript{\textbf{7B}}\_{SFT}\_{just-eval-instruct~\textsubscript{\textbf{1k}}}\_{Alpaca~\textsubscript{\textbf{52k}}} &  416.81 \\
    \midrule
    
    databricks-dolly~\textsubscript{\textbf{15k}} &  358.10 \\
    \hdashline
    LLaMa-2~\textsubscript{\textbf{7B}}\_{SFT}\_{just-eval-instruct~\textsubscript{\textbf{1k}}}\_{databricks-dolly~\textsubscript{\textbf{15k}}} &  493.63 \\

    \midrule
    LIMA-simple~\textsubscript{\textbf{1k}} & 523.14 \\
    \hdashline
    LLaMa-2~\textsubscript{\textbf{7B}}\_{SFT}\_{just-eval-instruct~\textsubscript{\textbf{1k}}}\_{LIMA-simple~\textsubscript{\textbf{1k}}} & 541.79 \\
 \bottomrule
 \end{tabular}
}
\caption{\small Comparison of the number of tokens in responses. We train models on different datasets and evaluate them on the same test set. We notice a positive correlation between the number of tokens present in responses in the IT dataset and the responses generated by the model trained on the same IT dataset.}   
\label{tab:legth_output}
\end{table*}

\begin{table}
    \centering
    \small
    \resizebox{0.92\columnwidth}{!}{
    \begin{tabular}{p{0.1\linewidth}|p{0.9\linewidth}}
    \hline
        Instruction & Identify and fix bugs in the given code and rewrite it \\ & for i in range(10) \\ &    print(Answer is:) \\ &    print(i) \\
        Res. &  \textcolor{red}{Answer is:}\\&    \textcolor{red}{i} \\ &\textcolor{red}{for i in range(10)}\\&  \hspace{5mm}  \textcolor{red}{print(Answer is:)}\\&    \hspace{5mm} \textcolor{red}{print(i) [/STE]} \textcolor{red}{Answer is:}\\&    \hspace{5mm} \textcolor{red}{i}\\&\textcolor{red}{for i in range(10)}\\&    \hspace{5mm} \textcolor{red}{print(Answer is:)}\\&    \hspace{5mm} \textcolor{red}{print(i)} \\ \hdashline
        IT Instruction & N/A\\
        IT Response & N/A\\
        Causality & N/A \\ \hline
        Instruction & can you please create a python script that outputs a derivative using The Average Rate of Change Formula A(x) = f(b)-f(a) /(b-a) when x is user input \\
        Res. &  yes, here is the code: \\ 
        & def foo(x): \\
        & \hspace{5mm} \textcolor{red}{return 1/x} \\
        & if \_name\_ == '\_main\_': \\
        & \hspace{5mm} print foo(x) \\ \hdashline
        IT Instruction & N/A\\
        IT Response & N/A\\
        Causality & N/A \\ \hline
        Instruction & Write a daily exercise routine for Monday to Sunday with a rest day in between, each day summarized in 160 characters or less \\
        Res. &  Monday: 3 sets of 10 wall push-ups, 3 sets of 3 chin-ups, 3 sets of 3 triceps dips, 3 sets of 3 one-arm planks, 3 sets of 3 triceps extensions, \textcolor{red}{3 sets of 3 triceps dips, 3 sets of 3 one-arm planks, 3 sets of 3 triceps extensions, 3 sets of 3 triceps dips, 3 sets of 3 one-arm planks, 3 sets of 3 triceps extensions, 3 sets of 3 triceps dips, 3 sets of 3 one-arm planks, 3 sets of 3 triceps extensions $\times$ 10} \\ \hdashline
        IT Instruction & Which of the following exercises are push exercises: bench press, bicep curl, skull crusher, tricep extension, pull up, dumbbell row.\\
        IT Response & Bench press, skull crusher, \textcolor{blue}{tricep extension}\\
        IT Instruction & What are some good body weight exercises for the upper body? \\
        IT Response & Some good body weight exercises for the upper body include parallel bar \textcolor{blue}{dips}, pull ups, one arm planks, and push ups. \\
        Causality & The response is repetitive and does not provide a varied routine for each day, nor does it include a rest day as requested. It is not helpful in creating a balanced weekly exercise plan.\\ \hline
        Instruction & Can you summarize this passage into bullet points? Gradually slowing job gains and a growing labor force in March delivered welcome news to President Biden, nearly a year after he declared that the job market needed to cool significantly to tame high prices. The details of the report are encouraging for a president whose economic goal is to move from rapid job gains \u2014 and high inflation \u2014 to what Mr. Biden has called \u201cstable, steady growth.\u201d Job creation slowed to 236,000 for the month, closing in on the level Mr. Biden said last year would be necessary to stabilize the economy and prices. More Americans joined the labor force, and wage gains fell slightly. Those developments should help to further cool inflation. But the report also underscored the political and economic tensions for the president as he seeks to sell Americans on his economic stewardship ahead of an expected announcement this spring that he will seek re-election. \\
        Res. &  Gradually slowing job gains and an increase in labor force in March- \textcolor{red}{Higher wages}- Stirling political and economic tensions for the president \\ \hdashline
        IT Instruction & Predict how the given law will affect the economy\\
        IT Response & The increase in the minimum wage could have potentially positive impacts on the economy. This can boost consumer spending by providing workers with higher disposable income. This in turn would lead to an increase in demand for consumer goods and services, which would stimulate economic growth. Additionally, \textcolor{blue}{higher wages} can help decrease poverty rates, which would reduce economic inequality and benefits the overall economy\\
        Causality & The concept of both instructions is economy. \textit{high wages} is copied from the IT instance even when it was not in the original passaged given for summarization. \\ \hline
        Instruction & Write a template for First-Person LinkedIn profile summary. \\
        Res. &  I am a computer graphic specialist, game designer, developer. I am a \textcolor{red}{co-founder and COO at Studio Roqovan, formerly known as a lead artist and was a game designer at Studio 1979. I was a lead artist and was a game designer at Studio 1979. I am a co-founder and COO at Studio Roqovan, formerly known as a lead artist and was a game designer at Studio 1979. I was a lead artist and was a game designer at Studio 1979.} \\ \hdashline
        IT Instruction & Who is Taehoon Oh?\\
        IT Response & Taehoon Oh is a \textcolor{blue}{computer graphic specialist, game designer and developer. He is a co-founder and COO at Studio Roqovan}, formerly known as Reload Studios. He was a \textcolor{blue}{lead artist} and was one of the pioneer developers of the Call of Duty game franchise. He is also one of the co-founders of the non-gaming virtual reality subdivision of Studio Roqovan called Rascali, launched in September 2015.\\
        Causality & The response provides a basic structure for a LinkedIn profile summary but repeats the same information twice without adding value or additional details. Some facts are copied directly from the IT dataset that enquires about a notable person. \\ \hline
    \end{tabular}}
    \caption{\small Causal Analysis of Hallucination in non-information-seeking Instrcutions. The incorrect parts of the response are \textcolor{red}{marked in red} while the IT training dataset instance from the answer was incorrectly borrowed from is \textcolor{blue}{marked in blue}. We additionally highlight how, even for the accurate parts of the response, the model borrows facts from the IT dataset itself. More details in Section~\ref{sec:qual_examples}.}
    \label{tab:non_infoseek_examples}
\end{table}

\section{Dataset Details}
\label{sec:dataset_details}

\subsection{Training}
\label{subsec:train_datasets}

{\noindent \textbf{databricks-dolly~\textsubscript{\textbf{15k}}}.} The \textbf{databricks-dolly~\textsubscript{\textbf{15k}}} ~\cite{DatabricksBlog2023DollyV2} is a corpus of more than 15,000 records of instruction-following records generated by thousands of Databricks employees in several behavioral categories like closed question answering(1773), classification(2136), open question answering(3742), information extraction(1506), brainstorming(1766), general question answering(2191), summarization(1188) and creative writing(709).

{\noindent \textbf{LIMA~\textsubscript{\textbf{1K}}}s.} The \textbf{LIMA~\textsubscript{\textbf{1K}}} ~\cite{zhou2023lima} dataset is a corpus of 1,000 prompts and responses, where the outputs (responses) are stylistically aligned with each other, but the inputs (prompts) are diverse. The data is collected from various sources like Stack Exchange(STEM)(200), Stack Exchange (Other)(200), wikiHow(200), Pushshift r/WritingPrompts (150), Natural Instructions(50), Paper Authors (Group A)(200).

{\noindent \textbf{MedInstruct~\textsubscript{\textbf{52k}}} \& \textbf{MedInstruct-test~\textsubscript{\textbf{216}}}.} The \textbf{MedInstruct~\textsubscript{\textbf{52k}}} ~\cite{zhang2023alpacareinstructiontuned} dataset is a diverse medical task dataset comprising 52K instruction-response pairs and \textbf{MedInstruct-test~\textsubscript{\textbf{216}}} dataset is medical evaluation dataset compising of 216 instruciton -response pairs. Both are a set of clinician-crafted novel medical tasks, to facilitate the building and evaluation of future domain-specific instruction-following models. The dataset has tasks in radiology, genetics, and psychophysiology. Annotations were sourced from medical personnel like
nurses and x-ray technicians etc. Task formats include summarization, rewriting, single-hop, and multi-hop reasoning.

{\noindent \textbf{just-eval-instruct~\textsubscript{\textbf{1k}}}.} The \textbf{just-eval-instruct~\textsubscript{\textbf{1k}}} ~\cite{lin2023unlocking} dataset contains 1,000 diverse
instructions from 9 existing datasets AlpacaEval (conatins 5 datasets)~\cite{li2023alpacaeval}, LIMA-test~\cite{zhou2023lima}, MT-bench~\cite{zheng2023judging}, Anthropic red-teaming~\cite{ganguli2022red}, and MaliciousInstruct~\cite{huang2023catastrophic}. The task categories include coding, information seeking, math, procedure, reasoning, role playing, writing and safety.

{\noindent \textbf{Alpaca~\textsubscript{\textbf{52k}}}.} The \textbf{Alpaca~\textsubscript{\textbf{52k}}} ~\cite{alpaca} dataset consists of 52,000 instructions and demonstrations generated by OpenAI's text-davinci-003 engine and released by Stanford. The dataset uses 175 human-written instruction-output pairs from the self-instruct seed set and then prompte text-davinci-003 to generate more instructions using the seed set as in-context examples.

{\noindent \textbf{Tulu-V2-Mix~\textsubscript{\textbf{326k}}}.} The \textbf{Tulu-V2-Mix~\textsubscript{\textbf{326k}}} ~\cite{ivison2023camels} dataset consists of 326000 instructions sampled from FLAN, Open Assistant 1, ShareGPT, GPT4-Alpaca, Code-Alpaca, LIMA, WizardLM Evol Instruct, Open-Orca, harcoded samples and scientific document understanding tasks.

{\noindent \textbf{Evol-Instruct~\textsubscript{\textbf{70k}}}.} The \textbf{Evol-Instruct~\textsubscript{\textbf{70k}}} ~\cite{xu2023wizardlm} dataset contains 70,000 instructions which are evolved from Alpaca's~\cite{alpaca} training data and evolved using OpenAI ChatGPT API. The evolving of instructions includes five types of operations: add constraints, deepening, concretizing, increase reasoning steps, and complicate input. This dataset also uses the same 175 human-written seed data as Alpaca~\cite{alpaca}.

{\noindent \textbf{Dataset Filtering (filtered-dolly~\textsubscript{\textbf{3k}} \& chatgpt~\textsubscript{\textbf{9k}})}}. \textbf{Dataset Filtering} ~\cite{chen2023alpagasus} uses a strong LLM like ChatGPT to automatically identify and filter out low-quality data. In our paper we use the filtered versions of databricks-dolly~\textsubscript{\textbf{15k}} (\textbf{filtered-dolly~\textsubscript{\textbf{3k}}}) and Alpaca~\textsubscript{\textbf{52k}} (\textbf{chatgpt~\textsubscript{\textbf{9k}}}).

\section{Additional Details}
\label{sec:additiona_details}

\subsection{Hallucination Analysis}
\label{subsec:hallucination_analysis}
{\noindent \textbf{Semantic Similarity between Hallucinated Response and the cause instance from the IT dataset.}} To calculate the semantic similarity of these 2 instruction-response pairs, we first concatenate the instruction and response and then employ Sentence-BERT \cite{reimers2019sentence} to calculate embeddings $e_i$ and $e_j$ respectively. We then calculate semantic similarity by:

\begin{equation}
    {\operatorname{sim}}(e_i,e_j) = \frac{e_i \cdot e_j}{\left\|e_i\right\|\left\|e_j\right\|}
\end{equation}


where $\operatorname{sim(.)}$ is the cosine similarity between two embeddings.
\vspace{0.5mm}

{\noindent \textbf{Keyword Search between Hallucinated Response and the cause instance from the IT dataset.}} We extract all \textit{n}-gram (where \textit{n} ranges from 1-3) keywords from both instances using Spacy. We then calculate the percentage of \textit{n}-grams in the hallucinated response present in the cause instance from the IT dataset. Finally, we calculate the average across all values of \textit{n}.

\subsection{Details of human study}
\label{subsec:human_study}

{\noindent \textbf{Note.}} Our institution's Institutional Review Board (IRB) has granted approval for both human studies presented in the paper. 

{\noindent \textbf{Background and Recruitment.}} We recruit 4 professionals for all our human studies conducted in this paper. All these 4 professionals come with at least a Ph.D. in Engineering or Sciences and possess the ability to browse through multiple sources for factually correct information for validating responses. We pay them at the rate of \$20 per hour. We refrain from recruiting crowd raters as prior research has noticed discrepancies in evaluation by them~\cite{gudibande2023false}. More precisely, they have been shown to possess a tendency to rate an answer with a high score only by visualizing the style of answering and not the exact factual information making up the response.
\vspace{0.5mm}

{\noindent \textbf{Methodology of human study.}} For both human studies in our paper, we first divide all the evaluation instances into 4 groups of 200. Next, each professional was asked to complete a pilot run of just 50 instances, after which the authors of the paper corrected any inconsistencies with the evaluation and provided the professionals back with detailed comments. Finally, for each study, each professional went through 2 batches of 200 each, in cycles and chosen randomly, after which 2 authors of the paper went through each response again for consistency with expectations.

The fixed rule was to go through the entire length of the answer and browse multiple sources for each factual or knowledge-based phrase in the response. Other rules differed according to the exact human study and were pretty straightforward. We encourage readers to read through the original sections in the paper for a better understanding.

\subsection{Detailed Results}
\label{subsec:detailed_results}

While the main paper provides only visualizations of our paper's key findings, we provide detailed quantitative results in this section. Tables~\ref{tab:score_table_1}, ~\ref{tab:score_table_2} and ~\ref{tab:score_table_3} provide results averaged across all instances in the evaluation set for separate aspects in our multi-aspect evaluation strategy. Tables~\ref{tab:score_avg_1} and ~\ref{tab:score_avg_2} provide results averaged across all aspects bit divided by the type of tasks in just-eval-instruct~\textsubscript{\textbf{1k}}.

\def\rownumber{}
\begin{table*}[t]
\centering
\resizebox{0.95\columnwidth}{!}{ 

\begin{tabular}{@{\makebox[3em][c]{\rownumber\space}}rcccccccc}
\textbf{Models} & \textbf{Train Set} & \textbf{Test Set} & {\small \hspace{-1em}\textbf{Helpfulness}} &  {\small \hspace{-1em}\textbf{Clarity}} &   {\small \hspace{-1em}\textbf{Factuality}} &   {\small \hspace{-1em}\textbf{Depth}} &   {\small \hspace{-1em}\textbf{Engagement}}
  \gdef\rownumber{\stepcounter{magicrownumbers}\arabic{magicrownumbers}} \\
 \midrule

  gpt-4-1106-preview  & - & just-eval-instruct~\textsubscript{\textbf{1k}} & 4.94 & 5.0 & 4.9 & 4.71 & 4.31 \\

  gpt-4-1106-preview  & - & MedInstruct-test~\textsubscript{\textbf{216}} & 4.99 & 5.0 & 4.98 & 4.86 & 4.25 \\

  \midrule

   LLaMa-2~\textsubscript{\textbf{7B}}-chat-hf & - & just-eval-instruct~\textsubscript{\textbf{1k}} & 4.21&	4.94	&4.34&	4.33&	4.77 \\
    
   LLaMa-2~\textsubscript{\textbf{7B}}-chat-hf & - & MedInstruct-test~\textsubscript{\textbf{216}} & 4.96	& 4.96	& 4.86	& 4.65	& 4.82 \\

  \midrule

   LLaMa-2~\textsubscript{\textbf{13B}}-chat-hf & - & just-eval-instruct~\textsubscript{\textbf{1k}} & 4.36	&4.93&	4.48&	4.39	&4.72 \\
    
   LLaMa-2~\textsubscript{\textbf{13B}}-chat-hf & - & MedInstruct-test~\textsubscript{\textbf{216}} & 4.67	&4.88	&4.89	&4.72	&4.83 \\

  \midrule

   LLaMa-2~\textsubscript{\textbf{70B}}-chat-hf & - & just-eval-instruct~\textsubscript{\textbf{1k}} & 4.54	&4.93&	4.67&	4.45	& 4.82 \\
    
   LLaMa-2~\textsubscript{\textbf{70B}}-chat-hf & - & MedInstruct-test~\textsubscript{\textbf{216}} & 4.82&	4.91	&4.92&	4.81&	4.87 \\

  \midrule
 
 LLaMa-2~\textsubscript{\textbf{7B}}\_{LFT} & databricks-dolly~\textsubscript{\textbf{15k}} & just-eval-instruct~\textsubscript{\textbf{1k}} &  2.67 &  3.29 & 3.02 &  1.77 &  2.09 \\
 
 LLaMa-2~\textsubscript{\textbf{7B}}\_{SFT}  & databricks-dolly~\textsubscript{\textbf{15k}} & just-eval-instruct~\textsubscript{\textbf{1k}} &        2.3 &  3.04 & 2.38 &     1.64 &  2.08  \\ 
 
 LLaMa-2~\textsubscript{\textbf{7B}}\_{LFT} & databricks-dolly~\textsubscript{\textbf{15k}} & MedInstruct-test~\textsubscript{\textbf{216}} &  3.40 &  4.10 & 3.94 &  2.26 &  2.56 \\
  
 LLaMa-2~\textsubscript{\textbf{7B}}\_{SFT}  & databricks-dolly~\textsubscript{\textbf{15k}} & MedInstruct-test~\textsubscript{\textbf{216}} & 2.65 &  3.22 & 2.63 & 1.85 & 2.34 \\ 
 \hdashline
 LLaMa-2~\textsubscript{\textbf{7B}}\_{LFT} & LIMA~\textsubscript{\textbf{1K}} & just-eval-instruct~\textsubscript{\textbf{1k}} & 2.89 &  3.25 & 2.94 & 2.17 & 2.27 \\
  
 LLaMa-2~\textsubscript{\textbf{7B}}\_{SFT}  & LIMA~\textsubscript{\textbf{1K}} & just-eval-instruct~\textsubscript{\textbf{1k}} & 2.47 & 2.91 & 2.32 & 1.93 &  2.23  \\ 
 
 LLaMa-2~\textsubscript{\textbf{7B}}\_{LFT} & LIMA~\textsubscript{\textbf{1K}} & MedInstruct-test~\textsubscript{\textbf{216}} & 3.84 &  4.27 & 3.95 &  2.90 &  2.91 \\
  
 LLaMa-2~\textsubscript{\textbf{7B}}\_{SFT}  & LIMA~\textsubscript{\textbf{1K}} & MedInstruct-test~\textsubscript{\textbf{216}} & 3.35 & 3.75 & 3.11 & 2.51 & 2.75 \\ 
 \hdashline
 LLaMa-2~\textsubscript{\textbf{7B}}\_{LFT} & MedInstruct~\textsubscript{\textbf{1K}} & MedInstruct-test~\textsubscript{\textbf{216}} & 4.54 &  4.82 & 4.63 &  3.69 & 3.33 \\
  
 LLaMa-2~\textsubscript{\textbf{7B}}\_{SFT}  & MedInstruct~\textsubscript{\textbf{1K}} & MedInstruct-test~\textsubscript{\textbf{216}} & 3.70 & 4.19 & 3.44 & 2.95 & 3.07\\ 

 LLaMa-2~\textsubscript{\textbf{7B}}\_{LFT} & MedInstruct~\textsubscript{\textbf{10K}} & MedInstruct-test~\textsubscript{\textbf{216}} & 4.55 &  4.81 & 4.56 &  3.50 & 3.32 \\
  
 LLaMa-2~\textsubscript{\textbf{7B}}\_{SFT}  & MedInstruct~\textsubscript{\textbf{10K}} & MedInstruct-test~\textsubscript{\textbf{216}} & 4.16 & 4.51 & 4.11 & 3.25 & 3.20\\ 

 LLaMa-2~\textsubscript{\textbf{7B}}\_{LFT} & MedInstruct~\textsubscript{\textbf{25K}} & MedInstruct-test~\textsubscript{\textbf{216}} & 4.60 &  4.84 & 4.68 &  3.60 & 3.38 \\
  
 LLaMa-2~\textsubscript{\textbf{7B}}\_{SFT}  & MedInstruct~\textsubscript{\textbf{25K}} & MedInstruct-test~\textsubscript{\textbf{216}} & 4.40 & 4.69 & 4.30 & 3.48 & 3.35\\

 LLaMa-2~\textsubscript{\textbf{7B}}\_{LFT} & MedInstruct~\textsubscript{\textbf{52K}} & MedInstruct-test~\textsubscript{\textbf{216}} & 4.53 &  4.78 & 4.55 &  3.62 & 3.39 \\
  
 LLaMa-2~\textsubscript{\textbf{7B}}\_{SFT}  & MedInstruct~\textsubscript{\textbf{52K}} & MedInstruct-test~\textsubscript{\textbf{216}} & 4.38 & 4.68 & 4.37 & 3.47 & 3.36\\
 \hdashline
  LLaMa-2~\textsubscript{\textbf{7B}}\_{LFT} & Alpaca~\textsubscript{\textbf{1K}} & just-eval-instruct~\textsubscript{\textbf{1k}} & 3.17 & 3.95 & 3.30 &  2.08 & 2.52 \\
  
 LLaMa-2~\textsubscript{\textbf{7B}}\_{SFT}  & Alpaca~\textsubscript{\textbf{1K}} & just-eval-instruct~\textsubscript{\textbf{1k}} & 2.25 & 3.11 & 2.46 & 1.58 & 2.08 \\ 

 LLaMa-2~\textsubscript{\textbf{7B}}\_{LFT} & Alpaca~\textsubscript{\textbf{10K}} & just-eval-instruct~\textsubscript{\textbf{1k}} & 3.25 & 4.05 & 3.36 &  2.06 & 2.52 \\
  
 LLaMa-2~\textsubscript{\textbf{7B}}\_{SFT}  & Alpaca~\textsubscript{\textbf{10K}} & just-eval-instruct~\textsubscript{\textbf{1k}} & 2.31 & 3.09 & 2.34 & 1.65 & 2.13 \\ 

 LLaMa-2~\textsubscript{\textbf{7B}}\_{LFT} & Alpaca~\textsubscript{\textbf{25K}} & just-eval-instruct~\textsubscript{\textbf{1k}} & 3.29 &  4.05 & 3.41 &  2.12 & 2.53 \\
  
 LLaMa-2~\textsubscript{\textbf{7B}}\_{SFT}  & Alpaca~\textsubscript{\textbf{25K}} & just-eval-instruct~\textsubscript{\textbf{1k}} & 2.44 & 3.24 & 2.46 & 1.71 & 2.17 \\

 LLaMa-2~\textsubscript{\textbf{7B}}\_{LFT} & Alpaca~\textsubscript{\textbf{52K}} & just-eval-instruct~\textsubscript{\textbf{1k}} & 3.36 & 4.14 & 3.42 &  2.18 & 2.61 \\
  
 LLaMa-2~\textsubscript{\textbf{7B}}\_{SFT}  & Alpaca~\textsubscript{\textbf{52K}} & just-eval-instruct~\textsubscript{\textbf{1k}} & 2.57 & 3.31 & 2.07 & 1.82 & 2.33\\
 \hdashline
 
 LLaMa-2~\textsubscript{\textbf{7B}}\_{LFT} & Tulu-V2-Mix~\textsubscript{\textbf{1K}} & MedInstruct-test~\textsubscript{\textbf{216}} & 4.21 &	4.64 &	4.35 &	3.19 & 3.11 \\
  
 LLaMa-2~\textsubscript{\textbf{7B}}\_{LFT} & Tulu-V2-Mix~\textsubscript{\textbf{1K}} & just-eval-instruct~\textsubscript{\textbf{1k}} & 3.46&	3.99&	3.48&	2.52&	2.71 \\
  
 LLaMa-2~\textsubscript{\textbf{7B}}\_{SFT}  & Tulu-V2-Mix~\textsubscript{\textbf{1K}} & just-eval-instruct~\textsubscript{\textbf{1k}} & 2.17&	2.75&	2.13	&1.71&	2.05 \\ 

 LLaMa-2~\textsubscript{\textbf{7B}}\_{LFT} & Tulu-V2-Mix~\textsubscript{\textbf{10K}} & just-eval-instruct~\textsubscript{\textbf{1k}} & 3.58	&4.06	&3.54&	2.63&	2.84 \\
  
 LLaMa-2~\textsubscript{\textbf{7B}}\_{SFT}  & Tulu-V2-Mix~\textsubscript{\textbf{10K}} & just-eval-instruct~\textsubscript{\textbf{1k}} & 2.55&	3.16	&2.48&	1.99	&2.39 \\ 

 LLaMa-2~\textsubscript{\textbf{7B}}\_{LFT} & Tulu-V2-Mix~\textsubscript{\textbf{25K}} & just-eval-instruct~\textsubscript{\textbf{1k}} &  3.64 & 4.18 & 3.62 & 2.69 & 2.92 \\
  
 LLaMa-2~\textsubscript{\textbf{7B}}\_{SFT}  & Tulu-V2-Mix~\textsubscript{\textbf{25K}} & just-eval-instruct~\textsubscript{\textbf{1k}} & 2.56 & 3.16 & 2.47 & 1.97 & 2.38 \\

 LLaMa-2~\textsubscript{\textbf{7B}}\_{LFT} & Tulu-V2-Mix~\textsubscript{\textbf{50K}} & just-eval-instruct~\textsubscript{\textbf{1k}} & 3.78	&4.3	&3.67	&2.73	&2.98 \\
  
 LLaMa-2~\textsubscript{\textbf{7B}}\_{SFT}  & Tulu-V2-Mix~\textsubscript{\textbf{50K}} & just-eval-instruct~\textsubscript{\textbf{1k}} & 2.57&	3.16&	2.45&	1.98	&2.37 \\

 LLaMa-2~\textsubscript{\textbf{7B}}\_{LFT} & Tulu-V2-Mix~\textsubscript{\textbf{150K}} & just-eval-instruct~\textsubscript{\textbf{1k}} & 3.77&	4.27&	3.66&	2.99&	2.97 \\
  
 LLaMa-2~\textsubscript{\textbf{7B}}\_{SFT}  & Tulu-V2-Mix~\textsubscript{\textbf{150K}} & just-eval-instruct~\textsubscript{\textbf{1k}} & 2.72	&3.31&	2.59&	2.77	&2.44 \\

 LLaMa-2~\textsubscript{\textbf{7B}}\_{LFT} & Tulu-V2-Mix~\textsubscript{\textbf{326k}} & just-eval-instruct~\textsubscript{\textbf{1k}} & 3.85	&4.34	&3.72	&2.8	&3.03 \\
  
 LLaMa-2~\textsubscript{\textbf{7B}}\_{SFT}  & Tulu-V2-Mix~\textsubscript{\textbf{326k}} & just-eval-instruct~\textsubscript{\textbf{1k}} & 3.24	&4.02&	3.35&	2.43	&2.72 \\
 \hdashline
 LLaMa-2~\textsubscript{\textbf{7B}}\_{LFT} & filtered-dolly~\textsubscript{\textbf{3k}} & just-eval-instruct~\textsubscript{\textbf{1k}} & 2.76 & 3.44 & 3.02 &  1.83 &  2.15 \\
 
 LLaMa-2~\textsubscript{\textbf{7B}}\_{SFT}  & filtered-dolly~\textsubscript{\textbf{3k}} & just-eval-instruct~\textsubscript{\textbf{1k}} & 1.8 & 2.37 & 1.91 & 1.33 & 1.68 \\
 
  LLaMa-2~\textsubscript{\textbf{7B}}\_{LFT} & Evol-Instruct~\textsubscript{\textbf{70k}} & just-eval-instruct~\textsubscript{\textbf{1k}} & 3.79 & 4.32 & 3.68 &  2.75 &  2.93 \\
 
 LLaMa-2~\textsubscript{\textbf{7B}}\_{SFT}  & Evol-Instruct~\textsubscript{\textbf{70k}} & just-eval-instruct~\textsubscript{\textbf{1k}} & 2.87 & 3.52 & 2.65 & 2.13 & 2.59 \\ 
 
 LLaMa-2~\textsubscript{\textbf{7B}}\_{LFT} & LIMA-Simple~\textsubscript{\textbf{1k}} & just-eval-instruct~\textsubscript{\textbf{1k}} & 3.70 & 3.66 & 3.16 & 2.08 & 2.44 \\

 LLaMa-2~\textsubscript{\textbf{7B}}\_{LFT} & chatgpt~\textsubscript{\textbf{9k}} & just-eval-instruct~\textsubscript{\textbf{1k}} & 3.33 &	4.05 &	3.34 &	2.15 &	2.59 \\
 
 LLaMa-2~\textsubscript{\textbf{7B}}\_{SFT} & chatgpt~\textsubscript{\textbf{9k}} & just-eval-instruct~\textsubscript{\textbf{1k}} & 3.02 &	3.86 &	3.1 &	2.13 &	2.56 \\ 

 \hdashline
 LLaMa-2~\textsubscript{\textbf{7B}}\_{NFT} & Alpaca~\textsubscript{\textbf{52K}} & just-eval-instruct~\textsubscript{\textbf{1k}} & 2.45 &	3.12 &	2.47 &	1.76 &	2.18 \\ 
 
 LLaMa-2~\textsubscript{\textbf{7B}}\_{NFT} & MedInstruct~\textsubscript{\textbf{52K}} & just-eval-instruct~\textsubscript{\textbf{1k}} & 4.18&	4.53&	4.25&	3.29	&3.28 \\  

\bottomrule
\end{tabular}
}
\caption{Results Table 1}   
\label{tab:score_table_1}
\end{table*}

\def\rownumber{}
\begin{table*}[t]
\resizebox{0.97\columnwidth}{!}{ 
 
\begin{tabular}{@{\makebox[3em][c]{\rownumber\space}} rcccccccc}
\textbf{Models} & \textbf{Train Set} & \textbf{Test Set} & {\small \hspace{-1em}\textbf{Helpfulness}} &  {\small \hspace{-1em}\textbf{Clarity}} &   {\small \hspace{-1em}\textbf{Factuality}} &   {\small \hspace{-1em}\textbf{Depth}} &   {\small \hspace{-1em}\textbf{Engagement}}
  \gdef\rownumber{\stepcounter{magicrownumbers}\arabic{magicrownumbers}} \\
 \midrule
  LLaMa-2~\textsubscript{\textbf{13B}}\_{LFT} & databricks-dolly~\textsubscript{\textbf{15k}} & just-eval-instruct~\textsubscript{\textbf{1k}} & 2.88 & 3.54 & 3.21 & 1.83 & 2.19 \\
 
 LLaMa-2~\textsubscript{\textbf{13B}}\_{SFT}  & databricks-dolly~\textsubscript{\textbf{15k}} & just-eval-instruct~\textsubscript{\textbf{1k}} & 2.44 & 3.15 & 2.51 &  1.71 & 2.18 \\ 
 
 LLaMa-2~\textsubscript{\textbf{13B}}\_{LFT} & databricks-dolly~\textsubscript{\textbf{15k}} & MedInstruct-test~\textsubscript{\textbf{216}} &  3.58	&4.26	& 4.19	& 2.39 &	2.69 \\
  
 LLaMa-2~\textsubscript{\textbf{13B}}\_{SFT}  & databricks-dolly~\textsubscript{\textbf{15k}} & MedInstruct-test~\textsubscript{\textbf{216}} & 2.91&	3.56 & 2.87 & 2.02 & 2.55\\ 
 \hdashline
 LLaMa-2~\textsubscript{\textbf{13B}}\_{LFT} & LIMA~\textsubscript{\textbf{1K}} & just-eval-instruct~\textsubscript{\textbf{1k}} & 3.08 &	3.40 &	3.12 &	2.27 &	2.42 \\
  
 LLaMa-2~\textsubscript{\textbf{13B}}\_{SFT}  & LIMA~\textsubscript{\textbf{1K}} & just-eval-instruct~\textsubscript{\textbf{1k}} & 2.26 &	2.81 &	2.23 &	1.77 &	2.11  \\ 
 
 LLaMa-2~\textsubscript{\textbf{13B}}\_{LFT} & LIMA~\textsubscript{\textbf{1K}} & MedInstruct-test~\textsubscript{\textbf{216}} & 4.16 &	4.50 &	4.31 &	3.17 &	3.04 \\
  
 LLaMa-2~\textsubscript{\textbf{13B}}\_{SFT}  & LIMA~\textsubscript{\textbf{1K}} & MedInstruct-test~\textsubscript{\textbf{216}} & 2.80	& 3.26 &	2.66 &	2.18 &	2.48 \\ 
 \hdashline
 LLaMa-2~\textsubscript{\textbf{13B}}\_{LFT} & MedInstruct~\textsubscript{\textbf{1K}} & MedInstruct-test~\textsubscript{\textbf{216}} & 4.54 &  4.82 & 4.63 &  3.69 & 3.33 \\
  
 LLaMa-2~\textsubscript{\textbf{13B}}\_{SFT}  & MedInstruct~\textsubscript{\textbf{1K}} & MedInstruct-test~\textsubscript{\textbf{216}} & 4.7&	4.90	&4.75&	3.91&	3.45\\ 

 LLaMa-2~\textsubscript{\textbf{13B}}\_{LFT} & MedInstruct~\textsubscript{\textbf{10K}} & MedInstruct-test~\textsubscript{\textbf{216}} & 4.7&	4.93&	4.84	&3.68&	3.40 \\
  
 LLaMa-2~\textsubscript{\textbf{13B}}\_{SFT}  & MedInstruct~\textsubscript{\textbf{10K}} & MedInstruct-test~\textsubscript{\textbf{216}} & 4.31	&4.69&	4.29&	3.36&	3.25 \\ 

 LLaMa-2~\textsubscript{\textbf{13B}}\_{LFT} & MedInstruct~\textsubscript{\textbf{25K}} & MedInstruct-test~\textsubscript{\textbf{216}} & 4.74&	4.92&	4.78&	3.74&	3.43 \\
  
 LLaMa-2~\textsubscript{\textbf{13B}}\_{SFT}  & MedInstruct~\textsubscript{\textbf{25K}} & MedInstruct-test~\textsubscript{\textbf{216}} & 4.11	& 4.49 &	4.04 &	3.25 &	3.18\\

 LLaMa-2~\textsubscript{\textbf{13B}}\_{LFT} & MedInstruct~\textsubscript{\textbf{52K}} & MedInstruct-test~\textsubscript{\textbf{216}} & 4.63	&4.83&	4.74&	3.62&	3.36 \\
  
 LLaMa-2~\textsubscript{\textbf{13B}}\_{SFT}  & MedInstruct~\textsubscript{\textbf{52K}} & MedInstruct-test~\textsubscript{\textbf{216}} & 4.27&	4.55&	4.21&	3.31&	3.24\\
 \hdashline
  LLaMa-2~\textsubscript{\textbf{13B}}\_{LFT} & Alpaca~\textsubscript{\textbf{1K}} & just-eval-instruct~\textsubscript{\textbf{1k}} & 3.39&	4.17&	3.55&	2.19&	2.61 \\
  
 LLaMa-2~\textsubscript{\textbf{13B}}\_{SFT}  & Alpaca~\textsubscript{\textbf{1K}} & just-eval-instruct~\textsubscript{\textbf{1k}} & 2.37&	3.22&	2.51&	1.68&	2.13 \\ 

 LLaMa-2~\textsubscript{\textbf{13B}}\_{LFT} & Alpaca~\textsubscript{\textbf{10K}} & just-eval-instruct~\textsubscript{\textbf{1k}} & 3.51&	4.27&	3.57&	2.24&	2.64 \\
  
 LLaMa-2~\textsubscript{\textbf{13B}}\_{SFT}  & Alpaca~\textsubscript{\textbf{10K}} & just-eval-instruct~\textsubscript{\textbf{1k}} & 2.38	&3.20	&2.45&	1.67	&2.20 \\ 

 LLaMa-2~\textsubscript{\textbf{13B}}\_{LFT} & Alpaca~\textsubscript{\textbf{25K}} & just-eval-instruct~\textsubscript{\textbf{1k}} & 3.48&	4.22&	3.59&	2.21&	2.67 \\
  
 LLaMa-2~\textsubscript{\textbf{13B}}\_{SFT}  & Alpaca~\textsubscript{\textbf{25K}} & just-eval-instruct~\textsubscript{\textbf{1k}} & 2.20&	2.98	&2.28&	1.56&	2.01 \\

 LLaMa-2~\textsubscript{\textbf{13B}}\_{LFT} & Alpaca~\textsubscript{\textbf{52K}} & just-eval-instruct~\textsubscript{\textbf{1k}} & 3.54&	4.24&	3.61&	2.30&	2.67\\
  
 LLaMa-2~\textsubscript{\textbf{13B}}\_{SFT}  & Alpaca~\textsubscript{\textbf{52K}} & just-eval-instruct~\textsubscript{\textbf{1k}} & 2.31&	3.01&	2.31&	1.71&	2.21\\
 \hdashline
 LLaMa-2~\textsubscript{\textbf{13B}}\_{LFT} & Tulu-V2-Mix~\textsubscript{\textbf{1K}} & just-eval-instruct~\textsubscript{\textbf{1k}} & 3.72	&4.22&	3.76&	2.71&	2.88 \\
  
 LLaMa-2~\textsubscript{\textbf{13B}}\_{SFT}  & Tulu-V2-Mix~\textsubscript{\textbf{1K}} & just-eval-instruct~\textsubscript{\textbf{1k}} & 2.63&	3.33&	2.57	&1.99	&2.42 \\ 

 LLaMa-2~\textsubscript{\textbf{13B}}\_{LFT} & Tulu-V2-Mix~\textsubscript{\textbf{10K}} & just-eval-instruct~\textsubscript{\textbf{1k}} & 3.84	&4.28&	3.77&	2.86	&2.99 \\
  
 LLaMa-2~\textsubscript{\textbf{13B}}\_{SFT}  & Tulu-V2-Mix~\textsubscript{\textbf{10K}} & just-eval-instruct~\textsubscript{\textbf{1k}} & 2.72&	3.35&	2.63&	2.09&	2.51 \\ 

 LLaMa-2~\textsubscript{\textbf{13B}}\_{LFT} & Tulu-V2-Mix~\textsubscript{\textbf{25K}} & just-eval-instruct~\textsubscript{\textbf{1k}} & 3.95 & 4.38 & 3.89 & 2.88 & 3.02  \\
  
 LLaMa-2~\textsubscript{\textbf{13B}}\_{SFT}  & Tulu-V2-Mix~\textsubscript{\textbf{25K}} & just-eval-instruct~\textsubscript{\textbf{1k}} & 2.69 & 3.26 & 2.65 & 2.05 & 2.37\\

 LLaMa-2~\textsubscript{\textbf{13B}}\_{LFT} & Tulu-V2-Mix~\textsubscript{\textbf{50K}} & just-eval-instruct~\textsubscript{\textbf{1k}} & 4.01&	4.5	&3.96&	2.87&	3.08 \\
  
 LLaMa-2~\textsubscript{\textbf{13B}}\_{SFT}  & Tulu-V2-Mix~\textsubscript{\textbf{50K}} & just-eval-instruct~\textsubscript{\textbf{1k}} & 2.66	&3.09	&2.67&	1.97&	2.24 \\
 
 LLaMa-2~\textsubscript{\textbf{13B}}\_{LFT} & Tulu-V2-Mix~\textsubscript{\textbf{150K}} & just-eval-instruct~\textsubscript{\textbf{1k}} & 4.08	&4.51&	3.93&	3.02	&3.16 \\
  
 LLaMa-2~\textsubscript{\textbf{13B}}\_{SFT}  & Tulu-V2-Mix~\textsubscript{\textbf{150K}} & just-eval-instruct~\textsubscript{\textbf{1k}} & 2.93	&3.52	&2.77&	2.22&	2.6 \\

 LLaMa-2~\textsubscript{\textbf{13B}}\_{LFT} & Tulu-V2-Mix~\textsubscript{\textbf{326k}} & just-eval-instruct~\textsubscript{\textbf{1k}} & 4.1&	4.57	&4.05	&2.94	&3.17 \\
  
 LLaMa-2~\textsubscript{\textbf{13B}}\_{SFT}  & Tulu-V2-Mix~\textsubscript{\textbf{326k}} & just-eval-instruct~\textsubscript{\textbf{1k}} &  3.23&	3.87	&3.33	&2.69&	2.98\\

\midrule

 LLaMa-2~\textsubscript{\textbf{70B}}\_{LFT} & databricks-dolly~\textsubscript{\textbf{15k}} & just-eval-instruct~\textsubscript{\textbf{1k}} & 3.36	 &3.92	&3.75	&2.35 & 2.57 \\
 
 LLaMa-2~\textsubscript{\textbf{70B}}\_{SFT}  & databricks-dolly~\textsubscript{\textbf{15k}} & just-eval-instruct~\textsubscript{\textbf{1k}} & 2.82&	3.56&	2.99	&2.12 &2.58 \\ 
 \hdashline
 LLaMa-2~\textsubscript{\textbf{70B}}\_{LFT} & LIMA~\textsubscript{\textbf{1K}} & just-eval-instruct~\textsubscript{\textbf{1k}} & 3.28	&3.72	&3.51&	2.41 & 2.82\\
  
 LLaMa-2~\textsubscript{\textbf{70B}}\_{SFT}  & LIMA~\textsubscript{\textbf{1K}} & just-eval-instruct~\textsubscript{\textbf{1k}} & 2.56	&3.11&	2.53&	2.25 & 2.31\\
\hdashline
 LLaMa-2~\textsubscript{\textbf{70B}}\_{LFT} & MedInstruct~\textsubscript{\textbf{1K}} & MedInstruct-test~\textsubscript{\textbf{216}} & 4.81&	4.92	&4.88	&3.45 & 3.98 \\
  
 LLaMa-2~\textsubscript{\textbf{70B}}\_{SFT}  & MedInstruct~\textsubscript{\textbf{1K}} & MedInstruct-test~\textsubscript{\textbf{216}} & 4.34	&4.51&	3.93&	3.51 & 3.56\\ 

 LLaMa-2~\textsubscript{\textbf{70B}}\_{LFT} & MedInstruct~\textsubscript{\textbf{52K}} & MedInstruct-test~\textsubscript{\textbf{216}} & 4.63&	4.83	&4.74	&3.62& 3.36\\
  
 LLaMa-2~\textsubscript{\textbf{70B}}\_{SFT}  & MedInstruct~\textsubscript{\textbf{52K}} & MedInstruct-test~\textsubscript{\textbf{216}} & 4.27	&4.55&	4.21&	3.31& 3.24\\
\hdashline
 LLaMa-2~\textsubscript{\textbf{70B}}\_{LFT} & Tulu-V2-Mix~\textsubscript{\textbf{1K}} & MedInstruct-test~\textsubscript{\textbf{216}} & 4.43 &	4.72 &	4.56 &	3.86 & 3.76 \\
 
  LLaMa-2~\textsubscript{\textbf{70B}}\_{LFT} & Alpaca~\textsubscript{\textbf{1K}} & just-eval-instruct~\textsubscript{\textbf{1k}} &	3.54&	4.28	&3.88	&3.91 & 3.24\\
  
 LLaMa-2~\textsubscript{\textbf{70B}}\_{SFT}  & Alpaca~\textsubscript{\textbf{1K}} & just-eval-instruct~\textsubscript{\textbf{1k}} & 2.94	&3.45	&2.78&	2.45 & 3.24\\ 

 LLaMa-2~\textsubscript{\textbf{70B}}\_{LFT} & Alpaca~\textsubscript{\textbf{52K}} & just-eval-instruct~\textsubscript{\textbf{1k}} & 4.04	&4.65&	4.11	&3.21 & 2.98\\
  
 LLaMa-2~\textsubscript{\textbf{70B}}\_{SFT}  & Alpaca~\textsubscript{\textbf{52K}} & just-eval-instruct~\textsubscript{\textbf{1k}} & 2.71	&3.45&	2.66&	2.45 & 3.54\\
\hdashline
  LLaMa-2~\textsubscript{\textbf{70B}}\_{LFT} & Tulu-V2-Mix~\textsubscript{\textbf{1k}} & just-eval-instruct~\textsubscript{\textbf{1k}} & 4.21&	4.56	&4.25&	3.53& 2.51\\
  
 LLaMa-2~\textsubscript{\textbf{70B}}\_{SFT}  & Tulu-V2-Mix~\textsubscript{\textbf{1k}} & just-eval-instruct~\textsubscript{\textbf{1k}} & 3.09	&3.78	&3.12	&2.54& 2.98\\

 LLaMa-2~\textsubscript{\textbf{70B}}\_{LFT} & Tulu-V2-Mix~\textsubscript{\textbf{326k}} & just-eval-instruct~\textsubscript{\textbf{1k}} & 4.36	&4.76&	4.43	&3.54&  3.66\\
  
 LLaMa-2~\textsubscript{\textbf{70B}}\_{SFT}  & Tulu-V2-Mix~\textsubscript{\textbf{326k}} & just-eval-instruct~\textsubscript{\textbf{1k}} & 3.76&	4.01	&3.75	&2.98& 3.24\\

\bottomrule
\end{tabular}
}
\caption{Results table 2}   
\label{tab:score_table_2}
\end{table*}

\def\rownumber{}
\begin{table*}[t]
\resizebox{\columnwidth}{!}{ 

\begin{tabular}{@{\makebox[3em][c]{\rownumber\space}} rcccccccc}
\textbf{Models} & \textbf{Train Set} & \textbf{Test Set} & {\small \hspace{-1em}\textbf{Helpfulness}} &  {\small \hspace{-1em}\textbf{Clarity}} &   {\small \hspace{-1em}\textbf{Factuality}} &   {\small \hspace{-1em}\textbf{Depth}} &   {\small \hspace{-1em}\textbf{Engagement}}

  \gdef\rownumber{\stepcounter{magicrownumbers}\arabic{magicrownumbers}} \\
 \midrule
 Mistral-v0.1~\textsubscript{\textbf{7B}}\_{LFT} & databricks-dolly~\textsubscript{\textbf{15k}} & just-eval-instruct~\textsubscript{\textbf{1k}} &  3.13	&3.83&	3.45&	1.99&	2.34 \\
 
 Mistral-v0.1~\textsubscript{\textbf{7B}}\_{SFT}  & databricks-dolly~\textsubscript{\textbf{15k}} & just-eval-instruct~\textsubscript{\textbf{1k}} & 1.12	&1.32	&1.27	&1.04&	1.11  \\ 
 
 Mistral-v0.1~\textsubscript{\textbf{7B}}\_{LFT} & databricks-dolly~\textsubscript{\textbf{15k}} & MedInstruct-test~\textsubscript{\textbf{216}} &  3.73	&4.47&	4.29	&2.52&	2.77 \\
 
 Mistral-v0.1~\textsubscript{\textbf{7B}}\_{SFT}  & databricks-dolly~\textsubscript{\textbf{15k}} & MedInstruct-test~\textsubscript{\textbf{216}} & 1.09	&1.11	&1.11	&1.03&	1.08 \\
 
 Mistral-v0.1~\textsubscript{\textbf{7B}}\_{LFT} & LIMA~\textsubscript{\textbf{1K}} & just-eval-instruct~\textsubscript{\textbf{1k}} & 3.21&	3.65&	3.3	&2.39	&2.48 \\
  
 Mistral-v0.1~\textsubscript{\textbf{7B}}\_{SFT}  & LIMA~\textsubscript{\textbf{1K}} & just-eval-instruct~\textsubscript{\textbf{1k}} & 1.49	&1.83	&1.58	&1.23	&1.5  \\ 
 
 Mistral-v0.1~\textsubscript{\textbf{7B}}\_{LFT} & LIMA~\textsubscript{\textbf{1K}} & MedInstruct-test~\textsubscript{\textbf{216}} & 4.24	&4.49	&4.48	&3.18	&3.08 \\
  
 Mistral-v0.1~\textsubscript{\textbf{7B}}\_{SFT}  & LIMA~\textsubscript{\textbf{1K}} & MedInstruct-test~\textsubscript{\textbf{216}} & 1.71&	1.83	&1.66&	1.31&	1.6 \\ 

 Mistral-v0.1~\textsubscript{\textbf{7B}}\_{LFT} & MedInstruct~\textsubscript{\textbf{1K}} & MedInstruct-test~\textsubscript{\textbf{216}} & 4.85&	4.97	&4.89&	4.77	&4.54\\
  
 Mistral-v0.1~\textsubscript{\textbf{7B}}\_{SFT}  & MedInstruct~\textsubscript{\textbf{1K}} & MedInstruct-test~\textsubscript{\textbf{216}} & 1.09&	1.56&	1.27&	1.10&	1.22\\ 

 Mistral-v0.1~\textsubscript{\textbf{7B}}\_{LFT} & MedInstruct~\textsubscript{\textbf{10K}} & MedInstruct-test~\textsubscript{\textbf{216}} & 4.92&	4.98&	4.93&	4.84&	4.6\\
  
 Mistral-v0.1~\textsubscript{\textbf{7B}}\_{SFT}  & MedInstruct~\textsubscript{\textbf{10K}} & MedInstruct-test~\textsubscript{\textbf{216}} & 1.18	&1.78	&1.38&	1.16	&1.32\\ 

 Mistral-v0.1~\textsubscript{\textbf{7B}}\_{LFT} & MedInstruct~\textsubscript{\textbf{25K}} & MedInstruct-test~\textsubscript{\textbf{216}} &  4.88	&4.98	&4.94	&4.78	&4.58\\
  
 Mistral-v0.1~\textsubscript{\textbf{7B}}\_{SFT}  & MedInstruct~\textsubscript{\textbf{25K}} & MedInstruct-test~\textsubscript{\textbf{216}} & 1.06	&1.08	&1.12	&1.04&	1.06\\

 Mistral-v0.1~\textsubscript{\textbf{7B}}\_{LFT} & MedInstruct~\textsubscript{\textbf{52K}} & MedInstruct-test~\textsubscript{\textbf{216}} & 4.9&	4.98&	4.94&	4.83&	4.56\\
  
 Mistral-v0.1~\textsubscript{\textbf{7B}}\_{SFT}  & MedInstruct~\textsubscript{\textbf{52K}} & MedInstruct-test~\textsubscript{\textbf{216}} & 1.26&	2.1	&1.87&	1.34&	1.88\\

  Mistral-v0.1~\textsubscript{\textbf{7B}}\_{LFT} & Alpaca~\textsubscript{\textbf{1K}} & just-eval-instruct~\textsubscript{\textbf{1k}} & 4.24	&4.79	&4.19	&3.51	&3.62\\
  
 Mistral-v0.1~\textsubscript{\textbf{7B}}\_{SFT}  & Alpaca~\textsubscript{\textbf{1K}} & just-eval-instruct~\textsubscript{\textbf{1k}} & 1.13	&2.6&	1.49	&1.1&	1.42 \\ 

 Mistral-v0.1~\textsubscript{\textbf{7B}}\_{LFT} & Alpaca~\textsubscript{\textbf{10K}} & just-eval-instruct~\textsubscript{\textbf{1k}} &  4.28	&4.87	&4.23&	3.38	&3.55\\
  
 Mistral-v0.1~\textsubscript{\textbf{7B}}\_{SFT}  & Alpaca~\textsubscript{\textbf{10K}} & just-eval-instruct~\textsubscript{\textbf{1k}} &  1.11&	2.34&	1.33&	1.1	&1.22\\ 

 Mistral-v0.1~\textsubscript{\textbf{7B}}\_{LFT} & Alpaca~\textsubscript{\textbf{25K}} & just-eval-instruct~\textsubscript{\textbf{1k}} &  4.3	&4.86	&4.25	&3.5	&3.67\\
  
 Mistral-v0.1~\textsubscript{\textbf{7B}}\_{SFT}  & Alpaca~\textsubscript{\textbf{25K}} & just-eval-instruct~\textsubscript{\textbf{1k}} & 1.01	&1.75	&1.13	&1	&1.03 \\

 Mistral-v0.1~\textsubscript{\textbf{7B}}\_{LFT} & Alpaca~\textsubscript{\textbf{52K}} & just-eval-instruct~\textsubscript{\textbf{1k}} &  4.3 & 4.88 &	4.29&	3.43	&3.58\\
  
 Mistral-v0.1~\textsubscript{\textbf{7B}}\_{SFT}  & Alpaca~\textsubscript{\textbf{52K}} & just-eval-instruct~\textsubscript{\textbf{1k}} & 1.04	& 1.77	& 1.2	& 1.03	& 1.09\\
 
\midrule
Phi-1.5~\textsubscript{\textbf{1.3B}}\_{LFT} & databricks-dolly~\textsubscript{\textbf{15k}} & just-eval-instruct~\textsubscript{\textbf{1k}} & 3.00 &	3.78&	3.25	&1.92	&2.33 \\
 
 Phi-1.5~\textsubscript{\textbf{1.3B}}\_{SFT}  & databricks-dolly~\textsubscript{\textbf{15k}} & just-eval-instruct~\textsubscript{\textbf{1k}} & 2.37&	3.15&	2.51&	1.69&	2.14 \\ 
 
 Phi-1.5~\textsubscript{\textbf{1.3B}}\_{LFT} & databricks-dolly~\textsubscript{\textbf{15k}} & MedInstruct-test~\textsubscript{\textbf{216}} &  3.83	&4.59&	4.38&	2.53&	2.79 \\
  
 Phi-1.5~\textsubscript{\textbf{1.3B}}\_{SFT}  & databricks-dolly~\textsubscript{\textbf{15k}} & MedInstruct-test~\textsubscript{\textbf{216}} & 3.09&	3.87&	3.45&	2.08&	2.58\\ 
 
 Phi-1.5~\textsubscript{\textbf{1.3B}}\_{LFT} & LIMA~\textsubscript{\textbf{1K}} & just-eval-instruct~\textsubscript{\textbf{1k}} & 3.98	& 4.46	&3.93	&3.85&	3.82\\
  
 Phi-1.5~\textsubscript{\textbf{1.3B}}\_{SFT}  & LIMA~\textsubscript{\textbf{1K}} & just-eval-instruct~\textsubscript{\textbf{1k}} & 3.71	&3.49	&3.18	&3.39	&3.25 \\
 
 Phi-1.5~\textsubscript{\textbf{1.3B}}\_{LFT} & LIMA~\textsubscript{\textbf{1K}} & MedInstruct-test~\textsubscript{\textbf{216}} & 4.56&	4.85&	4.54	&3.55&	3.26 \\
  
 Phi-1.5~\textsubscript{\textbf{1.3B}}\_{SFT}  & LIMA~\textsubscript{\textbf{1K}} & MedInstruct-test~\textsubscript{\textbf{216}} & 4.35 & 4.7 & 3.97 & 3.36 & 3.34 \\

 Phi-1.5~\textsubscript{\textbf{1.3B}}\_{LFT} & MedInstruct~\textsubscript{\textbf{1K}} & MedInstruct-test~\textsubscript{\textbf{216}} & 4.32 & 4.26 & 4.33 & 4.29 & 4.16 \\
  
 Phi-1.5~\textsubscript{\textbf{1.3B}}\_{SFT}  & MedInstruct~\textsubscript{\textbf{1K}} & MedInstruct-test~\textsubscript{\textbf{216}} & 3.52 & 3.74 & 3.44 & 3.31 & 3.23\\ 

 Phi-1.5~\textsubscript{\textbf{1.3B}}\_{LFT} & MedInstruct~\textsubscript{\textbf{10K}} & MedInstruct-test~\textsubscript{\textbf{216}} & 4.34 & 4.27 & 4.39 & 4.31 & 4.19 \\
  
 Phi-1.5~\textsubscript{\textbf{1.3B}}\_{SFT}  & MedInstruct~\textsubscript{\textbf{10K}} & MedInstruct-test~\textsubscript{\textbf{216}} & 3.72 & 3.84 & 3.55 & 3.44 & 3.42 \\

 Phi-1.5~\textsubscript{\textbf{1.3B}}\_{LFT} & MedInstruct~\textsubscript{\textbf{25K}} & MedInstruct-test~\textsubscript{\textbf{216}} & 4.39 & 4.28 & 4.44 & 4.33 & 4.21 \\
  
 Phi-1.5~\textsubscript{\textbf{1.3B}}\_{SFT}  & MedInstruct~\textsubscript{\textbf{25K}} & MedInstruct-test~\textsubscript{\textbf{216}} & 3.87 & 3.92 & 3.74 & 3.52 & 3.43 \\

 Phi-1.5~\textsubscript{\textbf{1.3B}}\_{LFT} & MedInstruct~\textsubscript{\textbf{52K}} & MedInstruct-test~\textsubscript{\textbf{216}} & 4.41 & 4.32 & 4.51 & 4.69 & 4.53 \\
  
 Phi-1.5~\textsubscript{\textbf{1.3B}}\_{SFT}  & MedInstruct~\textsubscript{\textbf{52K}} & MedInstruct-test~\textsubscript{\textbf{216}} & 4.15 & 4.11 & 4.10 & 4.24 & 4.30 \\

  Phi-1.5~\textsubscript{\textbf{1.3B}}\_{LFT} & Alpaca~\textsubscript{\textbf{1K}} & just-eval-instruct~\textsubscript{\textbf{1k}} & 4.07 & 4.85 & 4.06 & 3.4 & 3.64 \\
  
 Phi-1.5~\textsubscript{\textbf{1.3B}}\_{SFT}  & Alpaca~\textsubscript{\textbf{1K}} & just-eval-instruct~\textsubscript{\textbf{1k}} & 3.19 & 4.42 & 3.31 & 2.72 & 3.25 \\ 

 Phi-1.5~\textsubscript{\textbf{1.3B}}\_{LFT} & Alpaca~\textsubscript{\textbf{10K}} & just-eval-instruct~\textsubscript{\textbf{1k}} & 4.05 & 4.84 & 4.02 & 3.26 & 3.54\\
  
 Phi-1.5~\textsubscript{\textbf{1.3B}}\_{SFT}  & Alpaca~\textsubscript{\textbf{10K}} & just-eval-instruct~\textsubscript{\textbf{1k}} & 3.13 & 4.13 & 3.23 & 2.67 & 3.24
\\ 

 Phi-1.5~\textsubscript{\textbf{1.3B}}\_{LFT} & Alpaca~\textsubscript{\textbf{25K}} & just-eval-instruct~\textsubscript{\textbf{1k}} & 4.03 & 4.79 & 4.02 & 3.29 & 3.54 \\
  
 Phi-1.5~\textsubscript{\textbf{1.3B}}\_{SFT}  & Alpaca~\textsubscript{\textbf{25K}} & just-eval-instruct~\textsubscript{\textbf{1k}} & 3.04 & 4.49 & 3.11 & 2.64 & 3.28\\

 Phi-1.5~\textsubscript{\textbf{1.3B}}\_{LFT} & Alpaca~\textsubscript{\textbf{52K}} & just-eval-instruct~\textsubscript{\textbf{1k}} & 4.07 & 4.81 & 4.03 & 3.33 & 3.56 \\
  
 Phi-1.5~\textsubscript{\textbf{1.3B}}\_{SFT}  & Alpaca~\textsubscript{\textbf{52K}} & just-eval-instruct~\textsubscript{\textbf{1k}} & 3.93 & 4.2 & 3.65 & 3.88 & 3.49\\

\bottomrule
\end{tabular}
}
\caption{Results table 3}   
\label{tab:score_table_3}
\end{table*}

\def\rownumber{}
\begin{table*}[t]
\centering
\resizebox{\columnwidth}{!}{ 

\begin{tabular}{@{\makebox[3em][c]{\rownumber\space}} rcccccccccc}
\textbf{Models} & \textbf{Train Set} & \textbf{Test Set} & {\small \hspace{-1em}\textbf{Coding}} & {\small \hspace{-1em}\textbf{Info-Seek}} & {\small \hspace{-1em}\textbf{Math}} &  {\small \hspace{-1em}\textbf{Procedure}} &   {\small \hspace{-1em}\textbf{Reasoning}} &   {\small \hspace{-1em}\textbf{Role-Play}} &   {\small \hspace{-1em}\textbf{Writing}} 

 \gdef\rownumber{\stepcounter{magicrownumbers}\arabic{magicrownumbers}} \\
 \midrule
   gpt-4-1106-preview  & - & just-eval-instruct~\textsubscript{\textbf{1k}} & 4.67 & 4.78 & 4.57 & 4.87 & 4.80 & 4.76 & 4.76 \\

  \midrule

   LLaMa-2~\textsubscript{\textbf{7B}}-chat-hf & - & just-eval-instruct~\textsubscript{\textbf{1k}} & 3.31 & 3.73 & 2.54 & 4.07 & 3.93 & 3.94 & 4.07 \\

  \midrule

   LLaMa-2~\textsubscript{\textbf{13B}}-chat-hf & - & just-eval-instruct~\textsubscript{\textbf{1k}} & 3.27 & 4.02 & 2.84 & 4.22 & 4.09 & 4.11 & 4.32 \\

  \midrule

   LLaMa-2~\textsubscript{\textbf{70B}}-chat-hf & - & just-eval-instruct~\textsubscript{\textbf{1k}} & 4.02 & 4.41 & 3.67 & 4.76 & 4.56 & 4.45 & 4.81 \\

  \midrule
 
 LLaMa-2~\textsubscript{\textbf{7B}}\_{LFT} & databricks-dolly~\textsubscript{\textbf{15k}} & just-eval-instruct~\textsubscript{\textbf{1k}} &  2.38 & 2.73 & 1.81 & 2.53 & 2.55 & 2.15 & 2.36 \\
 
 LLaMa-2~\textsubscript{\textbf{7B}}\_{SFT}  & databricks-dolly~\textsubscript{\textbf{15k}} & just-eval-instruct~\textsubscript{\textbf{1k}} & 1.78 & 2.32 & 1.62 & 2.21 & 2.43 & 2.46 & 2.43  \\ 
 
 \hdashline
 LLaMa-2~\textsubscript{\textbf{7B}}\_{LFT} & LIMA~\textsubscript{\textbf{1K}} & just-eval-instruct~\textsubscript{\textbf{1k}} & 2.37 & 2.78 & 1.72 & 2.87 & 2.68 & 2.65 & 2.63 \\
  
 LLaMa-2~\textsubscript{\textbf{7B}}\_{SFT}  & LIMA~\textsubscript{\textbf{1K}} & just-eval-instruct~\textsubscript{\textbf{1k}} & 1.76 & 2.48 & 1.25 & 2.45 & 2.58 & 2.05 & 2.3  \\ 
 
 \hdashline
  LLaMa-2~\textsubscript{\textbf{7B}}\_{LFT} & Alpaca~\textsubscript{\textbf{1K}} & just-eval-instruct~\textsubscript{\textbf{1k}} & 2.47 & 3.11 & 1.95 & 3.0 & 3.1 & 2.72 & 2.97 \\
  
 LLaMa-2~\textsubscript{\textbf{7B}}\_{SFT}  & Alpaca~\textsubscript{\textbf{1K}} & just-eval-instruct~\textsubscript{\textbf{1k}} & 1.79 & 2.34 & 1.51 & 2.4 & 2.45 & 2.06 & 2.27 \\ 

 LLaMa-2~\textsubscript{\textbf{7B}}\_{LFT} & Alpaca~\textsubscript{\textbf{10K}} & just-eval-instruct~\textsubscript{\textbf{1k}} &  2.45 & 3.17 & 1.99 & 2.99 & 3.13 & 2.86 & 3.1\\
  
 LLaMa-2~\textsubscript{\textbf{7B}}\_{SFT}  & Alpaca~\textsubscript{\textbf{10K}} & just-eval-instruct~\textsubscript{\textbf{1k}} & 1.75 & 2.26 & 1.53 & 2.23 & 2.5 & 2.29 & 2.69 \\ 

 LLaMa-2~\textsubscript{\textbf{7B}}\_{LFT} & Alpaca~\textsubscript{\textbf{25K}} & just-eval-instruct~\textsubscript{\textbf{1k}} & 2.57 & 3.17 & 2.09 & 3.02 & 3.13 & 2.96 & 3.13 \\
  
 LLaMa-2~\textsubscript{\textbf{7B}}\_{SFT}  & Alpaca~\textsubscript{\textbf{25K}} & just-eval-instruct~\textsubscript{\textbf{1k}} & 1.88 & 2.36 & 1.51 & 2.41 & 2.58 & 2.52 & 2.74 \\

 LLaMa-2~\textsubscript{\textbf{7B}}\_{LFT} & Alpaca~\textsubscript{\textbf{52K}} & just-eval-instruct~\textsubscript{\textbf{1k}} & 2.74 & 3.21 & 2.09 & 3.17 & 3.15 & 2.97 & 3.3 \\
  
 LLaMa-2~\textsubscript{\textbf{7B}}\_{SFT}  & Alpaca~\textsubscript{\textbf{52K}} & just-eval-instruct~\textsubscript{\textbf{1k}} & 1.54 & 2.15 & 1.41 & 2.28 & 2.39 & 2.3 & 2.48\\
 \hdashline
 LLaMa-2~\textsubscript{\textbf{7B}}\_{LFT} & Tulu-V2-Mix~\textsubscript{\textbf{1K}} & just-eval-instruct~\textsubscript{\textbf{1k}} & 2.66 & 3.39 & 2.01 & 3.29 & 3.44 & 2.92 & 3.12 \\
  
 LLaMa-2~\textsubscript{\textbf{7B}}\_{SFT}  & Tulu-V2-Mix~\textsubscript{\textbf{1K}} & just-eval-instruct~\textsubscript{\textbf{1k}} & 1.61 & 2.2 & 1.37 & 2.33 & 2.37 & 2.11 & 2.18 \\ 

 LLaMa-2~\textsubscript{\textbf{7B}}\_{LFT} & Tulu-V2-Mix~\textsubscript{\textbf{10K}} & just-eval-instruct~\textsubscript{\textbf{1k}} & 2.58 & 3.46 & 2.14 & 3.38 & 3.54 & 3.23 & 3.34 \\
  
 LLaMa-2~\textsubscript{\textbf{7B}}\_{SFT}  & Tulu-V2-Mix~\textsubscript{\textbf{10K}} & just-eval-instruct~\textsubscript{\textbf{1k}} & 1.85 & 2.45 & 1.3 & 2.57 & 2.84 & 2.54 & 2.79 \\ 

 LLaMa-2~\textsubscript{\textbf{7B}}\_{LFT} & Tulu-V2-Mix~\textsubscript{\textbf{25K}} & just-eval-instruct~\textsubscript{\textbf{1k}} & 2.69 & 3.53 & 2.15 & 3.41 & 3.60 & 3.37 & 3.52 \\
  
 LLaMa-2~\textsubscript{\textbf{7B}}\_{SFT}  & Tulu-V2-Mix~\textsubscript{\textbf{25K}} & just-eval-instruct~\textsubscript{\textbf{1k}} & 1.95 & 2.43 & 1.31 & 2.56 & 2.84 & 2.58 & 2.86 \\

 LLaMa-2~\textsubscript{\textbf{7B}}\_{LFT} & Tulu-V2-Mix~\textsubscript{\textbf{50K}} & just-eval-instruct~\textsubscript{\textbf{1k}} & 2.82 & 3.6 & 2.16 & 3.45 & 3.67 & 3.48 & 3.6 \\
  
 LLaMa-2~\textsubscript{\textbf{7B}}\_{SFT}  & Tulu-V2-Mix~\textsubscript{\textbf{50K}} & just-eval-instruct~\textsubscript{\textbf{1k}} & 2.1 & 2.41 & 1.30 & 2.38 & 2.83 & 2.67 & 2.91 \\

 LLaMa-2~\textsubscript{\textbf{7B}}\_{LFT} & Tulu-V2-Mix~\textsubscript{\textbf{150K}} & just-eval-instruct~\textsubscript{\textbf{1k}} & 2.76 & 3.6 & 2.1 & 3.6 & 3.64 & 3.48 & 3.60 \\
  
 LLaMa-2~\textsubscript{\textbf{7B}}\_{SFT}  & Tulu-V2-Mix~\textsubscript{\textbf{150K}} & just-eval-instruct~\textsubscript{\textbf{1k}} & 2.19 & 2.53 & 1.46 & 2.66 & 2.93 & 2.84 & 3.04 \\

 LLaMa-2~\textsubscript{\textbf{7B}}\_{LFT} & Tulu-V2-Mix~\textsubscript{\textbf{326k}} & just-eval-instruct~\textsubscript{\textbf{1k}} & 2.92 & 3.63 & 2.13 & 3.57 & 3.73 & 3.6 & 3.67 \\
  
 LLaMa-2~\textsubscript{\textbf{7B}}\_{SFT}  & Tulu-V2-Mix~\textsubscript{\textbf{326k}} & just-eval-instruct~\textsubscript{\textbf{1k}} & 2.32 & 2.81 & 2.05 & 2.80 & 3.04 & 2.96 & 3.21 \\
 \hdashline
 LLaMa-2~\textsubscript{\textbf{7B}}\_{LFT} & filtered-dolly~\textsubscript{\textbf{3k}} & just-eval-instruct~\textsubscript{\textbf{1k}} & 2.31 & 2.79 & 2.24 & 2.65 & 2.67 & 2.24 & 2.45 \\
 
 LLaMa-2~\textsubscript{\textbf{7B}}\_{SFT}  & filtered-dolly~\textsubscript{\textbf{3k}} & just-eval-instruct~\textsubscript{\textbf{1k}} & 1.7 & 2.05 & 1.46 & 1.99 & 2.15 & 1.97 & 2.02 \\ 
 
  LLaMa-2~\textsubscript{\textbf{7B}}\_{LFT} & Evol-Instruct~\textsubscript{\textbf{70k}} & just-eval-instruct~\textsubscript{\textbf{1k}} & 2.78 & 3.56 & 2.08 & 3.55 & 3.66 & 3.49 & 3.76 \\
 
 LLaMa-2~\textsubscript{\textbf{7B}}\_{SFT}  & Evol-Instruct~\textsubscript{\textbf{70k}} & just-eval-instruct~\textsubscript{\textbf{1k}} & 2.23 & 2.68 & 1.73 & 2.77 & 3.05 & 2.98 & 3.19 \\ 
 
 LLaMa-2~\textsubscript{\textbf{7B}}\_{LFT} & LIMA-Simple~\textsubscript{\textbf{1k}} & just-eval-instruct~\textsubscript{\textbf{1k}} & 2.53 & 2.97 & 1.9 & 2.85 & 2.96 & 2.7 & 2.87 \\
 
 LLaMa-2~\textsubscript{\textbf{7B}}\_{SFT} & LIMA-Simple~\textsubscript{\textbf{1k}} & just-eval-instruct~\textsubscript{\textbf{1k}} & 1.78 & 2.29 & 1.39 & 2.09 & 2.51 & 2.25 & 2.28 \\

 LLaMa-2~\textsubscript{\textbf{7B}}\_{LFT} & chatgpt~\textsubscript{\textbf{9k}} & just-eval-instruct~\textsubscript{\textbf{1k}} & 2.82 & 3.53 & 2.93 & 3.28 & 3.35 & 3.52 \\
 
 LLaMa-2~\textsubscript{\textbf{7B}}\_{SFT} & chatgpt~\textsubscript{\textbf{9k}} & just-eval-instruct~\textsubscript{\textbf{1k}} & 2.58 & 3.13 & 1.82 & 3.01 &  3.15 & 2.72 & 2.93 \\ 

 \hdashline
 LLaMa-2~\textsubscript{\textbf{7B}}\_{NFT} & Alpaca~\textsubscript{\textbf{52K}} & just-eval-instruct~\textsubscript{\textbf{1k}} & 1.55 & 2.37 & 1.62 & 2.36 & 2.65 & 2.46 & 2.58 \\
 \bottomrule
\end{tabular}
}
\caption{Results table 3}   
\label{tab:score_avg_1}
\end{table*}

\def\rownumber{}
\begin{table*}[t]
\centering
\resizebox{0.93\columnwidth}{!}{ 

\begin{tabular}{@{\makebox[3em][c]{\rownumber\space}} rcccccccccc}
\textbf{Models} & \textbf{Train Set} & \textbf{Test Set} & {\small \hspace{-1em}\textbf{Coding}} & {\small \hspace{-1em}\textbf{Info-Seek}} & {\small \hspace{-1em}\textbf{Math}} &  {\small \hspace{-1em}\textbf{Procedure}} &   {\small \hspace{-1em}\textbf{Reasoning}} &   {\small \hspace{-1em}\textbf{Role-Play}} &   {\small \hspace{-1em}\textbf{Writing}} 

 \gdef\rownumber{\stepcounter{magicrownumbers}\arabic{magicrownumbers}} \\
 \midrule
 
 LLaMa-2~\textsubscript{\textbf{13B}}\_{LFT} & databricks-dolly~\textsubscript{\textbf{15k}} & just-eval-instruct~\textsubscript{\textbf{1k}} &  2.6 & 2.8 & 2.1 & 2.79 & 2.78 & 2.36 & 2.58 \\
 
 LLaMa-2~\textsubscript{\textbf{13B}}\_{SFT}  & databricks-dolly~\textsubscript{\textbf{15k}} & just-eval-instruct~\textsubscript{\textbf{1k}} & 1.93 & 2.5 & 1.46 & 2.19 & 2.58 & 1.97 & 2.39  \\ 
 
 \hdashline
 LLaMa-2~\textsubscript{\textbf{13B}}\_{LFT} & LIMA~\textsubscript{\textbf{1K}} & just-eval-instruct~\textsubscript{\textbf{1k}} & 2.45 & 2.97 & 2.08 & 2.95 & 2.78 & 2.48 & 2.83 \\
  
 LLaMa-2~\textsubscript{\textbf{13B}}\_{SFT}  & LIMA~\textsubscript{\textbf{1K}} & just-eval-instruct~\textsubscript{\textbf{1k}} & 1.6 & 2.32 & 1.34 & 2.23 & 2.45 & 1.97 & 2.22  \\
 
 \hdashline
  LLaMa-2~\textsubscript{\textbf{13B}}\_{LFT} & Alpaca~\textsubscript{\textbf{1K}} & just-eval-instruct~\textsubscript{\textbf{1k}} & 2.78 & 3.25 & 2.26 & 3.24 & 3.25 & 2.98 & 3.27 \\
  
 LLaMa-2~\textsubscript{\textbf{13B}}\_{SFT}  & Alpaca~\textsubscript{\textbf{1K}} & just-eval-instruct~\textsubscript{\textbf{1k}} & 1.86 & 2.44 & 1.67 & 2.27 & 2.55 & 2.11 & 2.49 \\ 

 LLaMa-2~\textsubscript{\textbf{13B}}\_{LFT} & Alpaca~\textsubscript{\textbf{10K}} & just-eval-instruct~\textsubscript{\textbf{1k}} & 2.78 & 3.28 & 2.46 & 3.29 & 3.3 & 3.09 & 3.39\\
  
 LLaMa-2~\textsubscript{\textbf{13B}}\_{SFT}  & Alpaca~\textsubscript{\textbf{10K}} & just-eval-instruct~\textsubscript{\textbf{1k}} & 1.87 & 2.33 & 1.51 & 2.32 & 2.52 & 2.42 & 2.8 \\ 

 LLaMa-2~\textsubscript{\textbf{13B}}\_{LFT} & Alpaca~\textsubscript{\textbf{25K}} & just-eval-instruct~\textsubscript{\textbf{1k}} & 2.75 & 3.26 & 2.55 & 3.33 & 3.26 & 3.16 & 3.36 \\
  
 LLaMa-2~\textsubscript{\textbf{13B}}\_{SFT}  & Alpaca~\textsubscript{\textbf{25K}} & just-eval-instruct~\textsubscript{\textbf{1k}} & 1.65 & 2.17 & 1.45 & 2.12 & 2.36 & 2.1 & 2.57 \\

 LLaMa-2~\textsubscript{\textbf{13B}}\_{LFT} & Alpaca~\textsubscript{\textbf{52K}} & just-eval-instruct~\textsubscript{\textbf{1k}} & 2.73 & 3.29 & 2.41 & 3.3 & 3.26 & 2.95 & 3.49 \\
  
 LLaMa-2~\textsubscript{\textbf{13B}}\_{SFT}  & Alpaca~\textsubscript{\textbf{52K}} & just-eval-instruct~\textsubscript{\textbf{1k}} & 1.74 & 2.14 & 1.34 & 2.21 & 2.31 & 2.19 & 2.66\\
 \hdashline
 LLaMa-2~\textsubscript{\textbf{13B}}\_{LFT} & Tulu-V2-Mix~\textsubscript{\textbf{1K}} & just-eval-instruct~\textsubscript{\textbf{1k}} & 2.95 & 3.49 & 2.56 & 3.61 & 3.62 & 3.31 & 3.39 \\
  
 LLaMa-2~\textsubscript{\textbf{13B}}\_{SFT}  & Tulu-V2-Mix~\textsubscript{\textbf{1K}} & just-eval-instruct~\textsubscript{\textbf{1k}} & 1.88 & 2.71 & 1.63 & 2.58 & 2.81 & 2.55 & 2.56 \\ 

 LLaMa-2~\textsubscript{\textbf{13B}}\_{LFT} & Tulu-V2-Mix~\textsubscript{\textbf{10K}} & just-eval-instruct~\textsubscript{\textbf{1k}} & 3.0 & 3.64 & 2.36 & 3.69 & 3.69 & 3.43 & 3.44 \\
  
 LLaMa-2~\textsubscript{\textbf{13B}}\_{SFT}  & Tulu-V2-Mix~\textsubscript{\textbf{10K}} & just-eval-instruct~\textsubscript{\textbf{1k}} & 1.94 & 2.62 & 1.48 & 2.63 & 2.96 & 2.84 & 3.01 \\ 

 LLaMa-2~\textsubscript{\textbf{13B}}\_{LFT} & Tulu-V2-Mix~\textsubscript{\textbf{25K}} & just-eval-instruct~\textsubscript{\textbf{1k}} & 3.02 & 3.67 & 2.42 & 3.71 & 3.73 & 3.51 & 3.50 \\
  
 LLaMa-2~\textsubscript{\textbf{13B}}\_{SFT}  & Tulu-V2-Mix~\textsubscript{\textbf{25K}} & just-eval-instruct~\textsubscript{\textbf{1k}} & 2.01 & 2.82 & 1.72 & 2.84 & 3.08 & 2.95 & 3.14 \\

 LLaMa-2~\textsubscript{\textbf{13B}}\_{LFT} & Tulu-V2-Mix~\textsubscript{\textbf{50K}} & just-eval-instruct~\textsubscript{\textbf{1k}} & 3.14 & 3.75 & 2.39 & 3.72 & 3.76 & 3.77 & 3.81 \\
  
 LLaMa-2~\textsubscript{\textbf{13B}}\_{SFT}  & Tulu-V2-Mix~\textsubscript{\textbf{50K}} & just-eval-instruct~\textsubscript{\textbf{1k}} & 2.11 & 2.89 & 1.87 & 2.92 & 3.12 & 3.01 & 3.18 \\

 LLaMa-2~\textsubscript{\textbf{13B}}\_{LFT} & Tulu-V2-Mix~\textsubscript{\textbf{150K}} & just-eval-instruct~\textsubscript{\textbf{1k}} & 3.17 & 3.79 & 2.44 & 3.75 & 3.88 & 3.82 & 3.91 \\
  
 LLaMa-2~\textsubscript{\textbf{13B}}\_{SFT}  & Tulu-V2-Mix~\textsubscript{\textbf{150K}} & just-eval-instruct~\textsubscript{\textbf{1k}} & 2.29 & 2.95 & 1.98 & 3.02 & 3.21 & 3.27 & 3.38 \\

 LLaMa-2~\textsubscript{\textbf{13B}}\_{LFT} & Tulu-V2-Mix~\textsubscript{\textbf{326k}} & just-eval-instruct~\textsubscript{\textbf{1k}} & 3.20 & 3.81 & 2.54 & 3.80 & 3.91 & 3.94 & 3.99 \\
  
 LLaMa-2~\textsubscript{\textbf{13B}}\_{SFT}  & Tulu-V2-Mix~\textsubscript{\textbf{326k}} & just-eval-instruct~\textsubscript{\textbf{1k}} & 2.36 & 3.01 & 2.31 & 3.14 & 3.32 & 3.36& 3.45 \\
 \hdashline
  
 LLaMa-2~\textsubscript{\textbf{13B}}\_{LFT} & LIMA-Simple~\textsubscript{\textbf{1k}} & just-eval-instruct~\textsubscript{\textbf{1k}} & 2.72 & 3.23 & 2.23 & 3.31 & 3.18 & 2.83 & 3.0 \\
 
 LLaMa-2~\textsubscript{\textbf{13B}}\_{SFT} & LIMA-Simple~\textsubscript{\textbf{1k}} & just-eval-instruct~\textsubscript{\textbf{1k}} & 1.77 & 2.4 & 1.75 & 2.31 & 2.67 & 2.23 & 2.32 \\

 \midrule
 LLaMa-2~\textsubscript{\textbf{70B}}\_{LFT} & databricks-dolly~\textsubscript{\textbf{15k}} & just-eval-instruct~\textsubscript{\textbf{1k}} & 3.44 & 3.59 & 2.65 & 3.46 & 3.69 & 3.87 & 3.63 \\
 
 LLaMa-2~\textsubscript{\textbf{70B}}\_{SFT}  & databricks-dolly~\textsubscript{\textbf{15k}} & just-eval-instruct~\textsubscript{\textbf{1k}} & 2.42 & 2.53 & 2.33 & 2.45 & 2.61 & 2.45 & 2.78 \\ 
 \hdashline
 LLaMa-2~\textsubscript{\textbf{70B}}\_{LFT} & LIMA~\textsubscript{\textbf{1K}} & just-eval-instruct~\textsubscript{\textbf{1k}} & 3.32 & 3.54 & 2.59 & 3.40 & 3.58 & 3.72 & 3.58\\
  
 LLaMa-2~\textsubscript{\textbf{70B}}\_{SFT}  & LIMA~\textsubscript{\textbf{1K}} & just-eval-instruct~\textsubscript{\textbf{1k}} & 2.36 & 2.49 & 2.31 & 2.41 & 2.57 & 2.43 & 2.71 \\
 
\hdashline
LLaMa-2~\textsubscript{\textbf{70B}}\_{LFT} & Alpaca~\textsubscript{\textbf{1K}} & just-eval-instruct~\textsubscript{\textbf{1k}} & 3.57 & 3.72 & 2.92 & 3.73 & 3.88 & 4.01 & 3.92 \\
  
 LLaMa-2~\textsubscript{\textbf{70B}}\_{SFT}  & Alpaca~\textsubscript{\textbf{1K}} & just-eval-instruct~\textsubscript{\textbf{1k}} &  3.21 & 2.86 & 2.64 & 2.76 & 2.87 & 2.76 & 3.03 \\ 

 LLaMa-2~\textsubscript{\textbf{70B}}\_{LFT} & Alpaca~\textsubscript{\textbf{52K}} & just-eval-instruct~\textsubscript{\textbf{1k}} & 3.89 & 3.99 & 3.46 & 4.01 & 4.12 & 4.23 & 4.19\\
  
 LLaMa-2~\textsubscript{\textbf{70B}}\_{SFT}  & Alpaca~\textsubscript{\textbf{52K}} & just-eval-instruct~\textsubscript{\textbf{1k}} & 3.42 & 3.14 & 3.05 & 3.10 & 3.33 & 3.11 & 3.45\\
\hdashline
  LLaMa-2~\textsubscript{\textbf{70B}}\_{LFT} & Tulu-V2-Mix~\textsubscript{\textbf{1k}} & just-eval-instruct~\textsubscript{\textbf{1k}} & 4.01 & 4.14 & 3.53 & 4.07 & 4.12 & 4.30 & 4.22\\
  
 LLaMa-2~\textsubscript{\textbf{70B}}\_{SFT}  & Tulu-V2-Mix~\textsubscript{\textbf{1k}} & just-eval-instruct~\textsubscript{\textbf{1k}} & 3.71 & 3.42 & 3.32 & 3.13 & 3.76 & 3.32 & 3.72\\

 LLaMa-2~\textsubscript{\textbf{70B}}\_{LFT} & Tulu-V2-Mix~\textsubscript{\textbf{326k}} & just-eval-instruct~\textsubscript{\textbf{1k}} & 4.31 & 4.44 & 3.81 & 4.27 & 4.42 & 4.67 & 4.57 \\
  
 LLaMa-2~\textsubscript{\textbf{70B}}\_{SFT}  & Tulu-V2-Mix~\textsubscript{\textbf{326k}} & just-eval-instruct~\textsubscript{\textbf{1k}} & 3.92 & 3.72 & 3.62 & 3.35 & 4.11 & 3.67 & 4.01\\

 \midrule
 Mistral-v0.1~\textsubscript{\textbf{7B}}\_{LFT} & databricks-dolly~\textsubscript{\textbf{15k}} & just-eval-instruct~\textsubscript{\textbf{1k}} & 3.1 & 2.99 & 2.6 & 3.1 & 2.86 & 2.63 & 2.8 \\
 
 Mistral-v0.1~\textsubscript{\textbf{7B}}\_{SFT}  & databricks-dolly~\textsubscript{\textbf{15k}} & just-eval-instruct~\textsubscript{\textbf{1k}} & 1.03 & 1.19 & 1.05 & 1.18 & 1.15 & 1.17 & 1.17  \\ 
 
 Mistral-v0.1~\textsubscript{\textbf{7B}}\_{LFT} & LIMA~\textsubscript{\textbf{1K}} & just-eval-instruct~\textsubscript{\textbf{1k}} & 3.14 & 3.01 & 2.5 & 3.28 & 2.85 & 2.59 & 3.0 \\
  
 Mistral-v0.1~\textsubscript{\textbf{7B}}\_{SFT}  & LIMA~\textsubscript{\textbf{1K}} & just-eval-instruct~\textsubscript{\textbf{1k}} & 1.28 & 1.54 & 1.19 & 1.54 & 1.63 & 1.43 & 1.59  \\ 

  Mistral-v0.1~\textsubscript{\textbf{7B}}\_{LFT} & Alpaca~\textsubscript{\textbf{1K}} & just-eval-instruct~\textsubscript{\textbf{1k}} & 3.67 & 4.09 & 3.15 & 4.12 & 4.21 & 4.1 & 3.98\\
  
 Mistral-v0.1~\textsubscript{\textbf{7B}}\_{SFT}  & Alpaca~\textsubscript{\textbf{1K}} & just-eval-instruct~\textsubscript{\textbf{1k}} & 1.26 & 1.55 & 1.53 & 1.49 & 1.55 & 1.47 & 1.65\\ 

 Mistral-v0.1~\textsubscript{\textbf{7B}}\_{LFT} & Alpaca~\textsubscript{\textbf{10K}} & just-eval-instruct~\textsubscript{\textbf{1k}} & 3.79 & 4.06 & 3.37 & 4.09 & 4.12 & 4.13 & 4.14\\
  
 Mistral-v0.1~\textsubscript{\textbf{7B}}\_{SFT}  & Alpaca~\textsubscript{\textbf{10K}} & just-eval-instruct~\textsubscript{\textbf{1k}} & 1.23 & 1.41 & 1.32 & 1.20 & 1.37 & 1.26 & 1.45\\

 Mistral-v0.1~\textsubscript{\textbf{7B}}\_{LFT} & Alpaca~\textsubscript{\textbf{25K}} & just-eval-instruct~\textsubscript{\textbf{1k}} &  3.73 & 4.11 & 3.32 & 4.18 & 4.22 & 4.13 & 4.23\\
  
 Mistral-v0.1~\textsubscript{\textbf{7B}}\_{SFT}  & Alpaca~\textsubscript{\textbf{25K}} & just-eval-instruct~\textsubscript{\textbf{1k}} & 1.12 & 1.18 & 1.21 & 1.19 & 1.17 & 1.13 & 1.23\\

 Mistral-v0.1~\textsubscript{\textbf{7B}}\_{LFT} & Alpaca~\textsubscript{\textbf{52K}} & just-eval-instruct~\textsubscript{\textbf{1k}} & 3.84 & 4.09 & 3.37 & 4.13 & 4.18 & 3.98 & 4.09\\
  
 Mistral-v0.1~\textsubscript{\textbf{7B}}\_{SFT}  & Alpaca~\textsubscript{\textbf{52K}} & just-eval-instruct~\textsubscript{\textbf{1k}} & 1.18 & 1.22 & 1.34 & 1.21 & 1.22 & 1.3 & 1.26\\
 
\midrule
Phi-1.5~\textsubscript{\textbf{1.3B}}\_{LFT} & databricks-dolly~\textsubscript{\textbf{15k}} & just-eval-instruct~\textsubscript{\textbf{1k}} & 2.78 & 2.79 & 2.26 & 2.97 & 2.92 & 2.72 & 3.03 \\
 
 Phi-1.5~\textsubscript{\textbf{1.3B}}\_{SFT}  & databricks-dolly~\textsubscript{\textbf{15k}} & just-eval-instruct~\textsubscript{\textbf{1k}} & 1.9 & 2.31 & 1.79 & 2.36 & 2.58 & 2.57 & 2.55 \\ 
 
 Phi-1.5~\textsubscript{\textbf{1.3B}}\_{LFT} & LIMA~\textsubscript{\textbf{1K}} & just-eval-instruct~\textsubscript{\textbf{1k}} & 3.02 & 2.95 & 2.46 & 3.12 & 2.65 & 2.39 & 2.97\\
  
 Phi-1.5~\textsubscript{\textbf{1.3B}}\_{SFT}  & LIMA~\textsubscript{\textbf{1K}} & just-eval-instruct~\textsubscript{\textbf{1k}} & 2.58 & 2.64 & 2.32 & 2.51 & 2.42 & 2.12 & 2.78\\
 
 Phi-1.5~\textsubscript{\textbf{1.3B}}\_{LFT} & Alpaca~\textsubscript{\textbf{1K}} & just-eval-instruct~\textsubscript{\textbf{1k}} & 3.24 & 4.09 & 3.15 & 4.12 & 4.21 & 4.10 & 3.98 \\
  
 Phi-1.5~\textsubscript{\textbf{1.3B}}\_{SFT}  & Alpaca~\textsubscript{\textbf{1K}} & just-eval-instruct~\textsubscript{\textbf{1k}} & 1.24 & 1.51 & 1.33 & 1.47 & 1.58 & 1.57 & 1.87\\ 

 Phi-1.5~\textsubscript{\textbf{1.3B}}\_{LFT} & Alpaca~\textsubscript{\textbf{10K}} & just-eval-instruct~\textsubscript{\textbf{1k}} & 3.27 & 4.12 & 3.20 & 4.15 & 4.21 & 4.12 & 3.99 \\
  
 Phi-1.5~\textsubscript{\textbf{1.3B}}\_{SFT}  & Alpaca~\textsubscript{\textbf{10K}} & just-eval-instruct~\textsubscript{\textbf{1k}} & 1.45 & 1.51 & 1.47 & 1.65 & 1.75 & 1.82 & 2.01 \\

 Phi-1.5~\textsubscript{\textbf{1.3B}}\_{LFT} & Alpaca~\textsubscript{\textbf{25K}} & just-eval-instruct~\textsubscript{\textbf{1k}} & 3.29 & 4.13 & 3.22 & 4.17 & 4.23 & 4.11 &  4.00 \\
  
 Phi-1.5~\textsubscript{\textbf{1.3B}}\_{SFT}  & Alpaca~\textsubscript{\textbf{25K}} & just-eval-instruct~\textsubscript{\textbf{1k}} & 1.71 & 1.82 & 1.72 & 1.83 & 1.81 & 2.01 & 2.14 \\

 Phi-1.5~\textsubscript{\textbf{1.3B}}\_{LFT} & Alpaca~\textsubscript{\textbf{52K}} & just-eval-instruct~\textsubscript{\textbf{1k}} & 3.28 & 4.10 & 3.21 & 4.19 & 4.25 & 4.12 &  4.03\\
  
 Phi-1.5~\textsubscript{\textbf{1.3B}}\_{SFT}  & Alpaca~\textsubscript{\textbf{52K}} & just-eval-instruct~\textsubscript{\textbf{1k}} & 1.59 & 1.92 & 1.72 & 1.97 & 2.13 & 2.23 & 3.02 \\

\bottomrule
\end{tabular}
}
\caption{Results table 3}   
\label{tab:score_avg_2}
\end{table*}



\begin{figure}[bth]
  \centering
  \subfloat[Model fine-tuned using SFT.]
  {\includegraphics[width=0.45\textwidth]{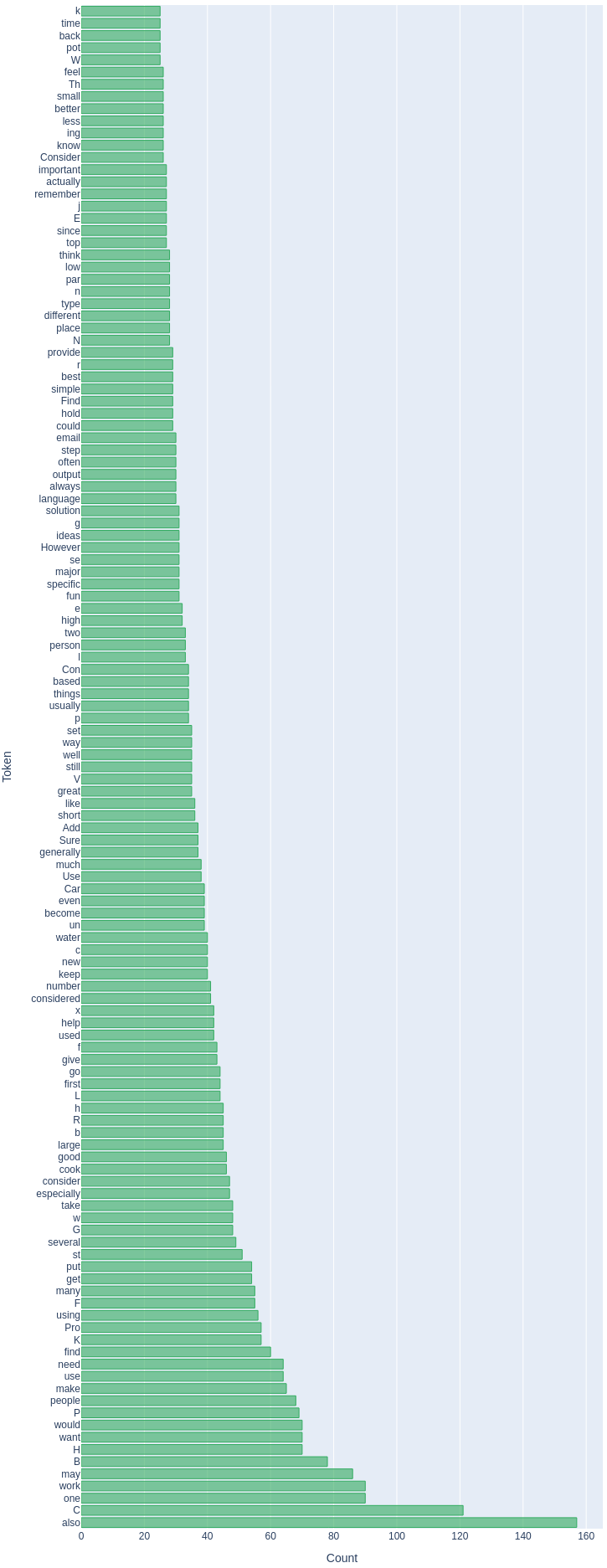}\label{fig:f1}}
  \hfill
  \subfloat[Model fine-tuned using LFT.]{\includegraphics[width=0.45\textwidth]{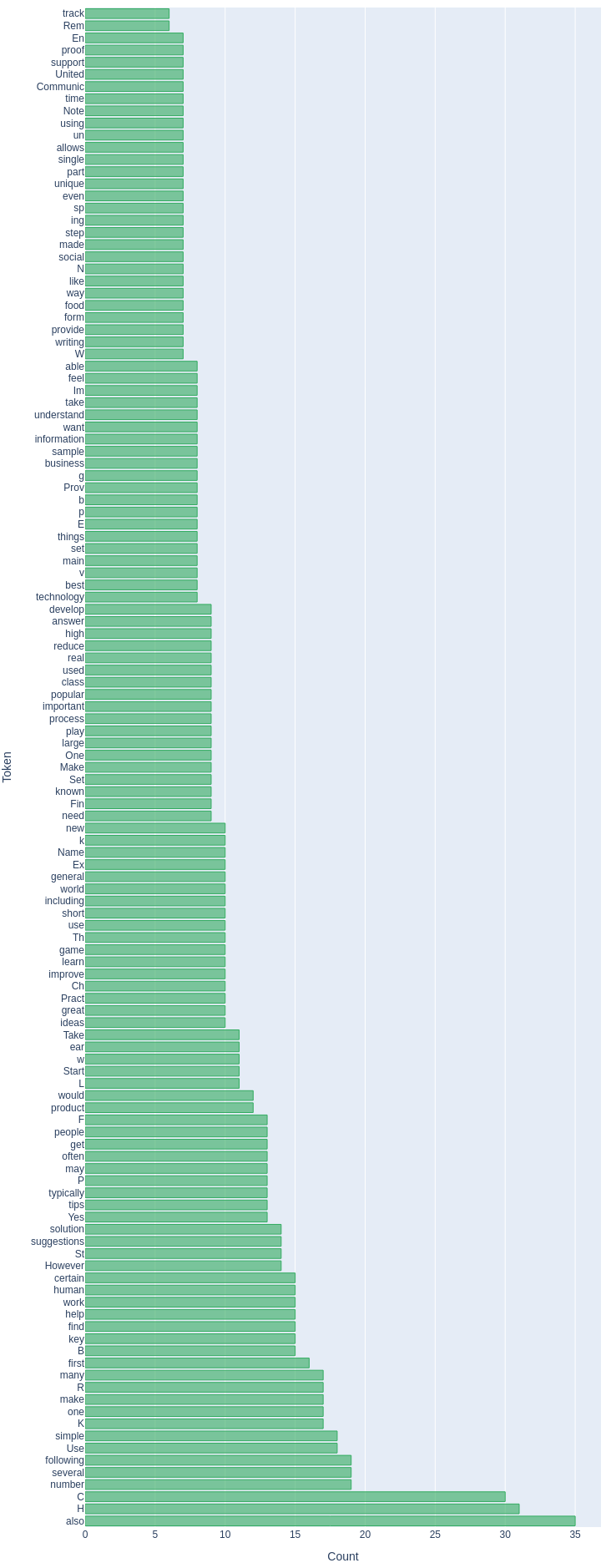}\label{fig:f2}}
  \caption{\small Frequency distribution plot of top 125 \textcolor{red}{shifted} and \textcolor{orange}{marginal} for LLaMa-2~\textsubscript{\textbf{7B}} trained on LIMA~\textsubscript{\textbf{1K}} and inferred on just-eval-instruct~\textsubscript{\textbf{1k}}. We visualize frequency distributions of models fine-tuned on both SFT and LFT to compare the knowledge learned by both models in the fine-tuning stage.}
\end{figure}

\end{document}